\documentclass[10pt]{article}
\usepackage{style}
\usepackage{placeins}
\usepackage{subcaption}
\usepackage{graphicx}
\usepackage{bm}

\usepackage[utf8]{inputenc} % allow utf-8 input
\usepackage[T1]{fontenc}    % use 8-bit T1 fonts
\usepackage{hyperref}       % hyperlinks
\usepackage{url}            % simple URL typesetting
\usepackage{booktabs}       % professional-quality tables
\usepackage{nicefrac}       % compact symbols for 1/2, etc.
\usepackage{microtype}      % microtypography

\usepackage{cite}
\usepackage{caption}
\usepackage{multirow}
\usepackage{amsfonts}
\usepackage{amsmath}
\usepackage{amssymb}
\usepackage{mathrsfs}
\usepackage[dvipsnames]{xcolor}
\usepackage{mathtools}
\usepackage{float}
\usepackage{soul}
\usepackage{subcaption}
\usepackage{hyperref} 
\usepackage{cleveref}

\hypersetup{
	colorlinks=true,
	linkcolor=blue,
	citecolor=blue,
	urlcolor=blue,
}

\newcommand\comment[1]{}

\newcommand\EndFCPINO{\textsc{PINO In--FC}}
\newcommand\FCPINO{\textsc{FC--PINO}}
\newcommand\padPINO{\textsc{Pad--PINO}}
\newcommand\PINOPadOut{\textsc{PINO Out--Pad}}
\newcommand\standardPINO{\textsc{Standard PINO}}
\newcommand\outsideFCPINO{\textsc{PINO Out--FC}}

\title{FC-PINO: High-Precision Physics-Informed \\ Neural Operators via Fourier Continuation}

\author{%
  {\fontsize{11}{13}\selectfont
    Adarsh Ganeshram, Valentin Duruisseaux, Haydn Maust, Zongyi Li, Yixuan Wang,\\
    Daniel Leibovici, Oscar P. Bruno, Thomas Y. Hou, Anima Anandkumar}\\[0.4em]
  {\normalfont\small\itshape
    California Institute of Technology, 
    Pasadena, CA, USA}
}

\begin{document}

\maketitle

\begin{abstract}

\vspace{-2.5mm}

The physics-informed neural operator (PINO) is a machine learning paradigm that has demonstrated promising results for learning solutions to partial differential equations (PDEs). It leverages the Fourier Neural Operator (FNO) to learn solution operators in function spaces and leverages physics losses during training to penalize deviations from known physics laws. Spectral differentiation provides an efficient way to compute derivatives for the physics losses, but it inherently assumes periodicity. When applied to non-periodic functions, this assumption of periodicity can lead to significant errors, including Gibbs phenomena near domain boundaries which degrade the accuracy of both function representations and derivative computations, especially for higher order derivatives. To overcome this limitation, we introduce the \FCPINO{} (Fourier‑Continuation-based Physics‑Informed Neural Operator) architecture which extends the accuracy and efficiency of PINO and spectral differentiation to non-periodic and non-smooth PDEs. In \FCPINO{}, we propose integrating Fourier continuation into the PINO framework, and test two different continuation approaches: \textsc{FC--Legendre} and \textsc{FC--Gram}. By transforming non-periodic signals into periodic functions on extended domains in a well-conditioned manner, Fourier continuation enables fast and accurate derivative computations. This approach avoids the discretization sensitivity of finite differences and the memory overhead of automatic differentiation. We demonstrate that standard PINO fails (without padding) or struggles (even with padding) to solve non-periodic and non-smooth PDEs with high precision, across challenging benchmarks.  
In contrast, the proposed \FCPINO{} provides accurate, robust, and scalable solutions, substantially outperforming PINO alternatives, and demonstrating that Fourier continuation is critical for extending PINO to a wider range of PDE problems when high-precision solutions are needed. 
\end{abstract}

\vspace{1mm}

\section{Introduction}

Partial differential equations (PDEs) are essential for modeling complex physical phenomena and dynamical systems in science and engineering. However, conventional numerical solvers become computationally intractable when applied to large-scale systems. To overcome these limitations, recent machine learning methods have been developed to learn solution operators of PDEs directly from data~\citep{kovachki2023neural}. Among these, neural operators~\citep{FNO_OG, azizzadenesheli2024neural, Berner2025Principles} have attracted considerable interest for their ability to process inputs defined on arbitrary grids and produce outputs at any spatial location. Moreover, they are supported by universal approximation guarantees for nonlinear operators~\citep{Kovachki2021_2}.

However, fully data-driven models can struggle in regimes with sparse or low-resolution data~\citep{PINO_OG}. To mitigate this, prior knowledge of physics or conservation laws can be integrated into the learning process. Physics-Informed Neural Networks (PINNs)~\citep{Raissi2017Part1,Raissi2017Part2,PINN_OG} are neural network representations of PDE solutions which are learned by minimizing deviations away from physics laws. While PINNs are flexible and powerful, the associated optimization problem is often difficult, characterized by a highly nonconvex and non-smooth loss landscape~\citep{fuks2020limitations,wang2021understanding, krishnapriyan2021characterizing,PINN_nonlinear_2022}. 
More recently, the Physics-Informed Neural Operator (PINO)~\citep{PINO_OG} has been proposed and shown empirically to outperform PINNs on many tasks~\citep{rosofsky2022applications}. The original PINO relies on a Fourier Neural Operator (FNO)~\citep{FNO_OG,duruisseaux2025FNOGuide} which parametrizes the solution operator to a family of PDEs. Learning this operator in function space can be easier than the optimization problem arising when learning a single solution to a single PDE instance using PINNs. While PINO was first proposed with a FNO as the backbone neural operator, it naturally extends to other neural operators such as graph neural operators~\citep{lin2025mGNO}.

The accurate and efficient computation of derivatives is critical in PINO, as small numerical errors can be amplified through the physics losses and degrade solution quality. This is particularly critical in applications where high-precision solutions are required such as blowup modeling in fluid equations. Singularity formation, or blowups, for the Navier--Stokes equation poses one of the Clay Prize problems and is one of the most challenging problems in PDE analysis, with implications for applications in fluid modeling and beyond.  While proving finite-time blowup for fluid equations under certain initial conditions remains a theoretical challenge, empirical approaches can play a valuable role in identifying potential blowup scenarios and approximate candidates for such events. If we can bound the error on these blowup candidates, then there is hope that (non)linear stability theory is sufficient to conclude that blowup does actually occur at a perturbation around the blowup candidate~\citep{chen2021finite, chen2022stable, wang2025highprecisionpinnsunbounded}.

One simple method of computing derivatives is finite differences, but it can require very fine resolutions to be accurate and can struggle with multi-scale or rapidly-varying dynamics. Automatic differentiation, widely used in PINNs~\citep{Baydin2017}, computes exact pointwise derivatives by applying the chain rule over a computational graph. It avoids the discretization error of finite differences but its memory cost can be prohibitive in large or deep models. 

On the other hand, \textbf{spectral differentiation} (also know as Fourier differentiation) offers a compelling alternative for smooth, noise-free, periodic functions by computing derivatives in frequency space via simple multiplications of Fourier coefficients. It is accurate and has a cost of only $\mathcal{O}(N\log N)$ for $N$ grid points, with no increase in cost for higher derivatives. It is significantly more memory-efficient than automatic differentiation, especially for larger grids and larger models, as it avoids storing intermediate states.

However, like other spectral methods, spectral differentiation implicitly assumes periodicity. When applied to non-periodic functions, it suffers from the Gibbs phenomenon which results in a loss of accuracy. While this loss of accuracy may be acceptable in certain contexts, it is a fundamental bottleneck when attempting to achieve high-precision solutions. This limitation motivates the development of Fourier-based techniques that remain highly accurate and robust for non-periodic inputs in physics-informed machine learning.

One possibility is to use the Fourier series on a larger domain by padding or expanding the original domain. Such extension grants FNOs and PINOs the necessary expressivity to represent solutions for general non-periodic problems. However, a severe ill-conditioning issue can arise since the discrete Fourier transform is a Vandermonde matrix~\citep{Higham2002,press2007numerical,adcock2014numerical} if the extension is not done carefully. Zero-padding a signal on the sides introduces artificial discontinuities at the endpoints. When applying Fourier transforms to such signals, these discontinuities can result in Gibbs phenomena and spurious oscillations that do not reflect the true behavior of the underlying function (see  \Cref{fig:gibbs_visualizations} ). A common alternative to zero-padding is mirror padding, in which a signal is extended by reflecting it about one of its endpoints. This technique guarantees continuity of the signal at the mirror point, but introduces a discontinuity in the derivative unless the function is locally constant at that point. Consequently, the extended signal remains continuous but is not differentiable at the mirror point. Since the convergence rate of Fourier series depends on the regularity of the function being approximated, this lack of differentiability leads to very slow spectral convergence and persistent artifacts in the frequency domain representation. 

\begin{figure}[htb]  
\centering   
\includegraphics[width=\textwidth]{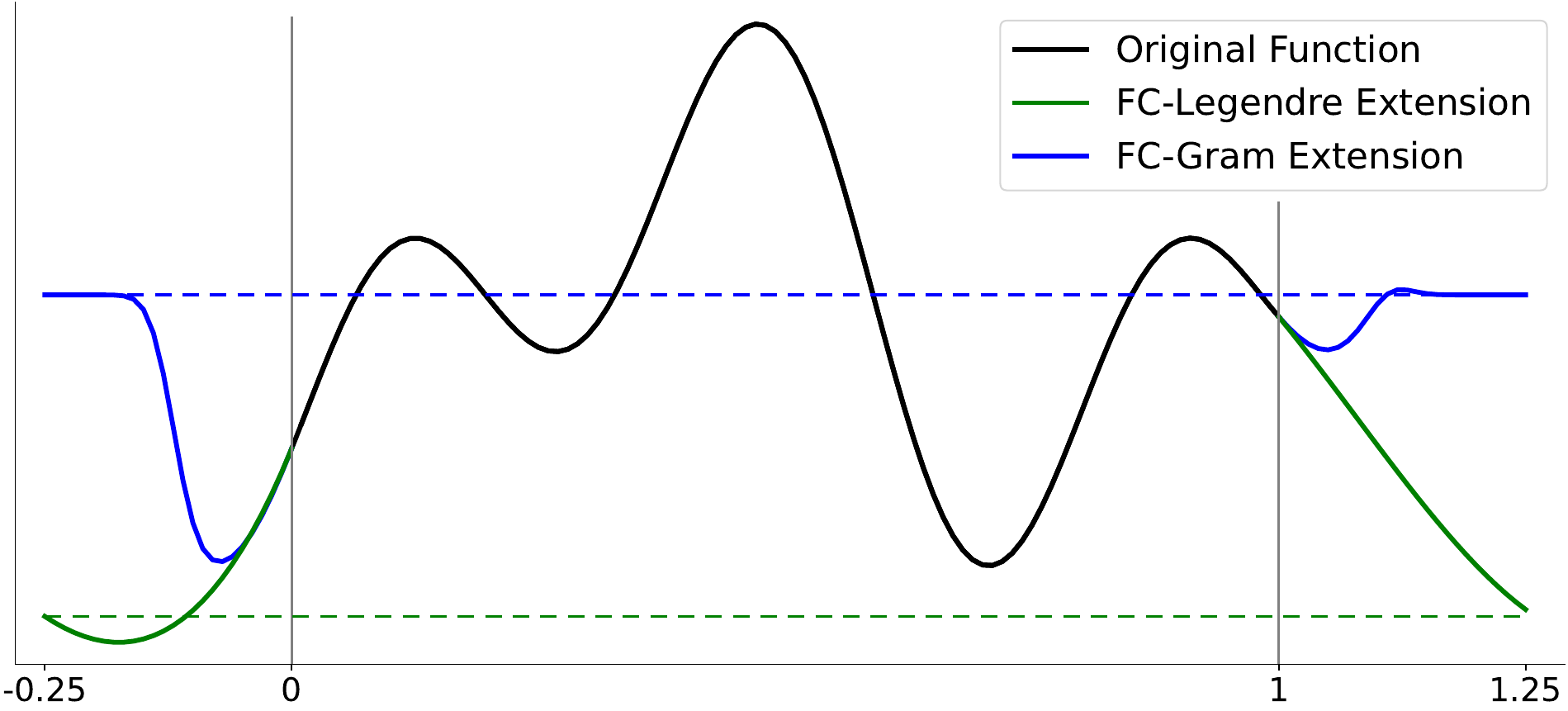} \vspace{0.1mm}

\includegraphics[width=\textwidth]{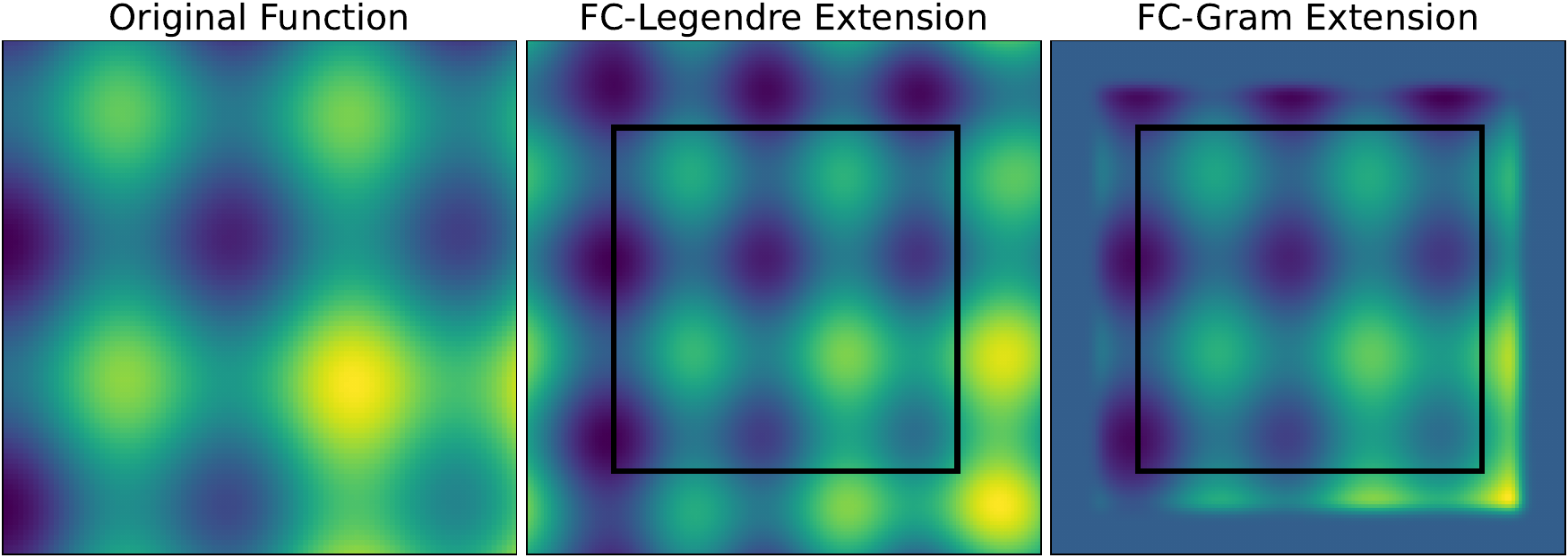} 
\caption{Examples of periodic extensions obtained using the Fourier continuation methods. \emph{(top)} Extensions of $f(x) = \sin(16x) - \cos(8x)$ from $[0,1]$ to $[-0.25,1.25]$. \emph{(bottom)} Extensions of $f(x,y) = \sin(12x) - \cos(14y) + 3xy$ from $[0,1]^2$ to $[-0.25,1.25]^2$. 
\label{fig: FC Examples} }  \vspace{1mm}
\end{figure}

Ideally, to apply Fourier methods to problems with non-periodic solutions, one must convert the signal of interest to a periodic function in a well-conditioned manner. \textbf{Fourier continuation (FC)} methods~\citep{Fourier_continuation_2011, fontana2020Fourier, bruno2022two} address this challenge by extending non-periodic signals to periodic ones over an enlarged domain through polynomial interpolation (see \Cref{fig: FC Examples} for illustrations). These methods have been shown empirically to achieve superalgebraic convergence rates, making them particularly effective for spectral approximations of smooth, non-periodic data. \\

\noindent \textbf{Proposed Approach.} We propose to use Fourier continuation (FC) as a mechanism that enables PINO to be applied to non-periodic functions with derivatives computed using spectral differentiation while retaining efficiency and high numerical precision. 
More precisely, we introduce the \textbf{Fourier-Continuation-based Physics-Informed Neural Operator} (\FCPINO{}), in which a FC layer is applied at the beginning of the FNO architecture. 
The FNO then operates on an extended domain, and its output is restricted immediately before the final projection operator. 
In \FCPINO{}, derivatives of the model output are computed using the chain rule by combining the analytic derivative of the projection operator with spectral differentiation on the extended domain of the pre-projection output. 
Because the FC layer enforces periodicity on the extended domain throughout the computation, spectral differentiation can be applied accurately and effectively before the restriction to the original domain. 
By construction, \FCPINO{} combines the universal operator approximation capabilities of neural operators with the convergence guarantees of FC, leading to errors in the solution derivatives that provably decay at an algebraic rate with increasing resolution. We also consider the use of two different FC methods as part of \FCPINO{}, the \textsc{FC--Legendre} and (Accelerated) \textsc{FC--Gram}~\citep{AMLANI2016} methods.  \\

\noindent \textbf{Summary of Results.} After introducing in details the \FCPINO{} approach and FC methods, we demonstrate that \FCPINO{} inherits the universal operator approximation guarantees of neural operators while leveraging Fourier continuation to control boundary artifacts. As a result, the $k$-th derivative errors converge at the algebraic rate $\mathcal{O}(N^{-(d-k)})$ as the resolution $N$ increases, where $d$ is a parameter of the Fourier continuation scheme.

We then conduct numerical experiments and ablation studies on the self-similar Burgers' equation, viscous 2D Burgers' equation, and 3D Navier--Stokes equations, where we are particularly interested in obtaining high-precision solutions. These problems exemplify the difficulties faced by PINO when spectral differentiation is used naively or only with zero-padding. Indeed, in our numerical experiments, the standard PINO failed to solve any of the PDEs considered, and even zero-padding applied outside the model before spectral differentiation did not remedy this issue. 
A more sophisticated padding strategy within the FNO architecture yielded some improvement, but remained inadequate for high-precision regimes. 
By contrast, the proposed \FCPINO{} robustly solved all problems with substantially higher accuracy, consistently outperforming the strongest padding baseline by several orders of magnitude. 
For example, on the self-similar Burgers’ equation, \FCPINO{} reduced PDE residuals from $10^{-8}$ to $10^{-12}$ while simultaneously lowering boundary and smoothness losses by 4--5 orders of magnitude. Altogether, our results demonstrate that \FCPINO{} can robustly handle non-periodic and non-smooth problems, while remaining scalable as a result of spectral differentiation, when compared with automatic differentiation in PINNs and PINO.

Overall, our findings strongly suggest that FC-based enhancements are critical in PINO when using spectral differentiation, especially when high-precision solutions are required. \FCPINO{} offers a principled and scalable path forward for extending PINO to a broader class of high-precision PDE problems.  \\

\section{Preliminaries}

\subsection{Problem Setup}

We consider both stationary and dynamic systems. Let $D \subset \mathbb R^d$ be a bounded open set, and let $g$ be a function on the boundary~$\partial D$. Let $\mathcal{U}$ and $\mathcal{V}$ be Banach spaces. 

In the stationary setting, given a coefficient function \(a \in \mathcal{A} \subseteq \mathcal{V}\), we want a solution \(u \in \mathcal{U}\) to the PDE 
\begin{equation}\label{eq: PDE setup stationary}
    \begin{split}
    \mathcal{P}(u,a) &= 0 \qquad \text{in } D \\
    u &= g \qquad \text{in } \partial D
    \end{split}
\end{equation}
where \(\mathcal{P}\) is a nonlinear partial differential operator. 

In the dynamic setting, given an initial condition \(a = u(0) \in \mathcal{A} \subseteq \mathcal{U}\), we want the solution \(u(t) \in \mathcal{U}\) for $t>0$ to the PDE 
\begin{align}\label{eq: PDE setup dynamic}
\begin{split}
    \frac{du}{dt} &= \mathcal{R}(u) \qquad  \text{in } D \times (0,\infty) \\
    u &= g \qquad \quad \:\:\: \text{in } \partial D \times (0,\infty) \\
    u &= a \qquad \quad \   \text{ in } (D \cup \partial D) \times \{0\}
\end{split}
\end{align}
where $\mathcal{R}$ is a nonlinear partial differential operator. \\

\hfill 

\subsection{Physics-Informed Machine Learning}

\subsubsection{Physics-Informed Neural Networks (PINNs)}

Physics-Informed Neural Networks (PINNs)~\citep{Raissi2017Part1,Raissi2017Part2,PINN_OG} are neural networks designed to approximate solutions to PDEs. Their parameters are optimized by minimizing discrepancies not only from known solutions of the PDEs (when available) but also from physics constraints such as conservation laws and the governing PDEs. Over the past few years, a wide range of PINN variants have been developed and successfully applied to solve PDEs across various domains~\citep{Jagtap2020_2,Cai2022,Yu2022}, including blowup problems~\citep{PINN_3rd_deriv, wang2025high}. Notably, PINNs can be trained without relying on data, using only knowledge of the governing differential equations and of the associated boundary condition. Note that a PINN only learns the solution for a single PDE instance and does not generalize to new PDE instances without further optimization. \\

To solve the stationary problem \eqref{eq: PDE setup stationary}, a PINN takes points in $D$ as input and produces an approximation~$u_\theta(x)$ to the solution at each point $x \in D$ by minimizing the residuals of equations \eqref{eq: PDE setup stationary} in an appropriate function space norm. A typical choice is \( L^2(D)\), giving the loss function
\begin{align}
\label{eq:pinns-stationary}
\begin{split}
    \mathcal{L}_{\text{pde}}(a, u_\theta)  & \ = \  \big\|\mathcal{P}(a,u_{\theta})\big\|^2_{L^2(D)} \ + \ \alpha \ \big\|u_{\theta}|_{\partial D} - g \big\|^2_{L^2(\partial D)} \\ 
    & \ = \  \int_D |\mathcal{P}(u_{\theta}(x),a(x))|^2\mathrm{d}x  \ + \ \alpha\int_{\partial D} |u_{\theta}(x) - g(x)|^2 \mathrm{d}x.
\end{split}
\end{align}

\hfill 

To solve the dynamic problem \eqref{eq: PDE setup dynamic}, a PINN takes a finite grid of points in $D \times (0,T]$ for a fixed final time $T>0$ as input and produces an approximation~$u_\theta(x,t)$ to the solution at each grid point $(x,t) \in D\times(0,T]$ by minimizing the residuals of equations \eqref{eq: PDE setup dynamic} in an appropriate function space norm. A typical choice is \( L^2(D;T)\), giving the loss function
\begin{align}
\label{eq:pinns-dynamic}
\mathcal{L}_{\text{pde}}(a, u_\theta) 
& \ = \  \Big\|\frac{du_{\theta}}{dt} - \mathcal{R}(u_{\theta})\Big\|^2_{L^2(T;D)}  \ + \  \alpha  \ \Big\|u_{\theta}|_{\partial D} - g \Big\|^2_{L^2(T; \partial D)} \ + \ \beta  \ \Big\|u_{\theta}|_{t=0} - a \Big\|^2_{L^2(D)} \nonumber \\ \begin{split}
& \ = \  \int_{0}^T \int_D \Big|  \frac{du_{\theta}}{dt}(t,x) -\mathcal{R}(u_{\theta})(t,x)\Big|^2 \mathrm{d}x \mathrm{d}t \\
& \qquad \quad   + \  \alpha \int_{0}^T \int_{\partial D} | u_{\theta}(t,x) - g(t,x)|^2 \mathrm{d}x \mathrm{d}t \ + \  \beta \int_{D} |u_{\theta}(0,x) - a(x)|^2 \mathrm{d}x.
\end{split}
\end{align}

\hfill 

In practice, the losses are evaluated on a discrete grid of points in the domain and summed up (possibly with quadrature weights) to approximate the loss integrals. The result is then used as the loss function for gradient-based optimization to train the model. The optimization could also be formulated via the variational form~\citep{weinan2018deep}. \\

In these loss functions, $\alpha, \beta \in \mathbb R^+$ are constant hyperparameters used to balance the contributions of the different terms. Multi-objective optimization with physics losses can prove challenging with complex loss landscapes~\citep{Krishnapriyan2021}. Gradients of the different losses can point in different directions with significant differences in magnitude, leading to unbalanced contributions and reductions of the different losses during training~\citep{Wang2021}. Manually tuning the loss coefficients can be computationally expensive, especially as the number of loss terms increases, but strategies  have been developed to adaptively update these coefficients and mitigate this issue \citep{Chen2018,Heydari2019,Bischof2021}. In this paper, we either conduct grid searches over the loss coefficients or employ \textsc{ReLoBRaLo} (Relative Loss Balancing with Random Lookback)~\citep{Bischof2021} which uses the history of loss decay across random lookback intervals to achieve uniform progress across the different terms and faster convergence overall. 

\clearpage

\subsubsection{Physics-Informed Neural Operators (PINOs)} \label{sec: PINO}

\paragraph{Neural Operators.}\!\!\!\!  Neural networks have been successful in many machine learning applications but are fundamentally designed to map between finite-dimensional vectors, which can cause them to overfit to a specific discretization. In contrast, solving PDEs often involves working with functions and functional relationships defined over infinite-dimensional function spaces. Neural operators are a principled way to generalize neural networks to learn mappings between functions, and in particular, to learn solution operators of PDEs~\citep{azizzadenesheli2024neural,Berner2025Principles}. They are applicable to functions given at any discretization and generate outputs that can be evaluated and show consistent accuracy across different resolutions. Neural operators also possess a universal approximation property for operators~\citep{kovachki2021universal}. Neural operators have been successfully applied to a wide range of problems~\citep{Kurth2022,gopakumar2023Fourier,Wen2023,zhou2024ai,Ghafourpour2025NOBLE}.

\begin{figure*}[h]
\centering
\includegraphics[width=0.9\textwidth]{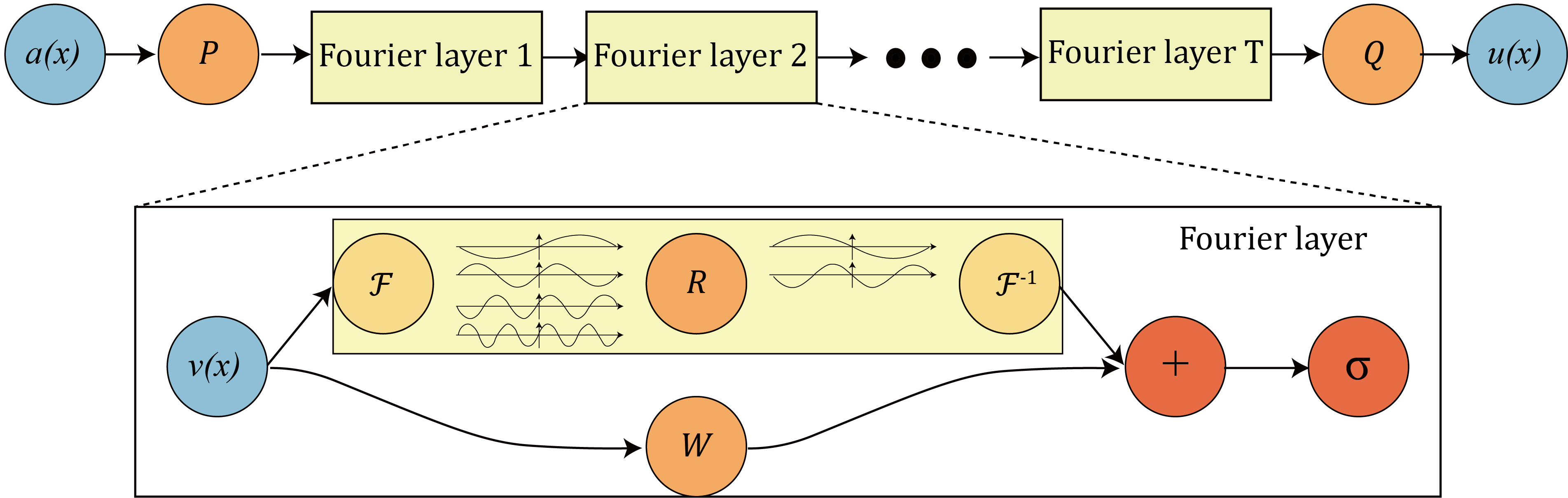} \vspace{-1mm}
\caption{The Fourier Neural Operator (FNO) architecture (extracted from~\citep{FNO_OG}). \label{fig: FNO} }
\end{figure*}

PINO relies on neural operators and typically on the Fourier Neural Operator (FNO)~\citep{FNO_OG,duruisseaux2025FNOGuide} which composes Fourier integral operator layers
with pointwise nonlinear activation functions $\sigma$ to approximate non-linear operators as
\begin{equation} \label{eq: FNO}
   \mathcal{Q} \ \circ \  \sigma (W_{L} + \mathcal{K}_{L} + b_L) \ \circ \  \cdots \ \circ \  \sigma(W_1 + \mathcal{K}_1 + b_1) \ \circ \  \mathcal{P}.
\end{equation}
In this expression, $\mathcal{K}$ denotes Fourier integral kernel operators
\begin{equation}
\label{eq:Fourier}
\bigl(\mathcal{K}(\phi)v_t\bigr)(x)=   
\mathcal{F}^{-1}\Bigl(R_\phi \cdot (\mathcal{F} v_t) \Bigr)(x) 
\end{equation}
where $R_\phi$ is the Fourier transform of a periodic function~$\kappa$ parameterized by \(\phi\). Also, \(\mathcal{P}\) and \(\mathcal{Q} \) are pointwise neural networks that encode the lower dimension function onto a higher dimensional space and project it back to the original space, respectively. In the Fourier integral operator layers, \(W_l \) are matrices, $b_l $ are bias terms, and \(\sigma\) are activation functions. The parameters of the FNO consist of the parameters in $\mathcal{P}, \mathcal{Q}, W_l, \mathcal{K}_l, b_l$. On uniform meshes, the Fourier transform $\mathcal{F}$ can be implemented using the fast Fourier transform. \citep{kossaifi2024neural} maintain a comprehensive open-source library for learning neural operators in PyTorch, which serves as the foundation for our implementation. The FNO architecture is displayed in Figure~\ref{fig: FNO}.

\paragraph{Physics Informed Neural Operators (PINOs).} \!\!\!\!\!\! Physics-Informed Neural Operators (PINOs)~\citep{PINO_OG} are closely related to PINNs, but rely on neural operators instead of neural networks to learn solution operators to entire families of PDEs simultaneously. In the PINO framework~\citep{PINO_OG}, we can first learn a neural operator to approximate the groundtruth operator for a family of PDEs, and then fine-tune the model on any specific PDE instance. We refer to the operator learning phase as \emph{pretraining} and the fine-tuning phase as \emph{instance fine-tuning} or \emph{test-time optimization}. In particular, with PINO, one can train from an unlimited amount of virtual PDE instances by drawing additional coefficient functions or initial conditions. Available data can also be used to supplement the training of the neural operator. A particularly helpful special case is to train a neural operator on low-resolution data instances with high-resolution PDE constraints. In this paper, we focus only on the more challenging setting where data is not available, although the entire exposition naturally applies to cases where data is available.

PINO has advantages over PINNs beyond those of neural operators over neural networks, and has been shown to outperform PINNs on many PDE solving tasks~\citep{rosofsky2022applications}. Optimizing the coefficients in a basis function representation is easier than optimizing a single function as in PINNs. Also, learning these basis functions in the operator learning phase can make the later instance fine-tuning easier. In PINO, we do not need to propagate the information from the initial condition and boundary condition to the interior, it only requires fine-tuning the solution function parameterized by the solution operator.

\subsubsection{Other Physics-informed Approaches} 

Various approaches enforce physics laws in surrogate models without using physics losses. This can be achieved, for instance, by using projection layers~\citep{jiang2020,duruisseaux_towards_2024,Harder2024} or by finding optimal linear combinations of learned basis functions that solve a PDE-constrained optimization problem~\citep{Negiar2022,chalapathi2024}. Another approach is to leverage well-studied inherent properties of the solution operator. For instance, incompressibility $\nabla \cdot u = 0$ can be enforced by representing $u$ as the curl of a vector field $v$, $u = \nabla \times v$, and learning $v$ directly~\citep{Mohan2023}, which guarantees the resulting $u$ is divergence-free. As another example, machine learning surrogates can preserve the symplectic structure underlying Hamiltonian systems~\citep{BurbyHenon,Jin2020,Duruisseaux2023NPMap, DuruisseauxLieFVINs, Valperga2022}, to ensure conservation of energy and other fundamental invariants, which leads to predictions that remain physically realistic over long time horizons. \citet{Cheng2026} use the governing PDE to separate parameter-dependent terms, enabling one operator to generalize across parameter regimes. \citet{yao2026} incorporate partial physics knowledge into pretrained surrogates to improve targeted out-of-distribution counterfactual predictions from factual data alone. 

When the system is well-understood and physical constraints can be explicitly encoded, these methods are particularly effective and can guarantee strict adherence to the governing physics. However, they are often tailored to very specific PDE structures and are limited to well-understood dynamical systems with well-characterized solutions. In contrast, methods that incorporate physics losses, such as PINNs and PINOs, provide greater flexibility and broader applicability, including PDEs whose solution properties are not well-known. By including physics-informed loss terms during training, as in PINNs and PINOs, the governing equations act as regularizers, offering flexibility, applicability across diverse PDEs, and the advantage of requiring only knowledge of the underlying equations. 

Hybrid strategies are also possible, where some constraints can be directly enforced in the architecture while others are imposed via physics losses during training. For example, in incompressible Navier–Stokes problems, a physics loss can penalize deviations from the momentum equation, while incompressibility can be explicitly enforced using projections as demonstrated in~\citep{jiang2020,duruisseaux_towards_2024}.

\subsection{Computing Derivatives for Physics Losses}

Effectively using physics-based losses to train PINNs or PINOs requires computing derivatives both accurately and efficiently. Even small numerical inaccuracies in the computed derivatives can become greatly amplified when they are combined and processed within the physics losses, which can substantially degrade the accuracy and reliability of the predicted solutions. This becomes particularly prohibitive in regimes requiring high-precision solutions.  

\vspace{2mm}

\paragraph{Finite Differences.} \!\!\!\!\!\!
One simple method for approximating derivatives is to use finite differences. This technique estimates derivatives by computing the differences between function values at neighboring points on a discrete grid. Finite differences are both computationally efficient and memory-friendly, requiring only $\mathcal{O}(n)$ operations for a grid with $n$ points. They are straightforward to implement and widely used in numerical simulations. However, finite differences face the same challenges as the corresponding numerical solvers: they can require high-resolution grids to achieve accurate results. If the grid is too coarse, derivative approximations can be poor, leading to significant errors. Consequently, finite differences can become computationally expensive or even intractable for multi-scale systems or for dynamics that vary rapidly in space or time. Note that higher-order finite difference schemes or stencils can sometimes help mitigate these issues.

\paragraph{Automatic Differentiation.}  \!\!\!\!\!\! Derivatives can be evaluated pointwise using automatic differentiation, which applies the chain rule to the sequence of operations in the model. Tools like Autograd~\citep{Autograd} automate this by constructing a computational graph during the forward pass and using reverse-mode differentiation to compute gradients in the backward pass. Automatic differentiation is often preferred over finite differences in PINNs~\citep{Baydin2017} because (1) it yields extremely accurate derivatives even at low resolutions, (2) it requires only a single function evaluation, and (3) it enables efficient computation of higher-order derivatives with minimal overhead. In contrast, finite differences require more evaluations and are more prone to numerical errors.

\vspace{2mm}

However, automatic differentiation has drawbacks: all operations must be differentiable, and storing intermediate values in the computational graph can lead to high memory consumption in deep models. It may also be slower than finite differences for problems where the latter are sufficiently accurate on low-resolution grids. Moreover, deep compositions in the physics loss can exacerbate vanishing or exploding gradient issues during backpropagation.  

\paragraph{Spectral Differentiation.} \!\!\!\!\!\!
Spectral differentiation (or Fourier differentiation) is a fast and memory-efficient method to approximate derivatives by working in frequency space \citep{Trefethen2000,boyd2013chebyshev}. Spectral differentiation is especially appealing when the solution is smooth and periodic, as it delivers very high accuracy, even at relatively low and moderate resolution, and can be implemented efficiently via the fast Fourier transform (FFT). Spectral differentiation exploits the fact that any sufficiently smooth and periodic function 
\(f : [0,2\pi] \to \mathbb{R}\) can be represented by its Fourier series
$$ 
  f(x) \;=\;\sum_{k=-\infty}^{\infty} \widehat f_{k}\,e^{ikx},
  \qquad
  \widehat f_{k} \;=\;\frac{1}{2\pi}\int_{0}^{2\pi}f(x)e^{-ikx}\,\mathrm{d}x.
$$
Taking derivatives in physical space corresponds to simple multiplications by powers of $ik$ in frequency space, where $k$ represents the wavenumber. More precisely, let \(n\) be the number of equidistant grid points. Then, the \(m\)-th derivative of \(f\) has Fourier coefficients
\[
  \mathcal{F}\bigl\{\partial^m f\bigr\}(k)
  = (ik)^m\,\widehat f_{k}, 
\]
After applying an inverse FFT, we recover the \(m\)-th derivative \(\partial^m f(2\pi j / n)\) of \(f\) at the $n$ equidistant grid points. A major advantage of spectral differentiation is that higher derivatives are obtained via multiplication by $(ik)^m$ for the same computational cost irrespective of the derivative order, whereas finite differences require wider stencils and greater computational cost as the derivative order increases. \\ Spectral differentiation is also much more memory‐efficient than automatic differentiation. When applied to high‐dimensional PDEs and deep networks, automatic differentiation  must record the entire computational graph and intermediate values, often leading to prohibitive memory overhead, complex implementation, and slow backward passes when training with physics losses. In contrast, spectral differentiation requires only the storage of the function values and its Fourier coefficients. \\ 

Note that spectral differentiation is highly efficient primarily when the underlying function is noise-free. When the data is noisy, it can become ill-conditioned and lead to a significant loss of accuracy. To mitigate the amplification of errors associated with high-frequency noise, spectral differentiation is often combined with low-pass spectral filtering in noisy settings, which reduces the impact of noise but inevitably degrades the accuracy of the resulting spectral derivatives. In these situations, alternative differentiation techniques, such as high-order finite difference schemes, can sometimes yield more reliable and accurate results. \\

\begin{figure}[t] 
    \centering \vspace{-12mm}
    \includegraphics[width=0.74\linewidth]{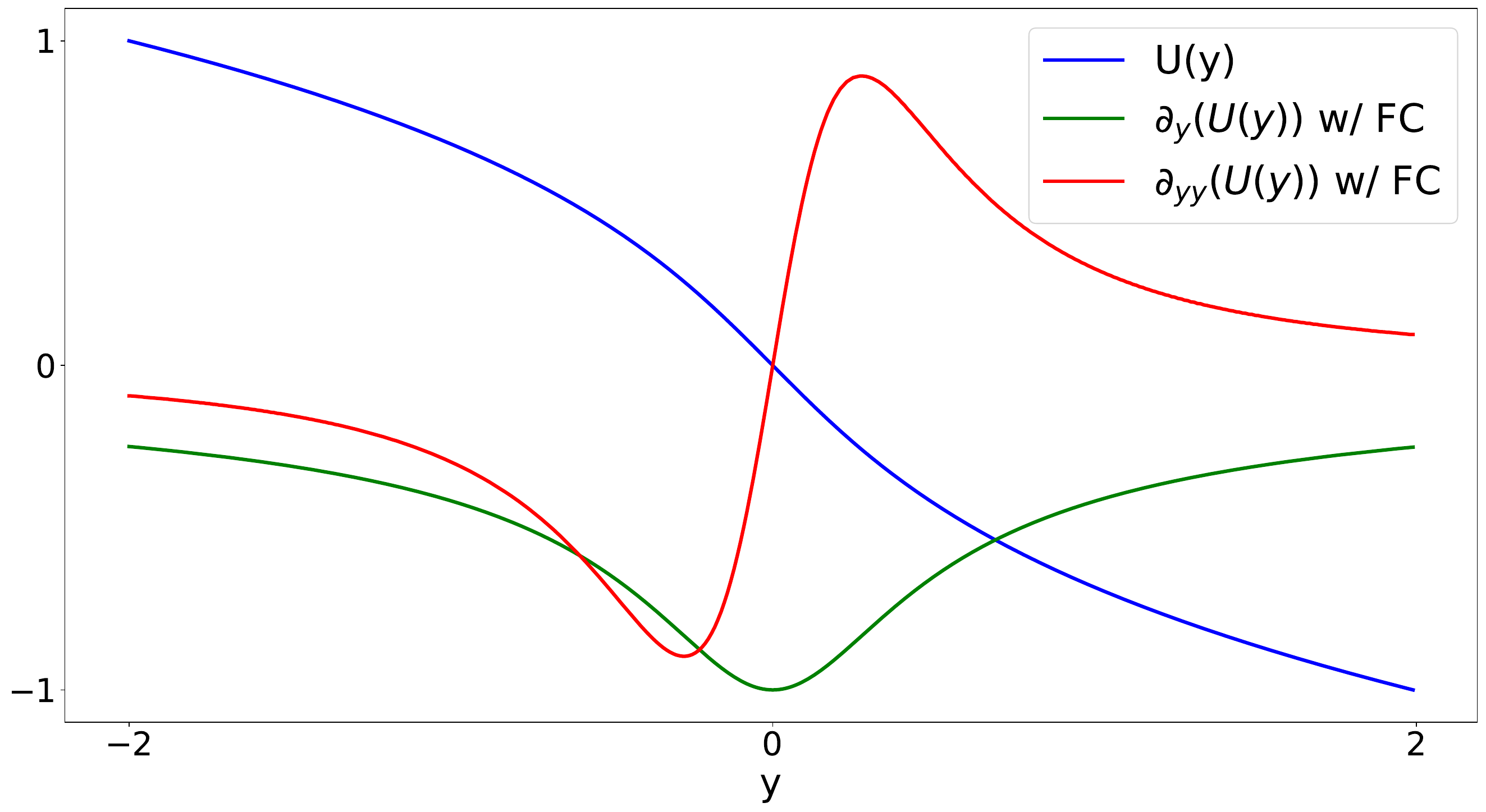}
        \centering
        \includegraphics[width=0.98\linewidth]{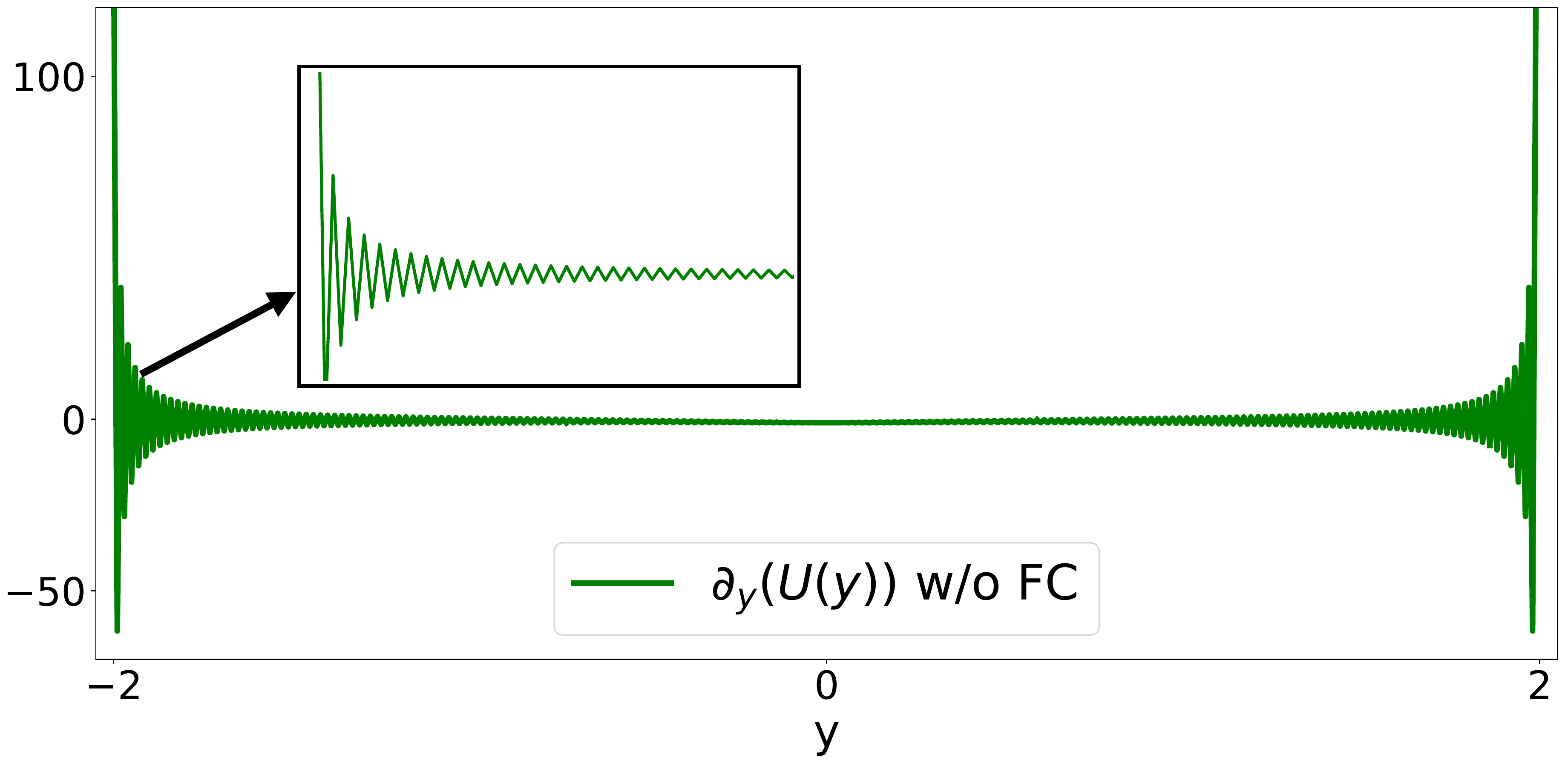}
        \includegraphics[width=0.82\linewidth]{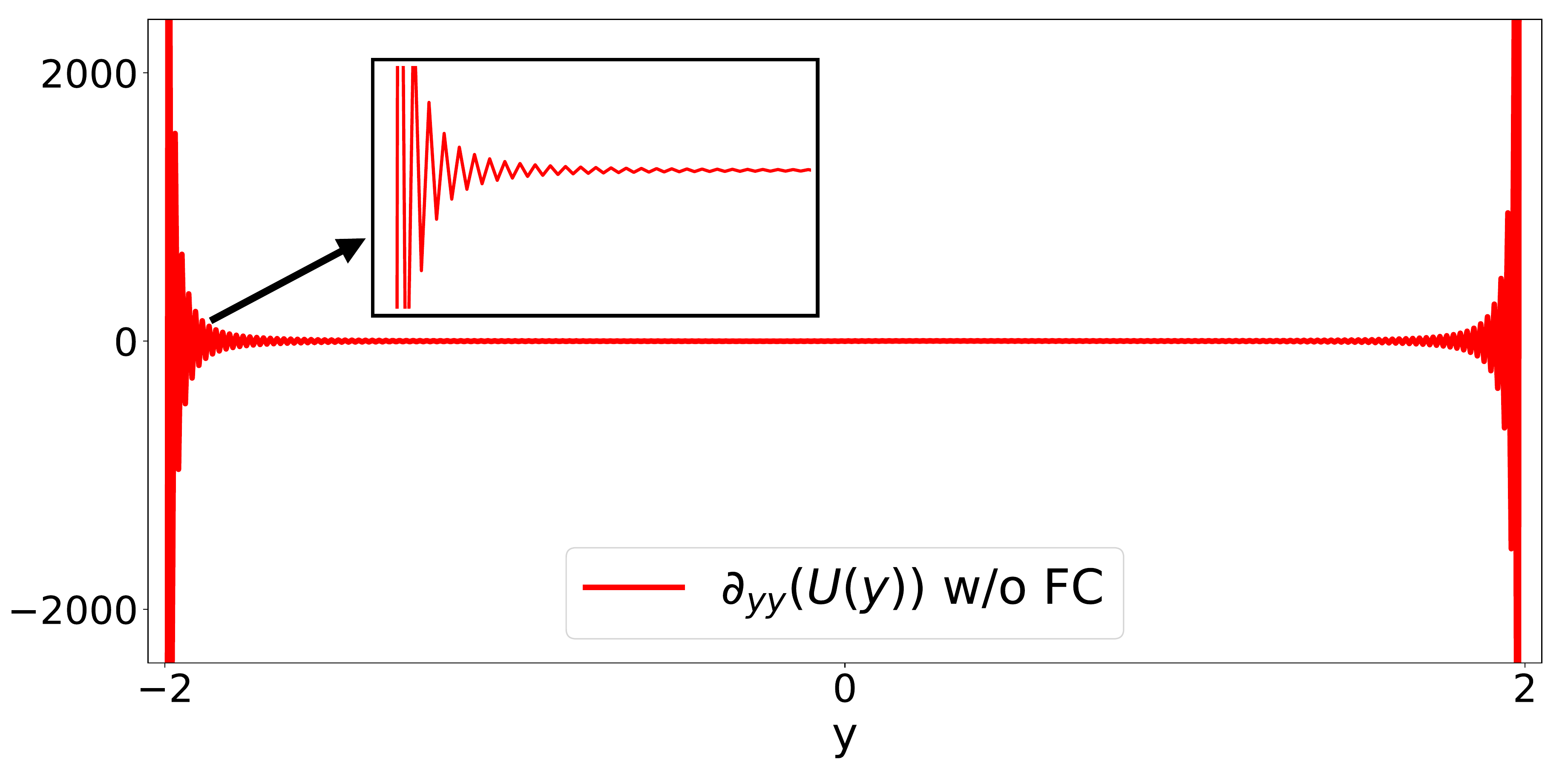} \vspace{-1mm}
    \caption{Visualization of the Gibbs phenomenon when applying spectral differentiation to a non-periodic solution $U(y)$ to the 1D Burgers' equation \eqref{eq: 1D Burger}. 
    (\emph{top}) Applying spectral differentiation on the extended domain after using \textsc{FC--Gram}. 
    (\emph{bottom two}) Applying spectral differentiation on the original domain without Fourier continuation. We observe large spurious oscillations, whose magnitude grows rapidly with the derivative order.}
    \label{fig:gibbs_visualizations}  \vspace{-4mm}
\end{figure}

Unfortunately, like other spectral techniques, spectral differentiation is inherently tailored to periodic functions. If the target function is non-periodic, the resulting Gibbs phenomenon and poor convergence of the Fourier series can degrade derivative accuracy (see \Cref{fig:gibbs_visualizations}), potentially undermining the reliability of physics losses for training models~\citep{Trefethen2000,boyd2013chebyshev}. This highlights the need for modified spectral differentiation methods that can robustly handle non-periodic functions for physics-informed training of machine learning models.

\clearpage

\clearpage

\section{Methods}

A central challenge in spectral differentiation in PINO is that Fourier series are well-defined only for periodic functions~\citep{Trefethen2000}. 
For non-periodic problems, one must first construct a periodic extension over a larger domain, commonly by padding, where the model produces outputs on an extended interval (e.g., \([-1,2]\)), while losses are evaluated solely on the original domain (e.g., \([0,1]\)). This partial observability where the model is optimized using data only from a subinterval becomes particularly problematic in Fourier layers, which rely on the discrete Fourier transform (DFT). The DFT can be written explicitly as a Vandermonde matrix, with entries $F_{jk} = e^{-2\pi i jk / N}$, mapping signals to their frequency coefficients. Then, the condition number of this Vandermonde matrix (i.e., ratio of its largest to smallest singular value) grows rapidly with resolution~\citep{Higham2002,press2007numerical,adcock2014numerical}. As a result, the transform becomes ill-conditioned, and small perturbations in the input can produce disproportionately large deviations in the output. This is exacerbated when the function being transformed is non-periodic or exhibits discontinuous derivatives at the boundaries due to padding, in which case spurious high-frequency components are introduced. Such ill-conditioning not only degrades solution accuracy but also poses a significant challenge for gradient-based optimization in PINO models, underscoring the need for careful boundary treatment and continuation methods.

To overcome this difficulty, it is essential to methods that mitigate ill-conditioning, thereby improving numerical stability and training robustness. One such way is Fourier continuation. We propose \FCPINO{} to incorporate Fourier continuation in the FNO architecture to enable efficient and accurate physics-informed learning with spectral derivatives. We also present two different Fourier continuation methods which can be used within \FCPINO{}.\\

\subsection{\FCPINO} \label{sec: FC-PINO architectures}

We propose \FCPINO{} where Fourier continuation (FC) is incorporated at the start of the FNO architecture \eqref{eq: FNO} to enable an efficient and accurate use of PINO with spectral derivatives. We first recall the formula for PINO with spectral differentiation. \\

\noindent \textbf{\standardPINO{}}. Recall the definition of the FNO from \Cref{sec: PINO},
\begin{equation} \label{eq: FNO 2}
   \mathcal{Q} \ \circ \  \sigma (W_{L} + \mathcal{K}_{L} + b_L) \ \circ \  \cdots \ \circ \  \sigma(W_1 + \mathcal{K}_1 + b_1) \ \circ \  \mathcal{P},
\end{equation}
where $\mathcal{K}$ are Fourier integral kernel operators, \(\mathcal{P}\) and \(\mathcal{Q} \) are pointwise lifting and projection neural networks, $b_l $ are bias terms, and \(\sigma\) are activation functions. In \standardPINO{}, the derivative of the model output is computed using spectral differentiation. 
\begin{figure}[h]
\vspace{1mm}
    \centering \includegraphics[width=0.74\textwidth]{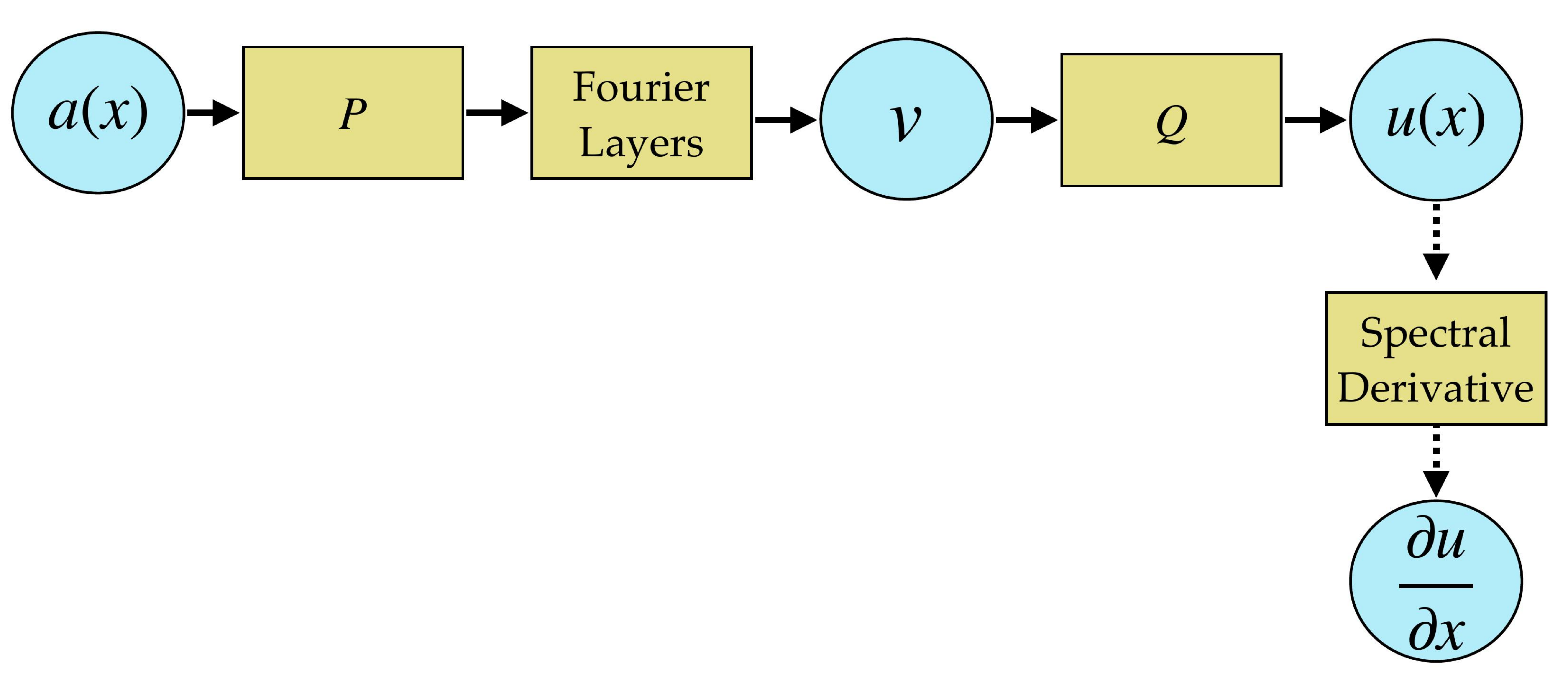}  \vspace{-14mm}
\end{figure}

\clearpage 
\noindent \textbf{\FCPINO{}} has the same structure as the FNO~\eqref{eq: FNO 2}, except it applies FC before the initial lifting operator $\mathcal P$ so each layer acts on a periodic function on an extended domain. The output is then restricted to the original domain before the final projection operator $ \mathcal Q$:
\begin{equation}
\label{eq:model 3}
 \mathcal Q \  \circ \  \textcolor{NavyBlue}{\mathsf{Restrict}} \ \circ  \ \sigma(W_L + \mathcal{K}_L + b_L) \ \circ \  \cdots \ \circ \  \sigma(W_1 + \mathcal{K}_1 + b_1) \ \circ   \ \mathcal P \  \circ  \ \textcolor{NavyBlue}{\mathsf{FC}}.
\end{equation}
Then, one can compute the derivative of the model output using the chain rule  on the derivative of $ \mathcal Q$ and the spectral derivative of the output before $ \mathcal Q$. Due to the FC layer applied earlier, the signal is periodic throughout the model and the output before $ \mathcal Q$ as well, so we can use spectral differentiation to compute its derivatives on the extended domain and then truncate back to the original domain. The derivatives of the projection operator $ \mathcal Q$ are also easy to obtain since it is a composition of linear layers and an activation function. Note that all the Fourier modes are kept for the spectral differentiation to maximize accuracy.     

\begin{figure}[h] 
\vspace{-1mm}\centering\includegraphics[width=0.94\textwidth]{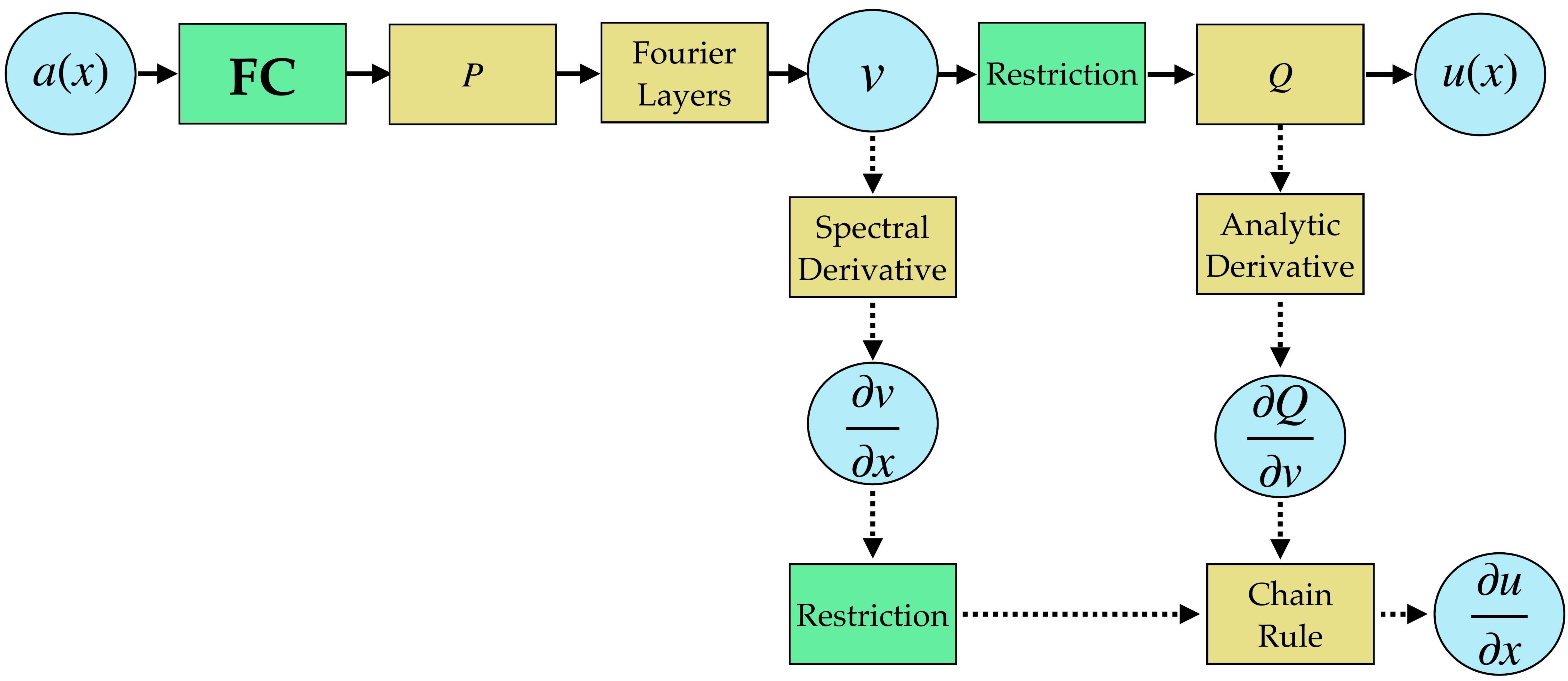}\label{fig:start_FC_diagram}
\end{figure}

\vspace{3mm}

\noindent In our numerical experiments, we consider the following baselines: 
\begin{itemize}
    \item \standardPINO{} where PINO is used with spectral differentiation, without FC,
    \item \padPINO{} which is similar to \FCPINO{} except that the FC layer at the beginning of the model is replaced by zero-padding,
    \item \PINOPadOut{} which is similar to \standardPINO{} except that zero-padding is applied to the model output before the derivative computation. 
\end{itemize} 

\vspace{2mm}

\noindent We also compare \FCPINO{} in \Cref{appx: alternative PINO FC} to two alternative ways of leveraging Fourier continuation when using PINO with spectral derivatives:
\begin{itemize}
    \item \outsideFCPINO{} which is similar to \standardPINO{} except that FC is applied to the model output before the derivative computation,
    \item \EndFCPINO{} where FC is applied inside the model but only before using spectral differentiation. Unlike \FCPINO{} where the Fourier layers act on periodic functions on the extended domain, \EndFCPINO{} retains the standard FNO architecture to produce the model output, and only leverages FC for the derivatives computation.
\end{itemize}

\noindent For all architectures, while the frequency spectrum is truncated in the Fourier layers, all the Fourier modes are kept for the spectral differentiation to maximize accuracy, and we do not introduce any learnable parameters within the spectral differentiation component.  \\

\noindent \textbf{Remark.} When performing spectral differentiation on an extended domain, it is essential to rescale the Fourier modes to ensure the use of correct frequencies. Given a physical domain of size \( L \), the Fourier modes correspond to frequencies $k_l =  2 \pi l /L$. If the domain is extended to size \( \tilde{L} > L \), the corresponding frequencies become $\tilde{k}_l =  2 \pi l /\tilde{L}$. If the differentiation operator is applied in the Fourier domain using the original frequencies~\( k_l \) instead of the rescaled frequencies \( k'_l \), the derivatives will be computed with respect to an incorrect frequency basis. Since \( m\text{th} \)-order spectral differentiation corresponds to multiplication by \( (i k_l)^m \) in Fourier space, using incorrect basis~\( k_l \) results in derivatives scaled by the wrong factor, introducing systematic errors that grow with the differentiation order and the size of the domain extension. Thus, rescaling Fourier modes according to the actual physical domain length is critical for accurate spectral differentiation on the extended domain.   \\

\subsection{Fourier Continuation}\label{FC section}

\subsubsection{Motivation}

Spectral differentiation offers a compelling approach for computing derivatives of smooth, periodic functions. It is significantly more memory-efficient than automatic differentiation, achieves high accuracy even on coarse grids unlike finite differences, and does not incur additional costs when computing higher-order derivatives. However, when applied to non-periodic functions, it suffers from the Gibbs phenomenon (e.g. see \Cref{fig:gibbs_visualizations}), leading to a loss of accuracy, which poses a critical barrier to achieving high-precision results.

To circumvent this issue, one can use spectral differentiation on a larger domain by padding or expanding the original domain. Such extension grants PINOs the necessary expressivity to represent solutions for general non-periodic problems, but the extension  needs to be done carefully. Zero-padding a signal on the sides introduces artificial discontinuities at the endpoints, which result in Gibbs phenomena and spurious oscillations that do not reflect the true behavior of the underlying function, when applying Fourier transforms. An alternative is mirror padding, where a signal is extended by reflecting it about one of its endpoints. While this preserves continuity at the mirror point, it typically introduces a discontinuity in the derivative at that point. Consequently, the extended signal remains continuous but is not differentiable at the mirror point. Since the convergence rate of Fourier series depends on the regularity of the function being approximated, this lack of differentiability leads to very slow spectral convergence and persistent artifacts in the frequency domain representation. 

Ideally, to apply Fourier methods to problems with non-periodic solutions, one must convert the signal of interest to a periodic function in a well-conditioned manner. This is precisely the purpose of Fourier continuation (FC) methods. They extend non-periodic signals to periodic ones over an extended domain through polynomial interpolation, while achieving  empirically superalgebraic convergence rates. We will introduce two Fourier continuation methods that can be integrated into \FCPINO{}: \textsc{FC--Legendre} and (Accelerated) \textsc{FC--Gram}~\citep{AMLANI2016}. \Cref{fig: FC Examples} displays examples of periodic extensions using \textsc{FC--Legendre} and \textsc{FC--Gram} on 1D and 2D non-periodic signals. \\

\subsubsection{General Setting}

\begin{figure}[t]
\centering
\includegraphics[width=0.94\textwidth]{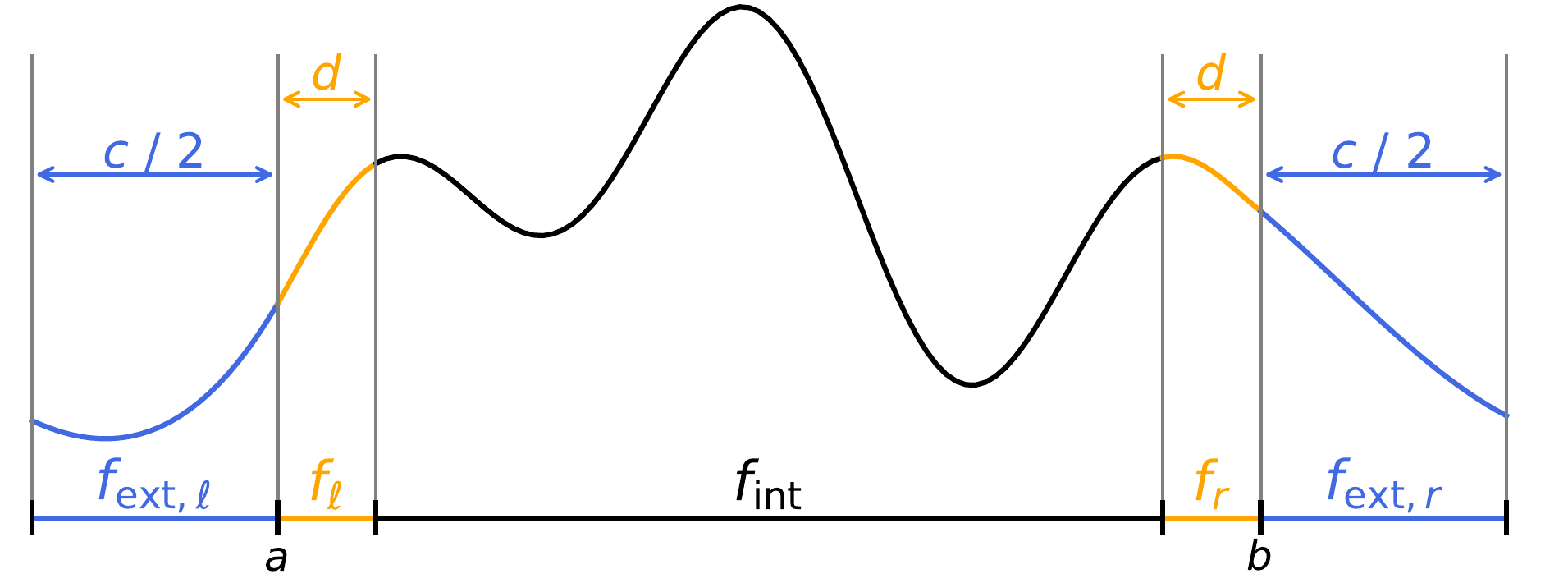} \vspace{-1.5mm}
\caption{Illustration of the various variables used in the Fourier continuation. The given signal \(f\) of $n$ points given on $[a,b]$ is extended to a periodic signal with $(n+c)$ points, and the extension is split and appended on both sides (each with $c/2$ points). \textsc{FC--Legendre} and \textsc{FC--Gram} construct periodic extension using only the information contained in the boundary vectors $f_{\ell}$ and $ f_r$ of width \(d\), and not the remaining part $f_{\text{int}}$ of the signal.   \label{fig: FC Defs} }  \vspace{2mm}
\end{figure}

We first detail the setting for the \textsc{FC--Legendre} and \textsc{FC--Gram} algorithms. \Cref{fig: FC Defs} displays a diagram illustrating the definitions of the variables $c,d, f_{\ell}, f_r, f_{\text{ext}}$ involved in the \textsc{FC--Legendre} and \textsc{FC--Gram} algorithms and defined in what follows.  \\

Let \(f:[a,b]\to\mathbb{R}\) be given on a grid of \(n\) points, with equidistant nodes \(x_k = a + hk\)  for   \(k = 0,\dots,n-1\), where \(h = \frac{b-a}{n-1}\). Denote $f_n = f(x_n)$. The goal is to produce a new sequence~\(\tilde f\) of length \((n+c)\) which is exactly periodic of period \((n+c)\) and agrees with \(f\) on the original interval. Define the left and right boundary vectors of width \(d\),
\begin{align}
f_\ell = (f_0, f_1, \ldots, f_{d - 1})^\top, \quad f_r = (f_{n - d}, \ldots, f_{n-1})^\top.
\end{align}

\textsc{FC--Legendre} and \textsc{FC--Gram} employ distinct strategies for constructing the extension signal $\textcolor{NavyBlue}{f_{\text{ext}}} \in \mathbb{R}^c$, and the periodic extended signal $\tilde f$ is then obtained by concatenating the original signal $f$ and the extension $\textcolor{NavyBlue}{f_{\text{ext}}}$ on both sides, 
\begin{align}
\tilde f & =  (\textcolor{NavyBlue}{f_{\text{ext},c/2},\dots,f_{\text{ext},c-1}},\,f_0,\dots,f_{n-1},\,\textcolor{NavyBlue}{f_{\text{ext},0},\dots,f_{\text{ext},c/2-1}}).
\end{align}

\hfill 

\noindent Note that both \textsc{FC--Legendre} and \textsc{FC--Gram} construct the extension $\textcolor{NavyBlue}{f_{\text{ext}}}$ using only the information contained in the boundary vectors $f_{\ell}$ and $ f_r$ of width \(d\), and not the remaining interior part $f_{\text{int}}$ of the original signal. \\

\noindent For functions in higher dimensions, the one‐dimensional extension is applied along each direction iteratively using the same extension matrices, i.e. first along the first dimension (left/right), then along the second dimension (top/bottom), and so on. \\

\noindent \textbf{Remark.} Instead of concatenating $f$ and $\textcolor{NavyBlue}{f_{\text{ext}}}$ on both sides, one could in principle use a one-sided extension $\tilde f = (f_0,\dots,f_{n-1},\textcolor{NavyBlue}{f_{{\rm ext},0},\dots,f_{{\rm ext},c-1}})$. Both extensions yield the same periodic function, except for a fixed time shift which leaves Fourier transform magnitudes unchanged. However, Gibbs phenomena typically happens near discontinuities, some of which are located at the endpoints when spectral methods are applied to non-periodic signals. The FC approaches typically resolve that issue at the endpoints by extending the signal to a periodic signal on an extended domain, but the two-sided extension is more robust as it keep any residual oscillations at the endpoints outside the original domain.

\subsubsection{\textsc{FC--Legendre}}

In \textsc{FC--Legendre}, the left and right boundary vectors are concatenated into the vector
$y = (f_{r}^\top, f_{\ell}^\top)^\top \in \mathbb{R}^{2d}$, and the extension $\textcolor{NavyBlue}{f_{\text{ext}}} \in \mathbb{R}^c$ is obtained via $ E y = \textcolor{NavyBlue}{f_{\text{ext}}} \in \mathbb{R}^c$. Here, $E \in \mathbb{R}^{c \times 2d}$ is an extension matrix, built from evaluations of shifted Legendre polynomials of degree \((2d-1)\), to do polynomial interpolation in between the right and left boundaries. For fixed parameters $(c, d)$, the \textsc{FC--Legendre} extension matrix $E \in \mathbb{R}^{c \times 2d}$ is independent of the input signal of length $n > 2d$ (by construction). As a result, the extension matrix $E$ can be precomputed offline and reused in \textsc{FC--PINO} without the need for recomputation. Details about the construction of the \textsc{FC--Legendre} extension matrix $E$ are provided in~\Cref{appx: Construction Legendre matrix}.

While only minimal tuning is typically required to obtain good solutions, more careful tuning might be required in high-precision settings. In these settings, tuning \textsc{FC--Legendre} effectively requires a careful balance between extension length, polynomial degree, and noise sensitivity to achieve accurate derivative approximations without compromising numerical stability. The extension length $c$ must be sufficient to capture boundary behavior and ensure smooth continuation, but overly long extensions can amplify noise and induce ill-conditioning. Likewise, higher-degree polynomials improve accuracy for smooth signals by reducing derivative errors, but excessively high degrees can worsen conditioning and introduce oscillatory artifacts. Techniques such as low-pass filtering of boundary data or adjusting the numerical rank in the pseudoinverse can help mitigate noise amplification. Naturally, increasing the signal resolution can also help further reduce PDE residuals.

With systematic tuning and validation, \textsc{FC--Legendre} delivers high-accuracy continuation and spectral differentiation, yielding high-precision derivatives even in challenging scenarios. More details about tuning \textsc{FC--Legendre} for high-precision applications are provided in \Cref{appx: tuning FC Legendre}. 

\vspace{2mm}

\subsubsection{\textsc{FC--Gram}}

We now describe the (Accelerated) \textsc{FC--Gram} method from \citep{AMLANI2016} which is inspired from earlier \textsc{FC--Gram} methods~\citep{Accurate_FC_2007,Fourier_continuation_2011}. See \citep{AMLANI2016} for more details and illustrations of \textsc{FC--Gram}.

In \textsc{FC--Gram}, the extension values $\textcolor{NavyBlue}{f_{\text{ext}}}$ are formed by projecting the left and right boundary vectors $f_\ell, f_r$ onto Gram bases via matrices \(Q_\ell\), \(Q_r\), and then blending their respective continuations using the continuation matrices \(A_\ell\), \(A_r\):
\begin{align}
\textcolor{NavyBlue}{f_{\text{ext}}} = A_\ell Q_\ell^\top f_\ell + A_r Q_r^\top f_r.
\end{align} 
For fixed parameters $(c, d)$, the matrices $A_{\ell}, A_r \in \mathbb{R}^{c \times d}$ and $Q_{\ell}, Q_r \in \mathbb{R}^{d \times d}$ are independent of the input signal of length $n > 2d$ (by construction). As a result, these matrices can be precomputed offline and reused in \textsc{FC--PINO} without the need for recomputation. Details about the construction of the \textsc{FC--Gram} extension matrices $A_{\ell}, A_r , Q_{\ell}, Q_r$ are provided in~\Cref{appx: Construction Gram matrix}.

Tuning \textsc{FC--Gram} is more involved, owing to its localized structure and the multiple construction stages, each with several parameters that can be adjusted. However, after extensive tuning conducted in previous \textsc{FC--Gram} works~\citep{Accurate_FC_2007,Fourier_continuation_2011,AMLANI2016}, fixed values of most parameters have been determined to provide an appropriate trade-off between overall computational cost and stability which works well in most settings. Then, we can simply focus on tuning the polynomial degree $d$, and refining the resolution as necessary since increasing the signal resolution can help further reduce PDE residuals. Higher $d$ allows to better resolve finer boundary features and higher-order derivatives, while excessively high values can cause ill-conditioning, noise overfitting, or oscillatory artifacts.  

Overall, \textsc{FC--Gram} combines flexibility and robustness: its localized blending approach produces smooth, stable derivative fields, and even with minimal tuning, it consistently achieves leads to high-precision solutions when paired with \FCPINO{} across a wide range of problems. More details about tuning \textsc{FC--Gram} for high-precision applications are provided in \Cref{appx: Tuninf FC Gram}.  \\

\subsection{Comparison of \textsc{FC--Legendre} and \textsc{FC--Gram}} \label{FC Comparison}

\textsc{FC--Gram} and \textsc{FC--Legendre} employ fundamentally different continuation strategies, each with unique pros and cons. As illustrated in \Cref{fig: FC Examples}, \textsc{FC--Legendre} constructs the extension by fitting a single Legendre polynomial of degree \( (2d -1) \) spanning the entire extended interval between the left and right boundaries. In contrast, \textsc{FC--Gram} takes a more local approach, projecting boundary data onto lower-order polynomial bases of degree \(d\) separately on each boundary and then smoothly blending the extensions outside the domain. While both approaches yield good results with minimal tuning alongside \FCPINO{}, one can offer advantages over the other in high-precision settings. \\

\textsc{FC--Legendre} is particularly well-suited for problems with very smooth, noise-free signals and when higher-order derivatives are required. Its global continuation using Legendre polynomials of degree \( (2d-1) \) ensures rapid decay of the continuation error, which remains dominant even after amplification by \(m\)-th order spectral derivatives involving factors \((ik)^m\). This global polynomial fit provides uniform approximation of all spectral modes without artificial damping, enabling high-precision derivative computation in the interior domain. Although \textsc{FC--Legendre} benefits from a simple formulation based on analytic polynomials and a single pseudoinverse which makes implementation and tuning more accessible, careful parameter choice is essential to prevent ill-conditioning and oscillatory artifacts that may emerge as the polynomial degree increases. \\

In contrast, \textsc{FC--Gram} excels for signals that are less smooth, noisy, or contain discontinuities, as it provides greater robustness and effective localized error control. By constructing localized orthonormal polynomial bases on each boundary and smoothly blending the extension, it confines boundary errors and high-frequency noise within the extension region, preventing their propagation into the main domain. This approach suppresses spurious oscillations near boundaries, resulting in cleaner derivatives without the need for manual filtering. Although \textsc{FC--Gram} is more computationally intensive and requires higher precision arithmetic, its QR factorizations performed on smaller local matrices produce better-conditioned continuation operators, which improve numerical stability and reduce sensitivity to round-off errors. Overall, its enhanced robustness and localized error confinement makes it especially well-suited for noisy or non-smooth problems.

Together, these two Fourier continuation methods offer complementary tools for spectral continuation, enabling users to select the approach best matched to their problem’s smoothness, noise level, derivative order, and computational constraints.

\subsection{Error Convergence for Solutions and Derivatives in \FCPINO{}}

Let $u_\theta$ denote the output of the PINO model, $\widetilde u_\theta$ its FC extension, and $u$ the true solution. In \FCPINO{}, spectral differentiation is performed on the continuation $\widetilde u_\theta$, so the error made on the derivatives can be decomposed as
\[
\|\partial^k \widetilde u_\theta - \partial^k u \|
\ \  \le \ \ 
 \underbrace{\ \| \partial^k \widetilde u_\theta - \partial^k u_\theta \| \ }_{\text{FC error}}
\ \  + \  \ \underbrace{\ \|\partial^k u_\theta - \partial^k u \| \ }_{\text{PINO error}} .
\]

For \textsc{FC Gram}, the continuation error on the physical interval consists of two parts, as demonstrated in~\citep{LyonThesis,lyon2010high}. The dominant term is the interpolation error from matching $d$ boundary values, which converges at order $\mathcal{O}(N^{-d})$. In addition, there is a blending term caused by forcing the continuation to vanish smoothly outside the interval. This blending error does not formally converge with $N$, but can be made numerically negligible. When computing a derivative of order $k$, the interpolation error is reduced to order $\mathcal{O}(N^{-(d-k)})$, while the blending error remains bounded and does not affect the overall convergence rate. 

For \textsc{FC--Legendre}, the extension matrix maps the boundary vector $y=(f_r^\top,f_\ell^\top)^\top\in\mathbb{R}^{2d}$ to the polynomial of degree less than $2d$ fitting these $2d$ values. The approximation order on the physical interval is limited by the boundary strip width $d$, since each side provides at most $d$ independent pieces of information, yielding an interpolation error $\mathcal{O}(N^{-d})$, with $k$ derivatives reducing it to $\mathcal{O}(N^{-(d-k)})$. The high polynomial degree ensures a smooth continuation, while SVD or pseudoinverse residuals are typically negligible.

Now, \FCPINO{} is based on the Fourier neural operator, which enjoys universal approximation for sufficiently smooth operator mappings \citep{Kovachki2021_2}, i.e., one can make the approximation error $\|u_\theta - \mathcal{G}(f)\|$ arbitrarily small. By the Sobolev embedding theorem, with sufficient regularity, the error in the derivatives can also be controlled and kept arbitrarily small. Combining this with the FC error estimates yields a $\mathcal{O}(N^{-(d-k)})$ rate for \FCPINO{}. Thus, \FCPINO{} inherits the algebraic convergence rate of FC-based spectral differentiation together with the universal approximation guarantees of PINO.

\subsection{Computational Complexity}

\subsubsection{Spectral Differentiation}

Given a one-dimensional signal of length \( n \), spectral differentiation first applies the fast Fourier transform (FFT) to compute the spectral coefficients, at a computational cost of \( \mathcal{O}(n \log n) \). The derivatives in Fourier space are then obtained by multiplying each coefficient by \( ik \), which require \( \mathcal{O}(n) \) operations. Note here that all the Fourier modes $k$ are kept in the FFT and used in the derivative computation to maximize accuracy. Finally, the inverse FFT is used to transform the result back to physical space, with an additional cost of \( \mathcal{O}(n \log n) \). Thus, the overall computational complexity of spectral differentiation is \( \mathcal{O}(n \log n) \).

When computing a spectral derivative along a single dimension for a signal on an \( m \)-dimensional grid with \( n^m \) points, one applies the one-dimensional FFT along that axis for each fixed combination of the other \( (m-1) \) indices. This results in \( n^{m-1} \) FFTs, each with cost \( \mathcal{O}(n \log n) \). Then, the derivatives are obtained in Fourier space, requiring \( n^{m-1} \) operations each with cost \( \mathcal{O}(n) \). Finally, the result is transformed back to physical space using \( n^{m-1} \) inverse FFTs, each with cost \( \mathcal{O}(n \log n) \). Thus, the total computational complexity for spectral differentiation of a \( m \)-dimensional signal along one dimension is
$\mathcal{O}(n^m \log n)$.

\subsubsection{Fourier Continuation} 

The computational cost of extending a signal from $n$ points to $(n+c)$ points using \textsc{FC--Legendre} based on boundary vectors of width \(d\) is the cost $\mathcal{O}(c\cdot d)$ associated to the matrix-vector multiplication of $ y = (f_{r}^\top, f_{\ell}^\top)^\top \in \mathbb{R}^{2d}$ by the precomputed Legendre extension matrix $E \in \mathbb{R}^{c \times 2d}$. Similarly, the computational cost of extending a signal of length $n$ into a signal of length $(n+c)$ using \textsc{FC--Gram} based on boundary vectors of width \(d\) is the cost $\mathcal{O}(c\cdot d)$ associated to the two matrix-vector multiplications of $ f_\ell, f_r \in \mathbb{R}^{d}$ by the precomputed Gram continuation matrices $\tilde{A}_\ell = A_\ell Q_\ell^\top \in \mathbb{R}^{c \times d}$ and $\tilde{A}_r = A_r Q_r^\top \in \mathbb{R}^{c \times d}$.

When applying Fourier continuation along a single dimension for a signal on an \( m \)-dimensional grid with \( n^m \) grid points, one applies the extension matrices along that axis for each fixed combination of the remaining \( (m-1) \) indices. This results in \( n^{m-1} \) Fourier continuations, each with a cost of \( \mathcal{O}(c \cdot d) \). Consequently, the total computational complexity for Fourier continuation of a \( m \)-dimensional signal along one dimension is
$\mathcal{O}(c \cdot d \cdot n^{m-1})$.

When applying Fourier continuation one dimension at a time to an \( m \)-dimensional signal of size \( n^m \), each successive extension operates on increasingly larger signals. Specifically, the cost of extending along the \( k \)-th dimension involves applying a matrix-vector multiplication of cost \( \mathcal{O}(c \cdot d) \) to \( (n + c)^{k-1} n^{m-k} \) slices. The total computational complexity is $\mathcal{O} \left( c \cdot d \cdot \sum_{k=0}^{m-1} (n + c)^k n^{m-1-k} \right)$ which is $\mathcal{O} \left( c \cdot d \cdot (n+c)^{m-1}   \right)$ since $c < n$ typically.

\subsubsection{Spectral Differentiation with Fourier Continuation} 

Given a signal on a \( m \)-dimensional grid with \( n^m \) grid points, Fourier continuation along one dimension costs $\mathcal{O}(c \cdot d \cdot n^{m-1}  )$. Based on the earlier result, computing the spectral partial derivative along that dimension for the extended signal with $n^{m-1} (n+c)$ points has a cost of 
$\mathcal{O}(n^{m-1} (n+c) \log (n+c))$. Since $c\cdot d \ll (n+c) \log (n+c)$, the overall cost of spectral differentiation with Fourier continuation is $\mathcal{O}(n^{m-1} (n+c) \log (n+c))$.

\subsubsection{FC-PINO}

Given a signal on a \( m \)-dimensional grid with \( n^m \) grid points, the computational cost of a Fourier layer is either dominated by the cost of the Fourier integral kernel integrator $\mathcal{K}_\ell$, or the cost of the pointwise linear operator $W_\ell + b_\ell$. Let $L$ denote the number of Fourier layers, and $\mathfrak{C}_{in},\mathfrak{C}_{out}$ denote the number of input and output channels of the layer, respectively.  

 For the \standardPINO{} approach, the computational complexity of an FNO inference is $\mathcal{O}\left(L \cdot \left(  \mathfrak{C}_{in} \cdot n^{m} \log n  + \mathfrak{C}_{in}  \cdot \mathfrak{C}_{out} \cdot n^m   \right) \right)$ or more concisely $\mathcal{O}\left(L \cdot \mathfrak{C}_{in} \cdot n^m \cdot \left(  \mathfrak{C}_{out}  +  \log n    \right) \right)$, and the cost of spectral differentiation is $\mathcal{O}(n^m \log n)$.

For \FCPINO{}, the computational cost of inference is slightly higher because the FNO layers are applied on the extended domain with $(n+c)$ points, leading to a computational complexity of $\mathcal{O}\left(L \cdot \mathfrak{C}_{in} \cdot (n+c)^m \cdot \left(  \mathfrak{C}_{out}  +  \log (n+c)    \right) \right)$. Then, the computational cost of spectral differentiation along one dimension of \FCPINO{} is the cost of spectral differentiation on the extended domain with $(n+c)$ points, that is $\mathcal{O}((n+c)^m \log (n+c))$. 

\hfill

\section{Numerical Experiments} \label{experiments section}

\vspace{2mm}

We now consider a suite of PDE benchmarks to test the proposed \FCPINO{} approach. Specifically, we study the 1D self-similar (inviscid) Burgers equation in \Cref{sec: 1D Self-Similar Burgers' Equation}, both in the single-instance setting and in the operator-learning setting over a family of parameters, the 2D viscous Burgers equation in \Cref{sec: 2D Burgers}, and the 3D incompressible Navier--Stokes equations for lid-driven cavity flow in velocity--pressure form in \Cref{sec: NS}. For each problem, we minimize a physics-informed loss, given by a weighted combination of an interior PDE residual and the appropriate constraint terms. \\ 

Derivatives appearing in the physics loss are computed via spectral differentiation in different ways, as discussed in \Cref{sec: FC-PINO architectures}. To ensure that the reported interior residual losses are comparable and not an artifact of a particular spectral differentiation pipeline, we also validate all of these results using automatic differentiation. \\

Note that obtaining correct PDE residuals using automatic differentiation can be intricate in practice. Since the FNO outputs a discrete field $u$ on a regular grid rather than a pointwise function, automatic differentiation cannot directly provide derivatives such as $\nabla_x u$ at arbitrary coordinates. To enable this verification, we need a differentiable query map $x \mapsto u(x)$ that can be evaluated at continuous points, so that $\nabla_x u(x)$ can be obtained by backpropagating through some (or all) of the layers. This can be achieved for instance by replacing the last inverse fast Fourier transform by an inverse discrete Fourier transform, and making sure the resulting computational graphs allow gradients to flow correctly from the query position $x$ to the queried solution $u(x)$ at $x$. Consequently, residual evaluation via automatic differentiation is substantially more memory intensive and computationally expensive than spectral differentiation in this setting, but it provides an important verification of the reported interior residual losses

\subsection{1D Self-Similar Burgers' Equation} \label{sec: 1D Self-Similar Burgers' Equation}

\vspace{2mm}

\subsubsection{Problem Setting}

We first consider the non-viscous Burgers' equation
\begin{equation}
    u_t + uu_x = 0 
\end{equation}
where $u(x,t)$ is the solution function. From there, if we make the substitution
\begin{equation}
    u = (1-t)^{\lambda} \ U \! \left(\frac{x}{(1-t)^{1+\lambda}}\right)
\end{equation}
for some fixed $\lambda \in \mathbb R^+$, the equation simplifies to
\begin{equation}\label{eq: 1D Burger}-\lambda U + \left((1+\lambda)y+U\right)U_y=0\end{equation}
where $U$ is the solution function and $U_y$ is the partial derivative of $U$ with respect to $y$. Equation ($\ref{eq: 1D Burger}$) is known as the self-similar Burgers' equation. In this form, there is only one input variable $y$, so this is a 1D problem. \\

We will train the models using the PDE residual from equation $\eqref{eq: 1D Burger}$ and the boundary condition $U(-2)=1$. Motivated by the results in~\citep{PINN_3rd_deriv}, we introduce an additional smoothness loss term which is the mean squared error of the derivative with respect to $y$ of the left side of equation~($\ref{eq: 1D Burger}$), to improve the smoothness of the solution. Hence the total loss is given by \begin{equation}
\label{eq: 1D Burgers Loss}
\begin{split}
 \mathcal{L}_{\text{pde}}(U_\theta) \ &= \  \int_D \big|  \mathcal{P}(U_\theta(y), y; \lambda) \big|^2\mathrm{d}y  \ + \   \alpha   | U_\theta(-2) - 1 |^2    \  + \    \beta \! \int_D \Big|\partial_y \mathcal{P}(U_\theta(y), y; \lambda)\Big|^2\mathrm dy \end{split}
\end{equation}
where \begin{equation} \mathcal{P}(U_\theta(y), y; \lambda) = \left[(1+\lambda)y+U_\theta(y)\right] \ \! \partial_y U_\theta(y)  \ - \
  \lambda U_\theta(y). \end{equation}

\hfill  

\subsubsection{Comparison of PINO Strategies using Spectral Differentiation}
\label{sec: Burgers 1D Comparison}

We first compare the performance of \FCPINO{} to the baseline models for solving a single instance of the self-similar Burgers' equation. For each architecture, we performed a hyperparameter grid search, varying the number of Fourier modes, learning rate, scheduler patience, and the $(c,d)$ continuation parameters. The optimal hyperparameters for each architecture are provided in \Cref{appx:hyper_1d_burgers}, while \Cref{tab:burgers1D_start_baseline} reports the errors achieved by the best model for each architecture. We see that \FCPINO{} substantially outperforms all the baselines by multiple orders of magnitude. 

\begin{table}[h] \vspace{2mm}
    \caption{PDE residual, boundary condition (BC) loss, smoothness loss, and total loss for \FCPINO{} with \textsc{FC--Gram} and \textsc{FC--Legendre}, and baseline PINO architectures when solving the self-similar Burgers' equation~\eqref{eq: 1D Burger} for $\lambda = 0.5$.}
    \label{tab:burgers1D_start_baseline} \vspace{-1.2mm}
    \centering
    \resizebox{\textwidth}{!}{%
    \begin{tabular}{lcccc}
        \toprule
        \textbf{Model} & \textbf{PDE Residual} & \textbf{BC Loss} & \textbf{Smoothness} & \textbf{Total Loss} \\
        \midrule
        \textbf{\textsc{FC--Gram}} \\
        \hspace{3mm} \FCPINO{}   & $1.9 \times 10^{-11}$ & \bm{$9.2 \times 10^{-15}$} & \bm{$5.5 \times 10^{-9}$} & \bm{$6.7 \times 10^{-10}$} \\
        \midrule
        \textbf{\textsc{FC--Legendre}} \\
        \hspace{3mm} \FCPINO{}   & \bm{$1.8 \times 10^{-12}$} & $1.9 \times 10^{-14}$ & $2.8 \times 10^{-8}$ & $2.8 \times 10^{-9}$ \ \\
        \midrule
        \textbf{\textsc{Baselines}} \\
        \hspace{3mm} \padPINO{}    & $1.4 \times 10^{-8}$ \   & $1.1 \times 10^{-10}$ & $4.9 \times 10^{-5}$ & $4.9 \times 10^{-6}$ \  \\
        \hspace{3mm} \PINOPadOut{} & $2.0 \times 10^{0}$ \  \ \   & $2.0 \times 10^{-6}$ \  & $9.1 \times 10^{-5}$ & $2.0 \times 10^{0}$ \ \ \ \\
        \hspace{3mm} \standardPINO{}     & $2.0 \times 10^{0}$ \ \ \     & $2.0 \times 10^{-6}$ \  & $2.0 \times 10^{0}$ \ \   & $2.0 \times 10^{0}$ \ \ \  \\
        \bottomrule
    \end{tabular}%
    } \vspace{8mm}
\end{table}

Examples of solutions obtained are displayed in \Cref{fig:1d_burgers_1}. We can see that \FCPINO{} produces a correct solution while maintaining smooth and accurate derivatives. In contrast, \padPINO{} generates the correct model output, but its derivatives exhibit oscillations and lack smoothness, whereas \PINOPadOut{} and \standardPINO{} converge to incorrect solutions, resulting in highly erratic derivatives. \\

Regarding computational time to train the different PINO models, \Cref{fig:threshold_crossing} shows that for this specific one-dimensional example, where $n=400$, that \FCPINO{} reaches a loss of $10^{-5}$ very rapidly and much faster than \padPINO{}, and then more slowly converges towards $10^{-8}$ or $10^{-9}$. Note that \FCPINO{} also reaches a threshold of $10^{-8}$ much before \padPINO{} approaches $10^{-6}$. \\

\subsubsection{Learning the Family of Solutions} \label{sec: learn family Burgers 1D}

\vspace{1mm}

In the PINO framework, we can also train a neural operator to approximate solutions for entire families of PDEs, which can then be fine-tuned on any specific PDE instance. This capability naturally extends to \FCPINO{}, and we use this framework to learn the solution operator of the self-similar Burgers' equation~\eqref{eq: 1D Burger} over many values of $\lambda$.  

It is known that for the Burgers equation to have a smooth self-similar solution, $\lambda$ has to take the form $\lambda = 1/(2i + 2)$ \citep{PINN_3rd_deriv}. Thus we obtain a collection of values of $\lambda = 1/(2i + 2)$ by randomly sampling $i \in [0,  20]$. By training the neural operator on the corresponding Burgers' equations, we effectively learn the solution operator over arbitrary values of $\lambda$ corresponding to smooth self-similar solutions.

\begin{table}[t]
    \centering
     \caption{PDE residual, boundary condition (BC) loss, and smoothness loss when solving the self-similar Burgers' equation with \FCPINO{} over a family of values of $\lambda$. Here, we report the errors obtained when fine-tuning the pretrained model on the individual values $\lambda \in \{ 1/2, 1/12, 1/22, 1/32, 1/52\}$, and the average over these 5 instances.} \vspace{-1mm}
    \label{tab:family_burgers_losses}
    \begin{tabular}{lccc}
        \toprule
        \textbf{Model (\textsc{FC--Gram)}} & \hspace{3mm}\textbf{PDE Residual}\hspace{3mm} & \hspace{3mm} \textbf{BC Loss}\hspace{3mm} & \hspace{4mm}\textbf{Smoothness} \hspace{4mm} \\
        \midrule

        \textbf{\boldmath$\lambda = 1/2$}   & $8.8 \times 10^{-10}$ & $3.7 \times 10^{-12}$ & $5.9 \times 10^{-8}$ \\
        \midrule
        \textbf{\boldmath$\lambda = 1/12$}  & $2.3 \times 10^{-10}$ & $6.7 \times 10^{-13}$ & $6.1 \times 10^{-8}$ \\
        \midrule
        \textbf{\boldmath$\lambda = 1/22$}  & $1.5 \times 10^{-10}$ & $8.3 \times 10^{-11}$ & $1.3 \times 10^{-7}$ \\
        \midrule
        \textbf{\boldmath$\lambda = 1/32$ }  & $8.3 \times 10^{-10}$ & $1.8 \times 10^{-10}$ & $1.6 \times 10^{-7}$ \\
        \midrule
        \textbf{\boldmath$\lambda = 1/52$}  & $2.2 \times 10^{-9}$ \  & $8.6 \times 10^{-10}$ & $3.4 \times 10^{-7}$ \\
        \midrule
        \textbf{Fine-tuning Average}    \ \ \                 & $8.6 \times 10^{-10}$ & $2.3 \times 10^{-10}$ & $1.5 \times 10^{-7}$ \\
        \bottomrule
    \end{tabular} \vspace{2mm}
\end{table}

\begin{figure}[t] 
\vspace{-0mm}
    \centering
    \begin{minipage}[t]{0.82\textwidth}
        \centering
        \includegraphics[width=\linewidth]{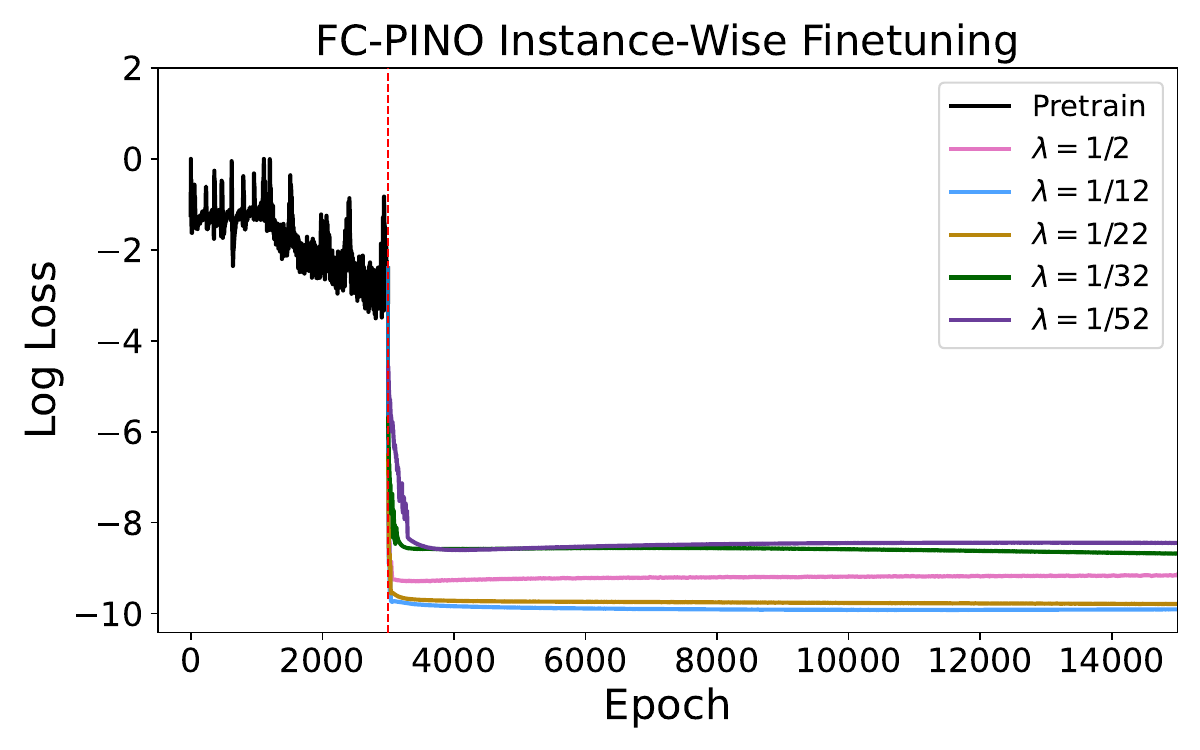}
    \end{minipage} \vspace{-3mm}
    \caption{PDE residual loss across different $\lambda$'s when fine-tuning the pretrained \FCPINO{}. The black curve represents the pretrain phase where we use the Adam optimizer for 3000 epochs, learning a solution operator over multiple $\lambda$ values. We subsequently switch to the L--BFGS optimizer, as signified by the dashed line, and fine-tune the operator individually on 5 different PDE instances corresponding to different values of $\lambda$. } 
    \label{fig:main_tex_family}  \vspace{6mm}
\end{figure}

We first pretrain the proposed \FCPINO{} architecture with \textsc{FC--Gram} for $3000$ epochs across the selected subset of values of $\lambda$. Then, we fine-tune the pretrained model on individual values of $\lambda$, for $\lambda \in \{ 1/2, 1/12, 1/22, 1/32, 1/52\}$, using the second-order L--BFGS optimizer for $40,\!000$ epochs. We embed $\lambda$ in the architecture by representing it as the amplitude of sines and cosines in a NeRF-style embedding~\citep{mildenhall2021nerf}, as $$\left[ \lambda, \ \lambda\sin(2^0  x), \  \lambda \cos(2^0  x) , \ \ldots, \  \lambda \sin(2^{K}  x), \ \lambda \cos(2^{K}  x) \right]$$ for some integer $K>0$, where $x$ denotes the discretized spatial coordinates. 

FNOs naturally leverage sines and cosines through the Fourier transform to learn mappings by capturing interactions between frequency components. Using an encoding with a richer frequency spectrum than a constant $\lambda$ function can allow more efficient learning of complex spectral dynamics. In this numerical experiment, $K=2$ proved sufficient.

 \Cref{tab:family_burgers_losses} reports the final component losses for the selected $\lambda$ values. Notably, \FCPINO{} achieves lower losses for the larger values of $\lambda$. \Cref{fig:main_tex_family} illustrates the convergence behavior of \FCPINO{}. When fine-tuning pretrained models on individual instances using the L--BFGS optimizer, the PDE residual initially drops rapidly to a value close to its final convergence point, followed by a much slower and more gradual reduction as it fully converges. We also display examples of model outputs in \Cref{appx:fine_tuning_sols}, and observe very small oscillations in the second derivative of the solutions for smaller values of $\lambda$.

 \hfill

\subsection{2D Burgers' Equation} \label{sec: 2D Burgers}

\subsubsection{Problem Setting}

\noindent Next, we consider the 2D viscous Burgers' equation
\begin{equation} \label{eq: 2d burger}
u_t + uu_x - \nu u_{xx} = 0
\end{equation}
where $u(x,t)$ is the solution function and $\nu=0.01$ is viscosity. We enforce both an initial condition $u(x,0) = \sin x$ and a boundary condition $u(0,t)=u(2\pi,t)=0$ on the overall domain $[0,2\pi]^2$.

Hence the total training loss is a weighted sum of the PDE residual, initial condition loss, and boundary condition loss,
\begin{equation}
    \begin{aligned}
\label{eq: 2D Burgers Loss}
    \mathcal{L}_{\text{pde}}(u_\theta) 
    \ & = \ \int  \!\!\!\! \int_{[0,2\pi]^2} \big|\partial_t u_\theta(x,t) + u_\theta(x,t) \  \partial_x u_\theta(x,t) - \nu \ \partial_{xx}u_\theta(x,t)\big|^2\mathrm{d}x   \ \! \mathrm{d}t 
    \\ & \qquad \quad  + \ \alpha \ \int_{[0,2\pi] \times \{0\}} |u_{\theta}(x,0) - \sin x|^2 \mathrm{d}x
 + \ \beta \ \int_{\{0,2\pi\} \times [0,2\pi]} |u_{\theta}(0,t) - 0|^2 \mathrm{d}t.
    \end{aligned}
\end{equation}

\hfill 

\subsubsection{Results}

We compare the performance of \FCPINO{} to baseline models for solving a single instance of the 2D Burgers' equation, where derivatives in the physics loss are computed using spectral differentiation. For each architecture, we performed a hyperparameter grid search, varying the number of Fourier modes, learning rate, scheduler patience, and the $(c,d)$ continuation parameters. The optimal hyperparameters for each architecture are provided in \Cref{appx:hyper_2d_burgers}, while \Cref{tab:burgers2D_start_baseline} reports the errors achieved by the best model for each architecture. \FCPINO{} performs very well, surpassing \padPINO{} by two orders of magnitude, while \PINOPadOut{} and \standardPINO{} again fail to satisfy the governing equations.

\begin{table}[h] \vspace{4mm}
    \caption{PDE residual, initial condition (IC) and boundary condition (BC) losses for \FCPINO{} with \textsc{FC--Gram} and \textsc{FC--Legendre}, and baseline PINO architectures for the 2D Burgers' equation \eqref{eq: 2d burger}.}
    \label{tab:burgers2D_start_baseline}
    \centering 
    \begin{tabular}{lccc}
        \toprule
        \textbf{Model} &  \hspace{1.5mm} \textbf{PDE Residual} \hspace{1.5mm} & \hspace{1.5mm} \textbf{IC Loss} \hspace{1.5mm}  & \hspace{1.5mm}  \textbf{BC Loss} \hspace{1.5mm}  \\
        \midrule
        \textbf{\textsc{FC--Gram}} \\
        \hspace{3mm} \FCPINO{}   & $1.35 \times 10^{-7}$ & $\bm{7.46 \times 10^{-7}}$ & $5.09 \times 10^{-7}$ \\
        \midrule
        \textbf{\textsc{FC--Legendre}} \\
        \hspace{3mm} \FCPINO{}   & $\bm{1.17 \times 10^{-7}}$ & $8.25 \ \times 10^{-7}$ & $\bm{5.06 \times 10^{-7}}$ \\
        \midrule
        \textbf{\textsc{Baselines}} \\
        \hspace{3mm} \padPINO{}    & $2.8 \times 10^{-5}$ & $7.0 \times 10^{-6}$ & $5.3 \times 10^{-7}$ \\
        \hspace{3mm} \PINOPadOut{} & $1.9 \times 10^{-2}$ & $7.7 \times 10^{-4}$ & $1.7 \times 10^{-5}$ \\
        \hspace{3mm} \standardPINO{}     & $1.6 \times 10^{-2}$ & $1.7 \times 10^{-3}$ & $2.4 \times 10^{-4}$ \\
        \bottomrule
    \end{tabular} \vspace{12mm}
\end{table}

Regarding computational time to train the models, \Cref{fig:threshold_crossing_2D} shows that, for this 2D example with $260 \times 260$ resolution, \FCPINO{} reaches a loss of $10^{-4}$ very rapidly and then more slowly converges towards $10^{-7}$. Notably, \FCPINO{} reaches the minimal loss achieved by \padPINO{} within the same short training time and then surpasses it significantly with further training.

Examples of solutions and corresponding PDE residual plots are displayed in \Cref{appx:sol_2D_burgers}. We see that the predicted solution of \FCPINO{} converges to a smooth and stable profile. On the other hand, \standardPINO{} and \PINOPadOut{} do not produce the correct solution, and exhibit undesired patterns and artifacts. In particular, the solution predicted by \standardPINO{} resembles a distorted version of the true solution, resulting from the incorrect imposition of periodicity on a non-periodic problem. \padPINO{} does better than the other baselines, but still shows much higher errors compared to \FCPINO, especially near the discontinuity.

\hfill

\subsection{Navier--Stokes Equations} \label{sec: NS}

\subsubsection{Problem Setting}

Finally, we test \FCPINO{} in 3D using the Navier--Stokes equations
\begin{equation}\label{eq:NS}
\begin{split}
\partial_t u(x,t) + u(x,t) \cdot \nabla u(x,t) &= -\frac{1}{\rho}\nabla p(x,t) + \frac{1}{Re} \Delta u(x,t), \qquad x \in (0,1)^2, t \in (0,T]  \\
\nabla \cdot u(x,t) &= 0, \qquad \qquad \qquad \qquad \quad x \in (0,1)^2, t \in [0,T]  \\
u(x,0) &= u_0(x), \qquad \qquad \qquad \quad x \in (0,1)^2 
\end{split}
\end{equation} We study the standard cavity flow on domain $D = (0,1)^2$ with $T=10$s, where $u$ is velocity, $p $ is pressure, $\rho=1$ is density, and $Re= 500$. We assume zero initial conditions, and the no-slip boundary condition where $u(x,t) = (0, 0)$ at left, bottom, and right walls and $u(x,t) = (1, 0)$ on top. We also start from zero initial conditions, $u(0,t) = (0,0)$ for $x\in (0,1)^2$. We are interested in learning the solution on the time interval $[5,10]$. The main challenge lies in handling the boundary conditions within the velocity–pressure formulation. The total training loss is a weighted sum of the PDE residual, boundary conditions loss, and initial condition loss.

\subsubsection{Results}

\vspace{1mm}

We compare the performance of \FCPINO{} to the baseline models for solving a single instance of the the Navier--Stokes equations, where derivatives in the physics loss are computed using spectral differentiation. For each architecture, we performed a hyperparameter grid search, varying the number of Fourier modes, learning rate, scheduler patience, and the $(c,d)$ continuation parameters. 

\clearpage

The optimal hyperparameters for each architecture are provided in \Cref{appx:hyper_Navier_Stokes}, while \Cref{tab:ns_start_baseline} reports the errors achieved by the best model for each architecture. In addition, \Cref{fig:NS_u_combined,fig:NS_v_combined} display the predicted solutions and the corresponding error plots for the different PINO architectures. We can see that \FCPINO{} converges to the correct solution with minimal PDE residual.

\begin{table}[h] \vspace{3mm}
    \caption{PDE residual, initial condition (IC) loss, and boundary condition (BC) loss for \FCPINO{} with \textsc{FC--Gram} and \textsc{FC--Legendre}, and baseline PINO architectures for the Navier--Stokes equations \eqref{eq:NS}.}
    \label{tab:ns_start_baseline}
    \centering \vspace{-1.2mm}
    \begin{tabular}{lccc}
        \toprule
        \textbf{Model} & \hspace{3mm} \textbf{PDE Residual}  \hspace{3mm} &  \hspace{3mm} \textbf{IC Loss}  \hspace{3mm}  &  \hspace{3mm} \textbf{BC Loss}  \hspace{3mm}  \\
        \midrule
        \textbf{\textsc{FC--Gram}} \\
        \hspace{6mm} \FCPINO{}   & \bm{{$7.5 \times 10^{-7}$}}  & \bm{$4.5 \times 10^{-7}$} & \bm{$6.7 \times 10^{-7}$} \\
        \midrule
        \textbf{\textsc{FC--Legendre}} \\
        \hspace{6mm} \FCPINO{}   & $1.1 \times 10^{-6}$  & $1.4 \times 10^{-6}$ & $1.1 \times 10^{-6}$ \\
        \midrule
        \textbf{\textsc{Baselines}} \\
        \hspace{6mm} \padPINO{}    & $8.0 \times 10^{-6}$  & $9.2 \times 10^{-6}$ & $3.5 \times 10^{-6}$ \\
        \hspace{6mm} \PINOPadOut{} & $3.7 \times 10^{-3}$  & $2.0 \times 10^{-2}$ & $6.3 \times 10^{-2}$ \\
        \hspace{6mm} \standardPINO{}     & $3.2 \times 10^{-2}$  & $5.9 \times 10^{-3}$ & $3.4 \times 10^{-3}$ \\
        \bottomrule
    \end{tabular} \vspace{6mm}
\end{table}

The results show that \PINOPadOut{} and \standardPINO{} fail to satisfy the governing equations. \padPINO{} is the only baseline that captures the solution correctly, yet as shown in \Cref{tab:ns_start_baseline} and \Cref{fig:NS_u_combined,fig:NS_v_combined}, it is still outperformed by one order of magnitude by \FCPINO{}. Regarding computational time to train the models, \Cref{fig:threshold_crossing_3D} shows that, for this specific 3D problem with $64\times64\times 50$ resolution, \standardPINO{} displays a faster convergence rate but learns an incorrect solution operator. Among the remaining architectures, \FCPINO{} reaches error thresholds the fastest, and converges to lower losses.

\hfill \\

\section*{Conclusion}

The reliable computation of highly accurate and computationally scalable  derivatives remains a central challenge in physics-informed machine learning for solving PDEs. Spectral differentiation stands out in this setting for its ability to compute derivatives with high accuracy and low computational cost via frequency-space operations. However, physics-informed neural operators (PINOs) with spectral differentiation suffer from a fundamental limitation: they assume periodicity, which can lead to artifacts such as the Gibbs phenomenon when applied to non-periodic functions. To address this, we proposed integrating Fourier continuation (FC) into the PINO framework, thereby defining the \FCPINO{} framework, enabling spectral derivatives computation for non-periodic functions without sacrificing efficiency or accuracy.   

The numerical experiments conducted on the self-similar and viscous Burgers' equations and Navier--Stokes equations exemplified the difficulties faced by the baseline PINO methods, where spectral differentiation is used naively or only with padding. In particular, \standardPINO{} and \PINOPadOut{} were not able to solve any of these 3 PDEs, and, while \padPINO{} displayed better results, it was still outperformed by several orders of magnitude on every aspect for every single problem by \FCPINO{}. Our empirical results demonstrate that \FCPINO{} can robustly handle non-periodic and non-smooth problems. In addition, they do so while remaining scalable thanks to the use of spectral differentiation. 

Overall, our findings strongly suggest that the proposed FC-based enhancement is critical in PINO when using spectral differentiation and offers a principled and scalable path forward for extending PINO to a broader class of PDE problems, especially when high-precision solutions are required.

One long-term goal of the machine learning community is to enable progress on the millennium question concerning the finite-time blowup of the Navier--Stokes equation, where ~\citep{PINN_3rd_deriv,wang2025high}, already demonstrated that machine learning can identify a very important candidate for a self-similar blowup scenario for the axis-symmetric Euler equation with boundary~\citep{luo2014potentially, luo2014toward}. Generally, to upgrade numerical candidates to rigorous proofs, one uses rigorous analysis to show stability around the approximate profile with high precision. Further research in this area will have important implications for fluid blowup research.

\hfill

\section*{Acknowledgements}
A. Ganeshram and H. Maust gratefully acknowledge the support of the Lynn A. Booth and Kent Kresa SURF Fellowship. Z. Li gratefully acknowledges the financial support from the Kortschak Scholars, PIMCO Fellows, and Amazon AI4Science Fellows programs. O. P. Bruno and D. Leibovici gratefully acknowledge support from NSF, AFOSR, and ONR through contracts DMS-2109831, FA9550-21-1-0373, FA9550-25-1-0015, and N00014-16-1-2808. T. Y. Hou is in part supported by the NSF grant DMS-2508463, the Choi Family Gift Fund, and the Charles Lee Powell endowed chair. A. Anandkumar is supported in part by the Bren endowed chair, ONR (MURI grant N00014-18-12624), and by the AI2050 senior fellow program at Schmidt Sciences. 

\hfill

\section*{Code availability}

The FC-PINO implementation and code used for the numerical experiments are publicly available at \href{https://github.com/neuraloperator/FC_PINO}{\nolinkurl{github.com/neuraloperator/FC_PINO}}.

\hfill 

\section*{Author contributions}

\noindent\textbf{AG:} Jointly designed the numerical experiments and ablation studies with VD. Conducted all experiments and ablation studies, and refined and expanded the earlier experiments. Set up most of the code used in the study, developed and polished the FC-PINO implementation, and prepared all figures. 

\noindent\textbf{VD:} Refined the FC-PINO approach and developed clean, robust FC-Gram and FC-Legendre implementations. Jointly designed the numerical experiments with AG and supervised code refinement and cleanup. Developed the background and methodological exposition, carried out the analysis, and wrote and revised the current manuscript in its entirety. 

\noindent\textbf{HM:} Conducted numerical experiments for the initial version of the study.

\noindent\textbf{ZL:} Originated the preliminary FC-PINO approach, contributed initial Fourier continuation and FC-PINO implementations, and supervised the initial phase of the project. 

\noindent\textbf{YW and TH:} Identified relevant blowup problems and provided scientific expertise on blowup phenomena.  

\noindent\textbf{DL and OB:} Contributed initial Fourier continuation implementations, particularly for FC-Gram. Drew on their prior experience applying Fourier continuation to numerical PDE solvers to provide valuable technical guidance on continuation methods and spectral differentiation. 

\noindent\textbf{AA:} Provided overall scientific supervision and feedback throughout the project. Helped shape the research direction, frame the study's objectives, and guide the selection and scope of the numerical experiments.\\

\bibliography{FC_PINO}
\bibliographystyle{plainnat}

\appendix

\newpage 

\section{Comparison with Alternative FC-based PINO Architectures} \label{appx: alternative PINO FC}

\vspace{6mm}

\subsection{Alternative FC-based PINO Architectures}

\vspace{4mm}

First recall the definition of the FNO from \Cref{sec: PINO},
\begin{equation} \label{eq: FNO 2 appx}
   \mathcal{Q} \ \circ \  \sigma (W_{L} + \mathcal{K}_{L} + b_L) \ \circ \  \cdots \ \circ \  \sigma(W_1 + \mathcal{K}_1 + b_1) \ \circ \  \mathcal{P},
\end{equation}
where $\mathcal{K}$ are Fourier integral kernel operators, \(\mathcal{P}\) and \(\mathcal{Q} \) are pointwise lifting and projection neural networks, $b_l $ are bias terms, and \(\sigma\) are activation functions. \\ 

\hfill 

\noindent \textbf{\EndFCPINO{}} has the same structure as the FNO~\eqref{eq: FNO 2 appx}, except that FC is used just before the final projection operator $ \mathcal Q$ in the derivative computation. Just like in \FCPINO{}, one can then compute the derivative of the model output using the chain rule on the derivative of the model up to the final projection operator $ \mathcal Q$ and the derivative of $ \mathcal Q$.

\vspace{8mm}

\begin{figure}[h]
    \centering
\includegraphics[width=0.92\textwidth]{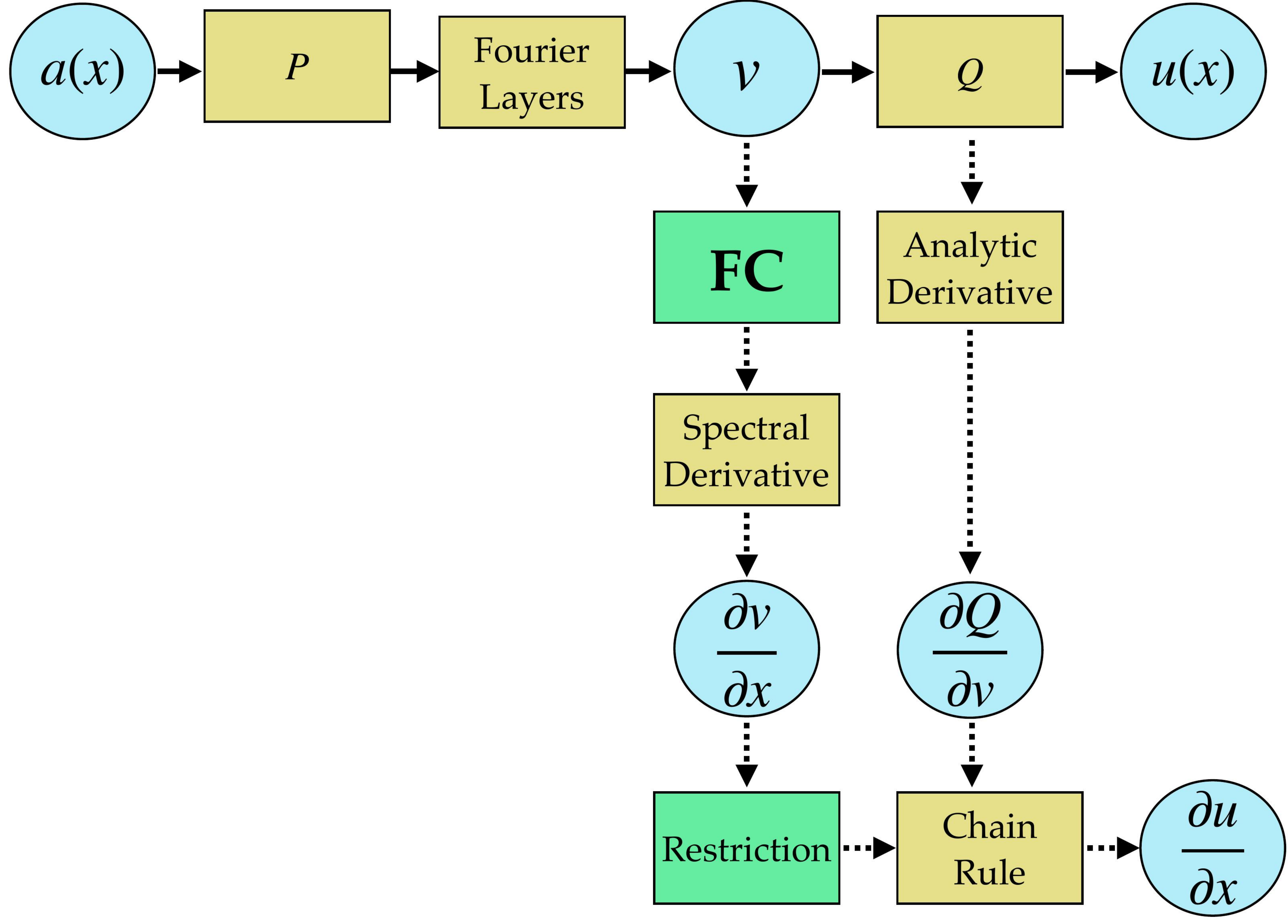}\\
    \label{fig:end_FC_diagram}
\end{figure}

\clearpage

\noindent \textbf{\outsideFCPINO{}} relies on the standard FNO architecture~\eqref{eq: FNO 2 appx}, and uses FC on the output of the model (outside the model) only before the spectral derivative computation. More explicitly, the model is the standard FNO, and the derivatives are computed outside the model using spectral differentiation on an extended domain

\vspace{5mm}

\begin{figure}[h]
    \centering
    \includegraphics[width=0.96 \textwidth]{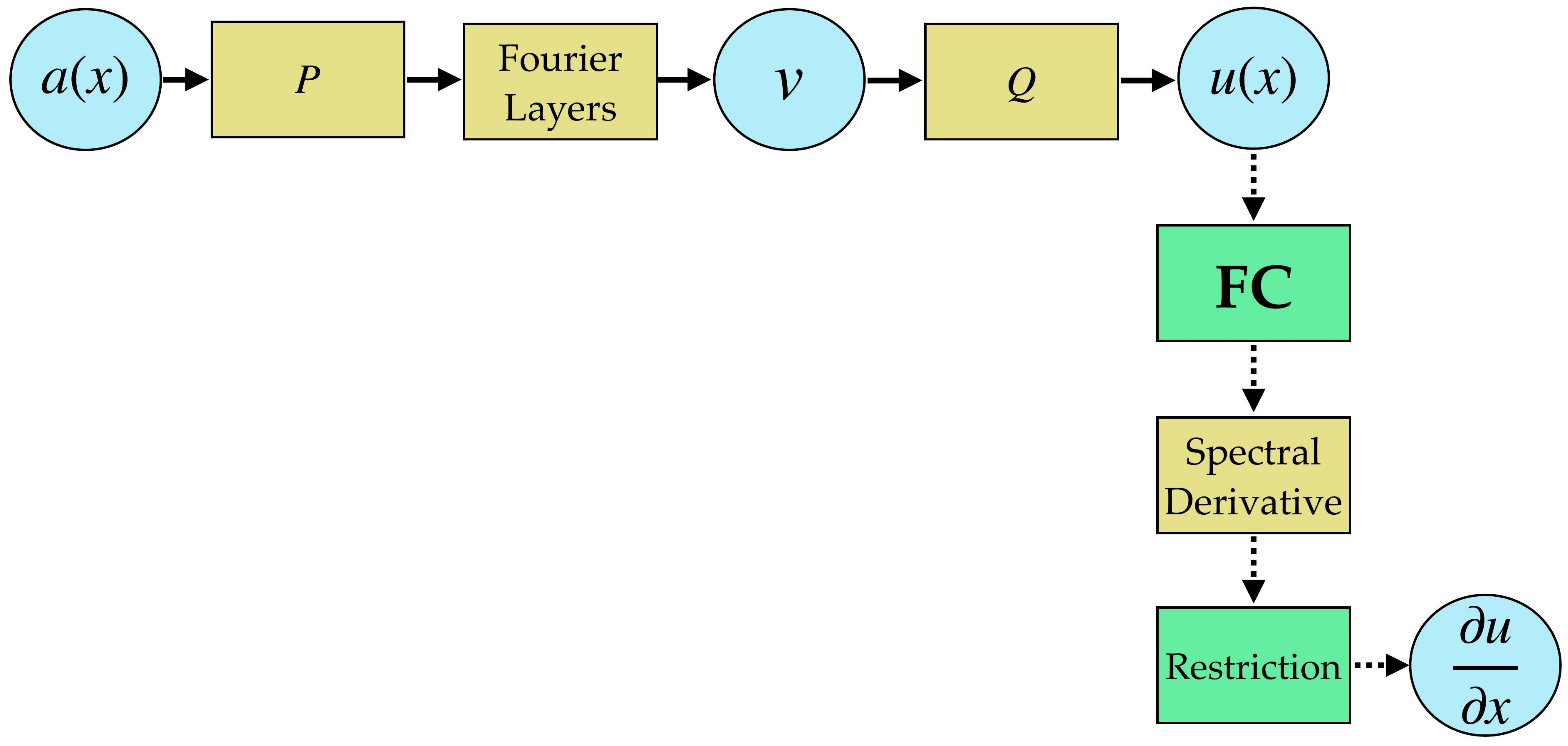}\\
    \label{fig:out_FC_diagram}
\end{figure}

\vspace{6mm}

Note that \EndFCPINO{} and \outsideFCPINO{} are mathematically equivalent, in the sense that they represent the same mapping and yield the same forward pass and derivatives in exact arithmetic. However, they induce different computational graphs for backpropagation, and therefore do not produce identical gradients in finite-precision arithmetic. In particular, even small discrepancies can be magnified when derivatives are computed via spectral differentiation and subsequently aggregated into the PDE residual loss. Consequently, although the two formulations are mathematically equivalent, the resulting backpropagated gradient can differ enough to steer the optimizer along substantially different trajectories, leading to different training dynamics and final model weights.

\hfill

\subsection{Comparison on Numerical Experiments}

\vspace{2mm}

We now compare the performance of \outsideFCPINO{} and \EndFCPINO{} against \FCPINO{} for the 1D and 2D Burgers' equations and the Navier--Stokes equations considered in this paper. \Cref{tab:burgers1D_legendre_gram_combined,tab:burgers2D_legendre_gram_combined,tab:ns_legendre_gram_combined} show the results obtained by these architectures. As before, for a fair comparison, we performed a hyperparameter grid search over the model, training, and continuation hyperparameters and display the results obtained with the best configuration. The optimal hyperparameters for each architecture are provided in \Cref{appx:hyperparameters}.

We first note that \outsideFCPINO{} performs well across all considered PDEs but is consistently several orders of magnitude behind the other two FC-based approaches. While \EndFCPINO{} achieves strong performance on all problems, \FCPINO{} consistently outperforms it slightly in nearly every aspect for every PDE considered here.

\begin{table}[h] \vspace{-10mm}
    \caption{PDE residual, boundary condition (BC) loss, smoothness loss, and total loss for the FC-based PINO architectures using \textsc{FC--Legendre} and \textsc{FC--Gram} when solving the 1D Burgers' equation for $\lambda = 0.5$.} \vspace{-1.5mm}
    \label{tab:burgers1D_legendre_gram_combined}
    \centering
    \resizebox{\textwidth}{!}{%
    \begin{tabular}{lcccc}
        \toprule
        \textbf{Model} & \textbf{PDE Residual} & \textbf{BC Loss} & \textbf{Smoothness} & \textbf{Total Loss} \\
        \midrule
        \textbf{\textsc{FC--Gram}} \\
        \hspace{3mm} \FCPINO{}   & $1.9 \times 10^{-11}$ & \bm{$9.2 \times 10^{-15}$} & \bm{$5.5 \times 10^{-9}$} & \bm{$6.7 \times 10^{-10}$} \\
        \hspace{3mm} \EndFCPINO{}     & \bm{$7.9 \times 10^{-13}$} & $2.9 \times 10^{-14}$ & $8.3 \times 10^{-8}$ & $8.3 \times 10^{-9}$ \\
        \hspace{3mm} \outsideFCPINO{} & $2.1\times 10^{-12}$ & $2.1 \times 10^{-13}$ & $1.3 \times 10^{-7}$ & $1.3 \times 10^{-8}$ \\
        \midrule
        \textbf{\textsc{FC--Legendre}} \\
        \hspace{3mm} \FCPINO{}   & $\bm{1.8 \times 10^{-12}}$ & $\bm{2.2 \times 10^{-15}}$ & $\bm{2.8 \times 10^{-8}}$ & $\bm{2.8 \times 10^{-9}}$ \\
        \hspace{3mm} \EndFCPINO{}     & $3.6 \times 10^{-12}$ & $6.8 \times 10^{-14}$ & $1.7 \times 10^{-7}$ & $1.7 \times 10^{-8}$ \\
        \hspace{3mm} \outsideFCPINO{} & $1.0 \times 10^{-9}$ & $1.6 \times 10^{-12}$ & $3.2 \times 10^{-5}$ & $3.2 \times 10^{-6}$ \\
        \bottomrule
    \end{tabular} \vspace{3mm}
    }
\end{table}

\begin{table}[h]  \vspace{3mm}
    \caption{PDE residual, initial condition (IC) loss, and boundary condition (BC) loss for the FC-based PINO architectures using \textsc{FC--Legendre} and \textsc{FC--Gram} when solving the 2D Burgers' equation \eqref{eq: 2d burger}.}
    \label{tab:burgers2D_legendre_gram_combined}
    \centering \vspace{-1.5mm}
    \begin{tabular}{lccc}
        \toprule
        \textbf{Model} & \hspace{2mm} \textbf{PDE Residual} \hspace{2mm} & \hspace{2mm} \textbf{IC Loss} \hspace{2mm}  & \hspace{2mm}  \textbf{BC Loss} \hspace{2mm}  \\
        \midrule
        \textbf{\textsc{FC--Gram}} \\
        \hspace{3mm} \FCPINO{}   & $\bm{1.35 \times 10^{-7}}$ & $\bm{7.46 \times 10^{-7}}$ & $\bm{5.09 \times 10^{-7}}$ \\
        \hspace{3mm} \EndFCPINO{}     & $1.73 \times 10^{-7}$ & $7.55 \times 10^{-7}$ & $5.21 \times 10^{-7}$ \\
        \hspace{3mm} \outsideFCPINO{} & $3.48 \times 10^{-5}$ & $6.89 \times 10^{-6}$ & $6.21  \times 10^{-7}$ \\
        \midrule
        \textbf{\textsc{FC--Legendre}} \\
        \hspace{3mm} \FCPINO{}   & $\bm{1.17 \times 10^{-7}}$ & $\bm{8.25 \times 10^{-7}}$ & $\bm{5.06 \times 10^{-7}}$ \\
        \hspace{3mm} \EndFCPINO{}     & $3.88 \times 10^{-7}$ & $9.23 \times 10^{-7}$ & $6.57 \times 10^{-7}$ \\
        \hspace{3mm} \outsideFCPINO{} \ \ \ & $2.46 \times 10^{-5}$ & $6.67 \times 10^{-6}$ & $5.74 \times 10^{-7}$ \\
        \bottomrule
    \end{tabular}
    \vspace{3mm}
\end{table}

\begin{table}[h] \vspace{3mm}
    \caption{PDE residual, initial condition (IC) loss, and boundary condition (BC) loss for the FC-based PINO architectures using \textsc{FC--Legendre} and \textsc{FC--Gram} when solving the Navier--Stokes equations \eqref{eq:NS}.}
    \label{tab:ns_legendre_gram_combined}
    \centering \vspace{-1.5mm}
    \begin{tabular}{lccc}
        \toprule
        \textbf{Model} & \hspace{2mm} \textbf{PDE Residual} \hspace{2mm} & \hspace{2mm} \textbf{IC Loss} \hspace{2mm} &\hspace{2mm}  \textbf{BC Loss} \hspace{2mm} \\
        \midrule
        \textbf{\textsc{FC--Gram}} \\
        \hspace{6mm} \FCPINO{}   & \bm{$7.5 \times 10^{-7}$}  & \bm{$4.5 \times 10^{-7}$} & \bm{$6.7 \times 10^{-7}$} \\
        \hspace{6mm} \EndFCPINO{}     & $1.3 \times 10^{-6}$ & $9.6 \times 10^{-7}$ & $1.9 \times 10^{-6}$ \\
        \hspace{6mm} \outsideFCPINO{} & $1.3 \times 10^{-5}$ & $1.7 \times 10^{-6}$ & $7.2 \times 10^{-6}$ \\
        \midrule
        \textbf{\textsc{FC--Legendre}} \\
        \hspace{6mm} \FCPINO{}   & $\bm{1.1 \times 10^{-6}}$  & $1.4 \times 10^{-6}$ & $\bm{1.1 \times 10^{-6}}$ \\
        \hspace{6mm} \EndFCPINO{}     & $1.3 \times 10^{-6}$  & $\bm{9.5 \times 10^{-7}}$ & $1.7 \times 10^{-6}$ \\
        \hspace{6mm} \outsideFCPINO{} & $2.0 \times 10^{-5}$  & $2.2 \times 10^{-6}$ & $1.5 \times 10^{-5}$ \\
        \bottomrule
    \end{tabular}
\end{table}         

\clearpage

\section{Extended Discussion of Fourier Continuation Methods}  \label{appx: FC}

\subsection{Construction of the Extension matrices}

Recall the general setting. Let \(f:[a,b]\to\mathbb{R}\) be given on a grid of \(n\) points, with equidistant nodes \(x_k = a + hk\)  for   \(k = 0,\dots,n-1\), where \(h = \frac{b-a}{n-1}\). Denote $f_n = f(x_n)$. The goal is to produce a new sequence~\(\tilde f\) of length \((n+c)\) which is exactly periodic of period \((n+c)\) and agrees with \(f\) on the original interval. Define the left and right boundary vectors of width \(d\),
\begin{align}
f_\ell = (f_0, f_1, \ldots, f_{d - 1})^\top, \quad f_r = (f_{n - d}, \ldots, f_{n-1})^\top.
\end{align}
\textsc{FC--Legendre} and \textsc{FC--Gram} employ distinct strategies for constructing the extension signal $\textcolor{NavyBlue}{f_{\text{ext}}} \in \mathbb{R}^c$, and the periodic extended signal $\tilde f$ is then obtained by concatenating the original signal $f$ and the extension $\textcolor{NavyBlue}{f_{\text{ext}}}$ on both sides, 
\begin{align}
\tilde f & =  (\textcolor{NavyBlue}{f_{\text{ext},c/2},\dots,f_{\text{ext},c-1}},\,f_0,\dots,f_{n-1},\,\textcolor{NavyBlue}{f_{\text{ext},0},\dots,f_{\text{ext},c/2-1}}).
\end{align}

\hfill

\subsubsection{\textsc{FC--Legendre}} \label{appx: Construction Legendre matrix}

In \textsc{FC--Legendre}, the left and right boundary vectors are concatenated into the vector
$y = (f_{r}^\top, f_{\ell}^\top)^\top \in \mathbb{R}^{2d}$, and the extension $\textcolor{NavyBlue}{f_{\text{ext}}} \in \mathbb{R}^c$ is obtained via $ E y = \textcolor{NavyBlue}{f_{\text{ext}}} \in \mathbb{R}^c$. Here, $E \in \mathbb{R}^{c \times 2d}$ is an extension matrix, built from evaluations of shifted Legendre polynomials of degree \((2d-1)\), to do polynomial interpolation in between the right and left boundaries. For fixed parameters $(c, d)$, the \textsc{FC--Legendre} extension matrix $E \in \mathbb{R}^{c \times 2d}$ is independent of the input signal of length $n > 2d$ (by construction; see details below). As a result, $E$ can be precomputed offline and reused in \textsc{FC--PINO} without the need for recomputation. \\

\noindent \textbf{Construction of the \textsc{FC--Legendre} extension matrix $E$.} We consider a larger grid of $(2d + c)$ points $x_k = a + hk$ spanning the extended interval \([a, b_{\text{ext}}]\), where $b_{\text{ext}} = a + (2d+c - 1)h$. We define a family of orthogonal polynomials on \([a, b_{\text{ext}}]\) by shifting and scaling the standard Legendre polynomials. These standard Legendre polynomials \(\{\mathsf{P}_j^{\text{std}}(\xi)\}\) are defined on the reference interval \([-1,1]\) and satisfy the recurrence relation 
\begin{align}
(j+1)\mathsf{P}_{j+1}^{\text{std}}(\xi) = (2j+1)\xi \mathsf{P}_j^{\text{std}}(\xi) - j\mathsf{P}_{j-1}^{\text{std}}(\xi), \quad j \geq 1,
\end{align}
with $\mathsf{P}_0^{\text{std}}(\xi) = 1$ and $
\mathsf{P}_1^{\text{std}}(\xi) = \xi$. The shifted Legendre polynomials are then defined on the interval \([a, b_{\text{ext}}]\) via $
\mathsf{P}_j(x) = \mathsf{P}_j^{\text{std}}\left( \frac{2(x - a)}{b_{\text{ext}} - a} - 1 \right)$, and are orthogonal with respect to the \(L^2\) inner product over \([a, b_{\text{ext}}]\). As a result, they form a natural basis for polynomial approximation on this interval.

Now let \(\{x_k^{\mathrm{fit}}\}_{k=0}^{2d-1}\) be the union of the \(d\) leftmost and \(d\) rightmost nodes from the original domain \([a, b]\), corresponding to the boundary strips with $f_{\ell}$ and $f_{r}$. Let \(\{x_i^{\mathrm{ext}}\}_{i=0}^{c-1}\) be the \(c\) interior grid points (in \([a, b_{\text{ext}}]\)) where we wish to evaluate the polynomial extension. We define the interpolation matrix \(X \in \mathbb{R}^{2d \times 2d}\) and evaluation matrix \(Q \in \mathbb{R}^{c \times 2d}\) by evaluating each shifted Legendre polynomial at these points:
\begin{align}
X_{jk}  = \mathsf{P}_j(x_k^{\mathrm{fit}}), \quad 0 \le j, k < 2d, \qquad  \ \ 
Q_{ij}  = \mathsf{P}_j(x_i^{\mathrm{ext}}), \quad 0 \le i < c, \; 0 \le j < 2d.
\end{align}
The extension matrix $E$ is then computed via the least-squares projection $E = Q X^{\dagger}$,
where \(X^{\dagger}\) denotes the Moore--Penrose pseudoinverse of \(X\). This matrix \(E\) maps the boundary data vector \(y \in \mathbb{R}^{2d}\) to the polynomial extrapolation \(Ey \in \mathbb{R}^c\) of degree less then \( \!2d\) that best fits the boundary data in the \(\ell^2\)-sense. Because the Legendre polynomials are smooth and the continuation uses a low-degree polynomial, the extended sequence \(\tilde{f}\) interpolates the original data \(f\) exactly on the interval \([a,b]\) and smoothly transitions into the periodic extension. Also note that the extension matrix $E$ is computed using only the information contained in the boundary vectors $f_{\ell}$ and $ f_r$ of width \(d\), and not the remaining interior part $f_{\text{int}}$ of the original signal. \\

\subsubsection{\textsc{FC--Gram}} \label{appx: Construction Gram matrix}

We now describe the (Accelerated) \textsc{FC--Gram} method from \citep{AMLANI2016} which is inspired from earlier \textsc{FC--Gram} methods~\citep{Accurate_FC_2007,Fourier_continuation_2011}. See \citep{AMLANI2016} for more details and illustrations of \textsc{FC--Gram}.\\

In \textsc{FC--Gram}, the extension values $\textcolor{NavyBlue}{f_{\text{ext}}}$ are formed by projecting the left and right boundary vectors $f_\ell, f_r$ onto Gram bases via matrices \(Q_\ell\), \(Q_r\), and then blending their respective continuations using the continuation matrices \(A_\ell\), \(A_r\):
\begin{align}
\textcolor{NavyBlue}{f_{\text{ext}}} = A_\ell Q_\ell^\top f_\ell + A_r Q_r^\top f_r.
\end{align} 
For fixed parameters $(c, d)$, the matrices $A_{\ell}, A_r \in \mathbb{R}^{c \times d}$ and $Q_{\ell}, Q_r \in \mathbb{R}^{d \times d}$ are independent of the input signal of length $n > 2d$ (by construction; see details below). As a result, they can be precomputed offline and reused in \textsc{FC--PINO} without the need for recomputation. \\

\noindent \textbf{Construction of the \textsc{FC--Gram} matrices $A_\ell,  Q_\ell,  A_r,  Q_r$.}
The left boundary vector $f_\ell$ is first projected onto an orthonormal basis of Gram polynomials \(B_\ell\) which spans the space of polynomials of degree less than \(d\) and are orthonormal with respect to the discrete inner product over the grid points corresponding to $f_\ell$. Here, the basis \(B_\ell = \{p_0, \dots, p_{d-1}\}\) is constructed by applying stabilized Gram–Schmidt orthogonalization to the Vandermonde matrix of monomials \(\{1, x, x^2, \ldots, x^{d-1}\}\) evaluated at the corresponding grid points. Letting \(P_\ell \in \mathbb{R}^{d \times d}\) be the Vandermonde matrix, the orthonormal basis is obtained via QR factorization \(P_\ell = Q_\ell R_\ell\), and the columns of \(Q_\ell \in \mathbb{R}^{d \times d}\) are the evaluations of the Gram basis functions on the grid. Similarly, the right boundary vector $f_r$ is projected onto an orthonormal basis of Gram polynomials \(B_r\), and an evaluation matrix $Q_r \in \mathbb{R}^{d \times d}$ is constructed. \\

Each Gram polynomial is smoothly extended beyond the matching region using a blending function defined over an extended interval consisting of: (i) a matching region of \(d\) points, (ii) a continuation region of \(c\) points, (iii) a region where the function vanishes. The extension is represented as a polynomial whose coefficients are precomputed by fitting the Gram polynomial in the matching region while ensuring decay in the zero region. This fitting uses oversampled grids and high-precision arithmetic to guarantee smoothness and numerical stability. The resulting functions are evaluated at the continuation nodes to form the columns of the continuation matrices \(A_\ell \in \mathbb{R}^{c \times d}\) and \(A_r \in \mathbb{R}^{c \times d}\). \\

We emphasize that the \textsc{FC--Gram} matrices are computed using only the information contained in the boundary vectors $f_{\ell}$ and $ f_r$ of width \(d\), and not the remaining interior part $f_{\text{int}}$ of the original signal. \\

\subsection{Tuning the Fourier Continuation Parameters}  \label{appx: tuning FC parameters}

\subsubsection{Tuning \textsc{FC--Legendre}.} \label{appx: tuning FC Legendre}

Tuning \textsc{FC--Legendre} involves carefully balancing the degree $(2d-1)$ of the interpolating polynomial, the extension length \( c \), and noise sensitivity to achieve optimal derivative accuracy and numerical stability. \\

The parameter $c$ sets the number of extension points, balancing the need for sufficient length to accurately capture complex boundary behavior and achieve smooth continuation against the risk of ill-conditioning and noise amplification from overly long extensions. $c$ must be sufficiently large to capture boundary behavior and ensure smooth continuation, but overly long extensions can amplify noise and induce ill-conditioning. As a result, increasing $c$ can initially reduce the PDE residual, but beyond a certain point, further increases would degrade performance due to numerical instability. Thus, for a given resolution and value of $d$, one can start with $c\approx 30$ and gradually increase it until the PDE residual no longer improves or begins to worsen.\\

\begin{figure}[t]
\vspace{5mm}
  \centering
  \begin{minipage}{0.494\linewidth}
    \centering
     \includegraphics[width=\linewidth]{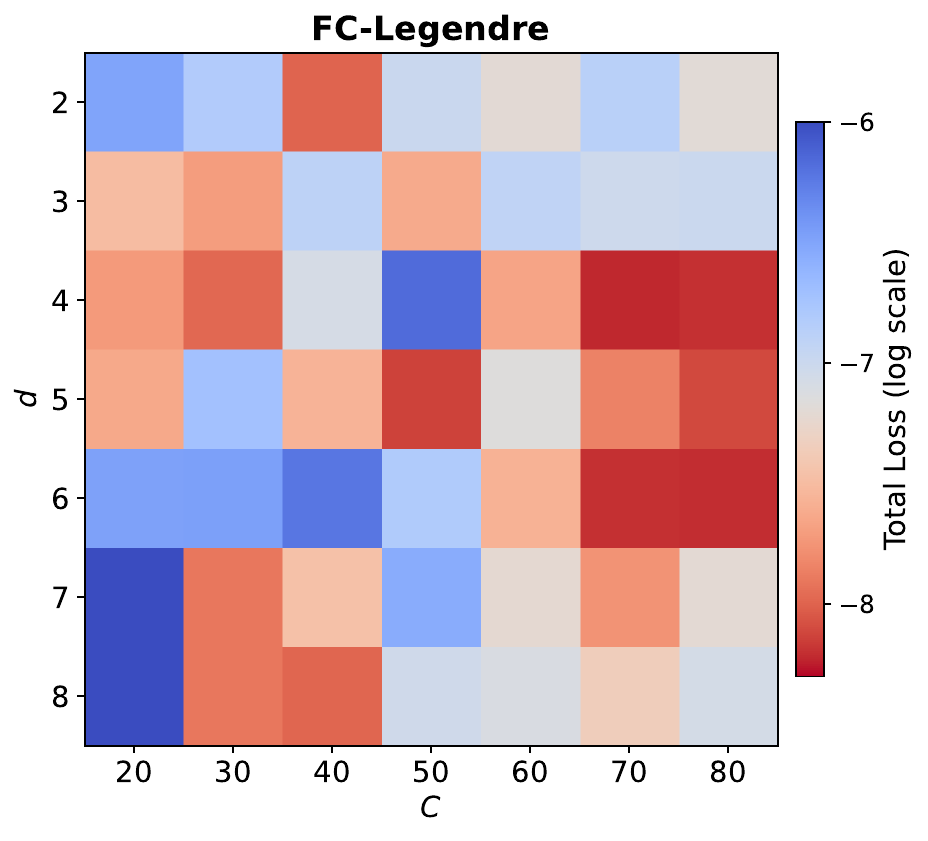}
     \vspace{-3mm}
   \end{minipage}%
   \hfill
   \begin{minipage}{0.494\linewidth}
     \centering
     \includegraphics[width=\linewidth]{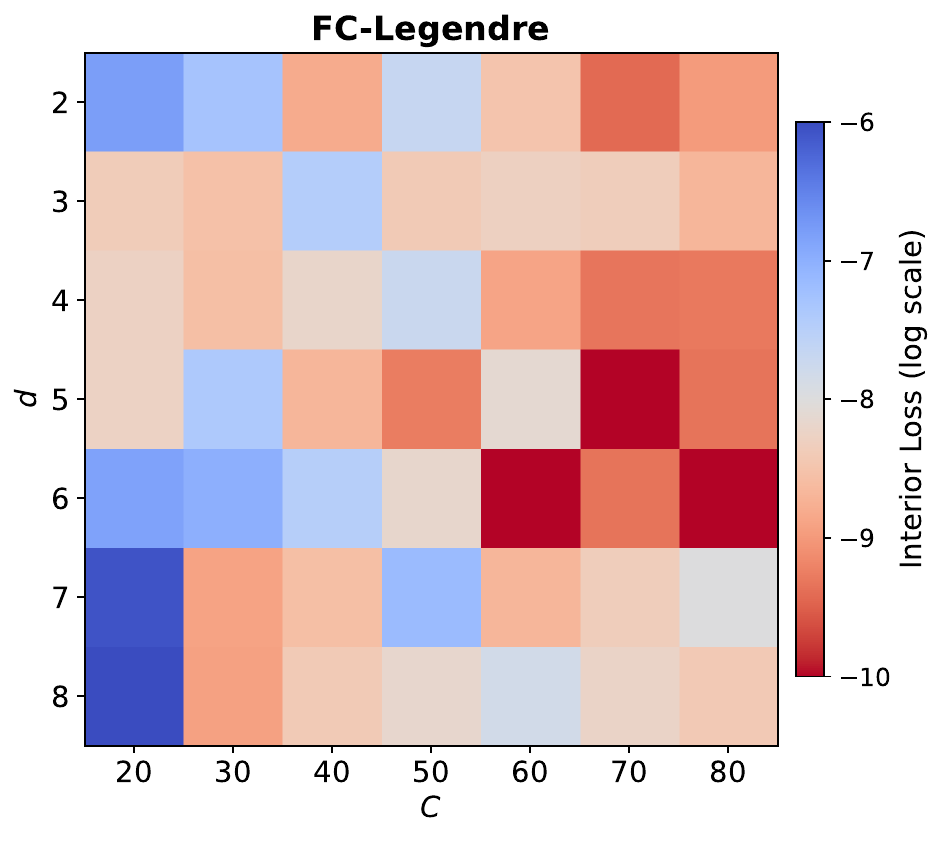}
  \end{minipage}  \vspace{-4mm}
  \caption{Effect of varying continuation parameters $c$ (continuation points) and d (degree) for \textsc{FC--Legendre} on Total Loss (left) and PDE Residual (right)}
  \label{fig:continuation-study-1d-burgers} \vspace{12mm}
\end{figure}

Regarding the degree of the polynomial interpolation, values of \( d \) between 4 and 6 are typically effective. Increasing \( d \) tends to improve accuracy for smooth signals without boundary noise since higher-degree polynomials provide a better approximation and help reduce derivative errors. However, as the value of \( d \) grows, the conditioning of the Legendre pseudoinverse worsens, so it is essential to monitor singular values and halt the increase once the conditioning degrades significantly. We observe that trade-off experimentally in \Cref{fig:continuation-study-1d-burgers} when fixing \( c \) large enough and increasing \( d \), with performance improving initially until a threshold where it starts degrading. Larger values of \( d \) can be preferable for problems with smoother solutions or when high-order derivatives are desired. For a fixed resolution and value of $c$ large enough, one can start with $d\approx 3$ and gradually increase it until the PDE residual no longer improves or begins to worsen. Note that a good choice of $c$ is dependent on the resolution of the problem.

Additionally, applying a low-pass filter to the boundary data or adjusting the numerical rank used in the pseudoinverse (e.g. via \texttt{rcond} in \texttt{np.linalg.pinv}) can be beneficial to reduce noise amplification in the higher modes. We also observe from \Cref{fig:continuation-study-1d-burgers} that \textsc{FC--Legendre} is more sensitive to the choice of \( (c, d) \) because of its global polynomial basis and the larger pseudoinverse problems it entails. This sensitivity means that careful tuning is necessary to avoid issues such as ill-conditioning and oscillatory artifacts. Increasing the signal resolution can help reduce the PDE residual error. When higher resolution is available, one can fix the continuation parameters, gradually increase the resolution, and monitor the resulting changes in the PDE residuals. This process can then be iterated, alternating between updating the continuation parameters and refining the resolution. \\

We emphasize that with all the configurations of $(c,d)$ tested, we achieve low errors in \Cref{fig:continuation-study-1d-burgers}. Careful tuning is only necessary when high-precision solutions are needed (e.g. in this case to go from $10^{-7}$ to $10^{-11}$). With systematic validation and parameter sweeps, \textsc{FC--Legendre} can be effectively tuned to deliver high-accuracy continuation and spectral differentiation, leading to low PDE residuals when paired with \FCPINO{}. \\

\subsubsection{Tuning \textsc{FC--Gram}.} \label{appx: Tuninf FC Gram}

Tuning \textsc{FC--Gram} can be more challenging due to its localized and more complex construction. Parameters to tune include (1) the polynomial degree \( d \), which controls the size of the local orthonormal basis and the number of matching points on each boundary segment, (2) the number of continuation points $c$, (3) the number of blending points, (4) the number of extra points to improve numerical stability, (5) the oversampling factor to ensure accurate fine-grid construction, (6) the number of modes to reduce in the SVD computation to suppress the smallest singular values without significantly impacting resolution, and (7) the symbolic computation precision. However, after extensive tuning conducted in previous papers~\citep{Accurate_FC_2007,Fourier_continuation_2011,AMLANI2016}, fixed values of all these parameters except $d$ have been determined to provide an appropriate trade-off between overall computational cost and stability which works well in most settings. In particular, fixing \( c = 50 \) works well across a range of \( d \) values and generally does not require further tuning. Typically, fewer than \( c = 50 \) extension points are needed to fully blend the signal to the target constant, ensuring a gradual taper that avoids sharp discontinuities or artifacts. \\

The remaining parameter \( d \) determines the degree of the local polynomials used to smoothly transition the signal from boundary values to a constant intermediate value. Higher \( d \) values allow better resolution of fine boundary features and improve approximation of higher-order derivatives, especially for smooth and noise-free data. However, increasing \( d \) excessively may cause local QR decompositions to become ill-conditioned, leading to overfitting of noise and oscillatory artifacts in the extension, so \( d \) should be kept small around 2 for problems with strong discontinuities. Increasing the signal resolution can help reduce the PDE residual error. When higher resolution is available, one can fix the continuation parameters, gradually increase the resolution, and monitor the resulting changes in the PDE residuals. This process can then be iterated, alternating between updating the continuation parameters and refining the resolution. \\

Overall, \textsc{FC--Gram} offers greater flexibility through its multiple tuning parameters but may require careful monitoring of local QR conditioning and noise overfitting. Its localized approach and blending taper yield robust and clean derivative fields without manual filtering. For the 1D Burgers' equation, \Cref{tab: Gram continuation parameters} shows that \textsc{FC--Gram} leads to high-precision solutions consistently across values of $d$, with minimal variations. This indicates that \textsc{FC--Gram} is very robust in this setting. With systematic validation and parameter sweeps, \textsc{FC--Gram} can be effectively tuned to deliver high-accuracy spectral continuation and differentiation. We emphasize that even without careful tuning of the parameter $d$ and keeping all other parameters fixed, we already obtained solutions with high accuracy.

\begin{table}[t] 
\centering  
\label{tab: Gram continuation parameters}
\caption{Effect of the choice of continuation parameters $d$ for \textsc{FC--Gram} on the Total Loss and PDE Residual when using \FCPINO{} to solve the 1D Burgers' equation~\eqref{eq: 1D Burger}.} 
\vspace{-1mm}
\large 
\renewcommand{\arraystretch}{1.1}

\begin{tabular}{c | c | c}

  \hspace{12mm}  & \hspace{6mm}Total Loss\hspace{6mm} 
    & \hspace{6mm}PDE Residual\hspace{6mm} \\

    \hline

    $d=2$ & $1.05\times10^{-8}$ \  
           & $3.74 \times 10^{-10}$ \\

    $d=3$ & $4.89\times10^{-9}$ \  
           & $9.52 \times 10^{-10}$ \\

    $d=4$ & $7.34\times10^{-10}$ 
           & $1.25 \times 10^{-10}$ \\

    $d=5$ & $8.29\times10^{-10}$ 
           & $1.12 \times 10^{-9}$ \  \\

    $d=6$ & $\mathbf{6.72\times10^{-10}}$ 
           & $\mathbf{1.92\times10^{-11}}$ \\

    $d=7$ & $6.60\times10^{-9}$ \ 
           & $1.07 \times 10^{-9}$ \  \\

    $d=8$ & $6.28\times10^{-9}$ \ 
           & $9.46 \times 10^{-10}$ \\

\end{tabular}

\vspace{1mm}
\end{table}

\vspace{5mm}

\subsection{Remark about Varying Resolution}

The extensions obtained by Fourier continuation can depend heavily on the continuation parameters $c$ and $d$. For a given \((c,d)\) pair, the FC layer always uses \(d\) boundary points on each side and produces an extension of length \(c\), independently of the grid size \(N\). As the resolution \(N\) increases and the grid spacing decreases, the physical width of the boundary strips and extension region shrinks, so the continuation is built from a progressively smaller portion of the physical domain. Beyond a certain resolution, the strip of width \(d\) can become too narrow in physical space to capture boundary layers or high frequency features, and the FC continuation may no longer reflect the true boundary behavior, even if the interior approximation remains accurate.

This behavior is analogous to the number of Fourier modes. If the number of modes is fixed while \(N\) grows, the representation becomes increasingly band limited and eventually fails to resolve the relevant dynamics. In the same way, fixed values of the continuation parameters \(c\) and \(d\) can become insufficient at high resolution, which can also prove problematic when the model is trained with losses at one resolution and evaluated at a much finer resolution in super-resolution or cross-resolution settings. In practice, one can choose \((c,d)\) that are robust over a target range of resolutions and further stabilize performance with multi-resolution training or progressive schedules that increase the number of Fourier modes together with the continuation parameters. The incremental FNO strategy~\citep{george2022incremental}, where the resolution and number of modes is increased progressively, can also be extended by progressively increasing the values of the Fourier continuation parameters $c$ and $d$.\\

\subsection{Properties and Comparison of \textsc{FC--Legendre} and \textsc{FC--Gram}}  \label{appx: properties and comparison of FC}

\vspace{1mm}

\textsc{FC--Gram} and \textsc{FC--Legendre} employ fundamentally different continuation strategies, each with unique pros and cons. As illustrated in \Cref{fig: FC Examples}, \textsc{FC--Legendre} constructs the extension by fitting a single Legendre polynomial of degree \( (2d -1) \) spanning the entire extended interval between the left and right boundaries. In contrast, \textsc{FC--Gram} takes a more local approach, projecting boundary data onto lower-order polynomial bases of degree \(d\) separately on each boundary and then smoothly blending the extensions outside the domain. While both approaches yield good results with minimal tuning alongside \FCPINO{}, one can offer advantages over the other in high-precision settings.

\vspace{6mm}

\subsubsection{Oscillation Attenuation and Stability}  

\vspace{1.5mm}

\textsc{FC--Gram} achieves robustness through its blending continuation strategy, which acts like a low-pass filter by smoothly damping boundary mismatches. This robustness is further strengthened by localized orthonormal Gram bases \(Q_{\ell}\) and \(Q_r\), constructed separately on the left and right boundary strips. These localized bases confine errors and high-frequency modes and noise within the extension region, preventing oscillations from contaminating the original domain signal. Consequently, spectral differentiation on the original domain interacts only with an attenuated boundary noise residual. This reduces the amplification of high-frequency components during differentiation and enhances spectral accuracy near boundaries. The localization also leads to better-conditioned QR factorizations of smaller Vandermonde matrices, reducing round-off errors and noise amplification during continuation.

In contrast, \textsc{FC--Legendre} employs a global Legendre polynomial basis for continuation, which can cause residual oscillations to propagate across the entire domain. Small boundary mismatches or noise generate oscillations near the endpoints that are amplified in spectral differentiation, due to \((ik)^m\) factors. Because the polynomial fit is global, even minor perturbations in boundary data affect all expansion coefficients, causing oscillations to “ring” throughout the spectrum and degrade accuracy. Although \textsc{FC--Legendre} can represent higher modes across the domain, its susceptibility to noise and the ill-conditioning of large pseudoinverse computations often reduce robustness and spectral accuracy near boundaries.

\vspace{6mm}

\subsubsection{Simplicity}

\vspace{1.5mm}

\textsc{FC--Legendre} benefits from a simpler implementation, relying on straightforward global Legendre polynomial interpolation with few parameters to tune. \textsc{FC--Gram}, by contrast, requires a more complex setup involving Gram–Schmidt orthogonalization, oversampled least-squares fitting, and blending tapering. These additional components provide greater robustness but introduce more parameters such as blend width, oversampling factor, and zero region size, that require careful tuning to optimize derivative accuracy and error confinement. However, after extensive tuning, we have found good parameters for most settings so that \textsc{FC--Gram} can be used with minimal adjustment (see \Cref{appx: Tuninf FC Gram}).\\

\subsubsection{Computational Cost of the Extension Matrices}  

\vspace{1mm}

\textsc{FC--Gram} generally requires higher-precision arithmetic than \textsc{FC--Legendre} due to fundamental differences in the polynomial basis construction. It builds its orthonormal basis numerically via a QR decomposition of Vandermonde-like matrices followed by least-squares projections. While this offers flexibility and localization, it is sensitive to rounding errors: poor scaling, near-dependent basis functions, and small matching intervals can degrade orthogonality, reducing spectral differentiation accuracy and increasing floating-point noise. Additionally, the matrix construction in \textsc{FC--Gram} is more computationally intensive than the analytic evaluation of Legendre polynomials in \textsc{FC--Legendre}. The latter benefits from analytically defined orthonormal polynomials computed with stable, optimized routines, resulting in better conditioning and less sensitivity to rounding errors. Both methods face conditioning challenges as $d$ grows, but \textsc{FC--Gram} is especially prone to amplified round-off errors. Consequently, although \textsc{FC--Gram} often offers greater noise robustness, it can require precision beyond \texttt{float128} to ensure accurate, stable spectral continuation. 

We emphasize however, that these computational costs and precision requirements only apply for the construction of the extension matrices. For both \textsc{FC--Legendre} and \textsc{FC--Gram}, these matrices can be precomputed offline, and then reused in \textsc{FC--PINO} without the need for recomputation at a similar computation. Once precomputed, the costs of applying \textsc{FC--Legendre} and \textsc{FC--Gram} are very similar, and higher numerical precision is not necessary, unless solutions are desired with much higher precision.

\vspace{5mm}

\subsubsection{Conditioning}

\vspace{1mm}

In \textsc{FC--Gram}, the continuation is built from two smaller QR decompositions on \( d \times d \) Vandermonde-like matrices, which tend to be well-conditioned. This modular approach ensures that the continuation matrix changes smoothly when parameters such as \( c \) or \( d \) vary. In contrast, \textsc{FC--Legendre} relies on a single large pseudoinverse computation involving a \( 2d \times 2d \) Legendre matrix, which is generally less well-conditioned. The near-singular nature of this larger inversion can amplify noise and round-off errors from boundary data, potentially degrading numerical stability. Thus, while \textsc{FC--Gram} may require higher precision in arithmetic, its construction benefits from better conditioning at the matrix level, reducing error magnification during continuation.

\vspace{5mm}

\subsubsection{Convergence of the derivatives}

\vspace{1mm}

Let $u_\theta$ denote the output of the PINO model, $\widetilde u_\theta$ its FC extension, and $u$ the true solution. \\

\noindent In \FCPINO{}, spectral differentiation is performed on the continuation $\widetilde u_\theta$, so the error made on the derivatives can be decomposed as
\[
\|\partial^k \widetilde u_\theta - \partial^k u \|
\ \ \le \ \ 
 \underbrace{\  \| \partial^k \widetilde u_\theta - \partial^k u_\theta \| \ }_{\text{FC error}}
\ \ + \ \ \underbrace{\ \|\partial^k u_\theta - \partial^k u \| \ }_{\text{PINO error}} .
\]

For \textsc{FC Gram}, the continuation error on the physical interval consists of two parts \citep{LyonThesis,lyon2010high}. The dominant term is the interpolation error from matching $d$ boundary values, which converges at order $\mathcal{O}(N^{-d})$. In addition, there is a blending term caused by forcing the continuation to vanish smoothly outside the interval. This blending error does not formally converge with $N$, but can be made numerically negligible. When computing a derivative of order $k$, the interpolation error is reduced to order $\mathcal{O}(N^{-(d-k)})$, while the blending error remains bounded and does not affect the overall convergence rate. 

For \textsc{FC--Legendre}, the extension matrix maps the boundary vector $y=(f_r^\top,f_\ell^\top)^\top\in\mathbb{R}^{2d}$ to the polynomial of degree less than $2d$ fitting these $2d$ values. The approximation order on the physical interval is limited by the boundary strip width $d$, since each side provides at most $d$ independent pieces of information, yielding an interpolation error $\mathcal{O}(N^{-d})$, with $k$ derivatives reducing it to $\mathcal{O}(N^{-(d-k)})$. The high polynomial degree ensures a smooth continuation, while SVD or pseudoinverse residuals are typically negligible.

Now, \FCPINO{} is based on the Fourier neural operator, which enjoys universal approximation for sufficiently smooth operator mappings \citep{Kovachki2021_2}, i.e., one can make the approximation error $\|u_\theta - \mathcal{G}(f)\|$ arbitrarily small. By the Sobolev embedding theorem, with sufficient regularity, the error in the derivatives can also be controlled and kept arbitrarily small. 

Combining this with the FC error estimates yields a $\mathcal{O}(N^{-(d-k)})$ rate for \FCPINO{}. Thus, \FCPINO{} inherits the algebraic convergence rate of FC-based spectral differentiation together with the universal approximation guarantees of PINO.

\vspace{2mm}

\subsubsection{Summary} 

\textsc{FC--Legendre} is particularly well-suited for problems with very smooth, noise-free signals and when higher-order derivatives are required. Its global continuation using Legendre polynomials ensures rapid decay of the continuation error, which remains dominant even after amplification by \(m\)-th order spectral derivatives involving factors \((ik)^m\). This global polynomial fit provides uniform approximation of all spectral modes without artificial damping, enabling high-precision derivative computation in the interior domain. Although \textsc{FC--Legendre} benefits from a simple formulation based on analytic polynomials and a single pseudoinverse which makes implementation and tuning more accessible, careful parameter choice is essential to prevent ill-conditioning and oscillatory artifacts that may emerge as the polynomial degree increases. 

In contrast, \textsc{FC--Gram} excels for signals that are less smooth, noisy, or contain discontinuities, as it provides greater robustness and effective localized error control. By constructing localized orthonormal polynomial bases on each boundary and smoothly blending the extension, it confines boundary errors and high-frequency noise within the extension region, preventing their propagation into the main domain. This approach suppresses spurious oscillations near boundaries, resulting in cleaner derivatives without the need for manual filtering. Although \textsc{FC--Gram} is more computationally intensive and requires higher precision arithmetic, its QR factorizations performed on smaller local matrices produce better-conditioned continuation operators, which improve numerical stability and reduce sensitivity to round-off errors. Overall, it sacrifices some asymptotic accuracy to gain robustness and localized error confinement, making it especially well-suited for noisy or non-smooth problems.

Together, these two Fourier continuation methods offer complementary tools for spectral continuation, enabling users to select the approach best matched to their problem’s smoothness, noise level, derivative order, and computational constraints.

\vspace{8mm}

\section{Hyperparameters}
\label{appx:hyperparameters}

\noindent We first do a grid search over the number of Fourier modes, learning rate, scheduler, and degree of polynomials that both FC functions are interpolating with. For \textsc{FC--Gram}, we fix the number of continuation points to 50, and for \textsc{FC--Legendre}, we additionally search over continuation points. Below, we report the models with the best accuracy.

\clearpage

\subsection{1D Self-Similar Burger's Equation}
\label{appx:hyper_1d_burgers}

We use a 1D FNO with 4 layers, each with 200 hidden channels. We train for 60,000 epochs minimizing the MSE using Adam and the ReduceLROnPlateau scheduler with factor 0.5. We conducted a grid search over the loss coefficients, and fixed them to $1$, $1$, and $0.1$ for the PDE residual, boundary, and smoothness losses, respectively. For the fine-tuning phase of \Cref{sec: learn family Burgers 1D}, we increase the smoothness loss coefficient to $1$. We then conducted a grid search over numbers of Fourier modes of $\{24, 32, 48, 100\}$, learning rates of $\{0.001, 0.005, 0.0001\}$, and patience values of $\{500, 1000, 2000\}$. For \textsc{FC--Gram}, we searched over $d$ values in $\{2, 3, 4, 5, 6, 7, 8\}$, and for \textsc{FC--Legendre} we searched over $c$ values in $\{20, 30, 40, 50, 60, 70, 80\}$ and $d$ values in $\{2, 3, 4, 5, 6, 7, 8\}$. The configurations reported in \Cref{tab: 1D Hyperparams} correspond to the best model obtained for each approach.

\begin{table}[h!] 
\centering
\caption{Optimal Hyperparameters for the PINO models for the 1D Burgers equation.} \vspace{-2mm}\label{tab: 1D Hyperparams}
\begin{tabular}{lcccccc}
\toprule
\textbf{Model} & \hspace{3mm} \textbf{FC Method} \hspace{3mm} & \textbf{$d$} & \textbf{$c$} & \textbf{Fourier Modes} & \textbf{Learning Rate} \\
\midrule
\multirow{2}{*}{\FCPINO} 
& \textsc{FC--Gram} & 6 & 50 & 24 & 0.0001 \\
& \textsc{FC--Legendre} & 4 & 70 & 48 & 0.001 \\
\midrule
\multirow{2}{*}{\EndFCPINO} 
& \textsc{FC--Gram} & 4 & 50 & 24 & 0.005 \\
& \textsc{FC--Legendre} & 5 & 70 & 24 & 0.0001 \\
\midrule
\multirow{2}{*}{\outsideFCPINO} 
& \textsc{FC--Gram} & 4 & 50 & 24 & 0.001 \\
& \textsc{FC--Legendre} & 6 & 50 & 48 & 0.005 \\
\midrule
\standardPINO & - & - & - & 32 & 0.0001 \\
\PINOPadOut & - & - & \text{100}  & 48 & 0.0005 \\
\padPINO & - & - & \text{100} & 32 & 0.0001 \\
\bottomrule
\end{tabular}
\end{table}

\hfill 

\subsection{2D Burger's Equation}
\label{appx:hyper_2d_burgers}

We use a 2D FNO with 4 layers, each with 40 hidden channels and $(100,100)$ Fourier modes. We train for 60,000 epochs minimizing the MSE using Adam and the ReduceLROnPlateau scheduler with factor 0.5. We then conducted a grid search over the values of $\{64, 100, 128\}$ for the Fourier modes, $\{0.001, 0.005, 0.0001\}$ for the learning rate, and $\{500, 1000, 2000\}$ for the scheduler patience. For \textsc{FC--Gram}, we searched over $d$ in $\{3, 4, 5, 6, 7, 8\}$, and for \textsc{FC--Legendre}, we searched over $c$ in $\{30, 40, 50, 60, 70, 80\}$ and $d$ in $\{3, 4, 5, 6, 7, 8\}$. Here, we employ \textsc{ReLoBRaLo} to adaptively scale the loss coefficients. The configurations reported in \Cref{tab: 2D Hyperparams} correspond to the best model obtained for each approach.

\begin{table}[h!]
\centering
\caption{Optimal Hyperparameters for the PINO models for the 2D Burgers equation.} \label{tab: 2D Hyperparams} \vspace{-2mm}
\begin{tabular}{lcccc}
\toprule
\textbf{Model} & \hspace{5mm} \textbf{FC Method} \hspace{5mm} & \textbf{$d$} & \textbf{$c$} & \hspace{2mm} \textbf{Learning Rate} \hspace{2mm}  \\
\midrule
\multirow{2}{*}{\FCPINO} 
& \textsc{FC--Gram} & 4 & 50 & 0.0001 \\
& \textsc{FC--Legendre} & 6 & 30 & 0.001 \\
\midrule
\multirow{2}{*}{\EndFCPINO} 
& \textsc{FC--Gram} & 6 & 50 & 0.0001 \\
& \textsc{FC--Legendre} & 6 & 40 & 0.001 \\
\midrule
\multirow{2}{*}{\outsideFCPINO} 
& \textsc{FC--Gram} & 6 &  50& 0.0001 \\
& \textsc{FC--Legendre} & 6 & 80 & 0.001 \\
\midrule
\standardPINO \ \ \ & - & - & - & 0.01 \\
\PINOPadOut & - & - & \text{100} & 0.01 \\
\padPINO & - & - & \text{100} & 0.001 \\
\bottomrule
\end{tabular} \vspace{4mm}
\end{table}

\hfill 

\subsection{Navier--Stokes Equations}
\label{appx:hyper_Navier_Stokes}

\noindent We use a 3D FNO with 4 layers, each with 60 hidden channels. We train for 60,000 epochs with a learning rate of 0.001 minimizing the MSE using Adam and the ReduceLROnPlateau scheduler with factor 0.5. We then conducted a grid search over learning rate in $\{0.001, 0.005, 0.0001\}$, scheduler patience in $\{500, 1000, 2000\}$, and Fourier modes in $\{24,32\}$. For \textsc{FC--Gram}, we searched over $d$ in $\{3, 4, 5, 6\}$, and for \textsc{FC--Legendre}, we searched over $c$ in $\{20, 30, 40, 50, 60\}$ and $d$ in $\{3, 4, 5, 6\}$. Here, we employ \textsc{ReLoBRaLo} to adaptively scale the loss coefficients. The best configurations are reported in \Cref{tab: NS Hyperparams}.

\begin{table}[h!]
\centering
\caption{Optimal Hyperparameters for the PINO models for the 3D Navier--Stokes equation.} 
\label{tab: NS Hyperparams} \vspace{-2mm}
\begin{tabular}{lcccccc}
\toprule
\textbf{Model} & \hspace{2mm} \textbf{FC Method} \hspace{2mm} & \textbf{$d$} & \textbf{$c$} & \textbf{Spatial Modes} & \textbf{Temporal Modes} \\
\midrule
\multirow{2}{*}{\FCPINO} 
& \textsc{FC--Gram} & 8 & 50 & 32 & 20 \\
& \textsc{FC--Legendre} & 7 & 40 & 32 & 20 \\
\midrule
\multirow{2}{*}{\EndFCPINO} 
& \textsc{FC--Gram} & 7 & 50 & 32 & 24 \\
& \textsc{FC--Legendre} & 8 & 60 & 32 & 24 \\
\midrule
\multirow{2}{*}{\outsideFCPINO} 
& \textsc{FC--Gram} & 8 & 50 & 32 & 20 \\
& \textsc{FC--Legendre} & 8 & 60 & 32 & 24 \\
\midrule
\standardPINO \ \ & - & - & - & 24 & 20 \\
\PINOPadOut & - & - & \text{50} & 24 & 20 \\
\padPINO & - & - & \text{50} & 32 & 24 \\
\bottomrule
\end{tabular}
\end{table}

\clearpage

\section{Solution Visualizations}

\label{appx:solution_visualizations}
In this appendix, we display examples of solutions to the PDEs considered in this paper, obtained using various PINO frameworks. 

\subsection{1D Self-Similar Burger's Equation}
\label{appx:sol_1d_burgers}

\subsubsection{Comparison of the Different Approaches on $\lambda = 0.5$}

\vspace{1mm}

\begin{figure}[h]
    \centering
    \includegraphics[width=0.52\linewidth]{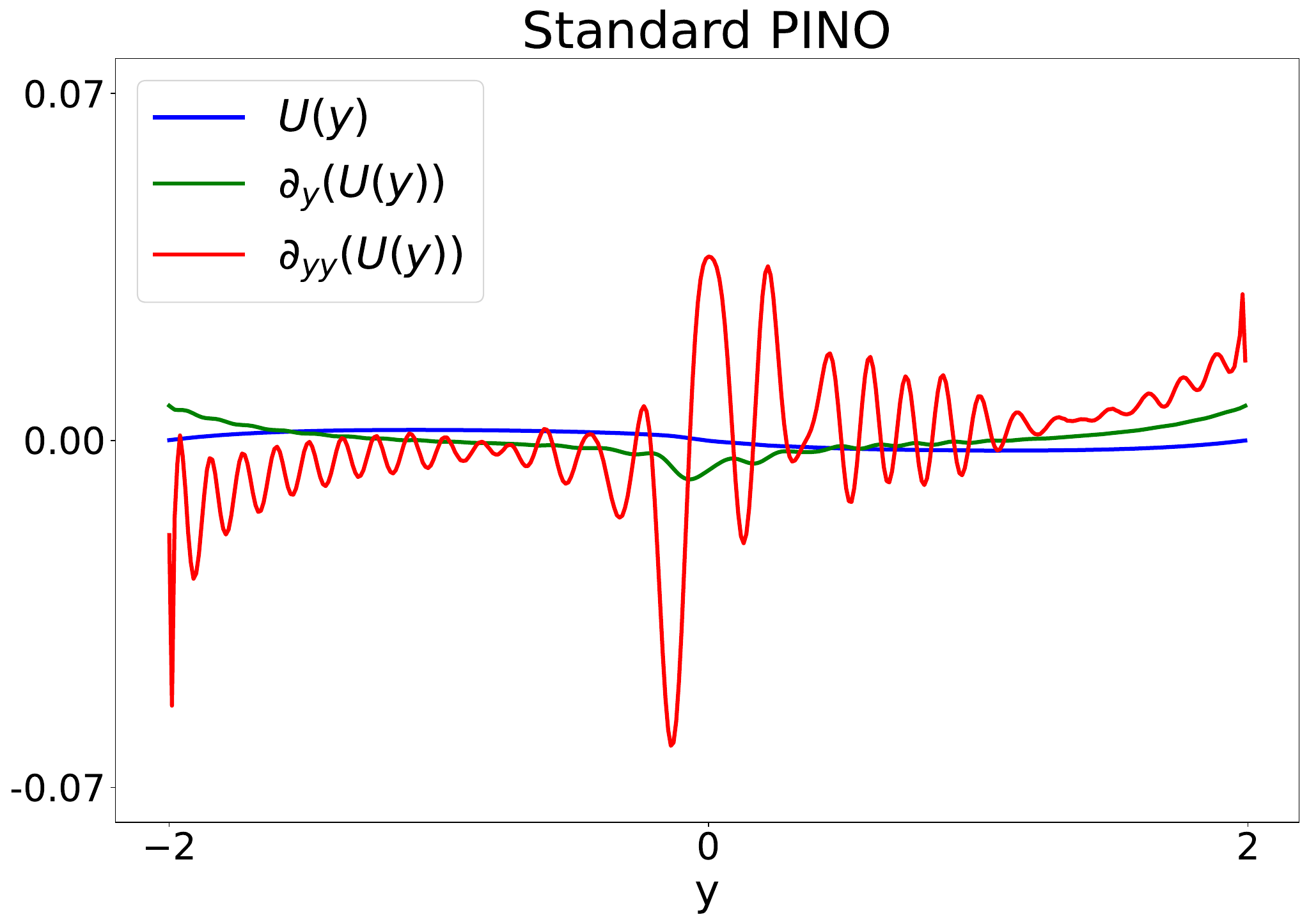}
    \begin{minipage}[t]{0.492\textwidth}
        \centering
        \includegraphics[width=\linewidth]{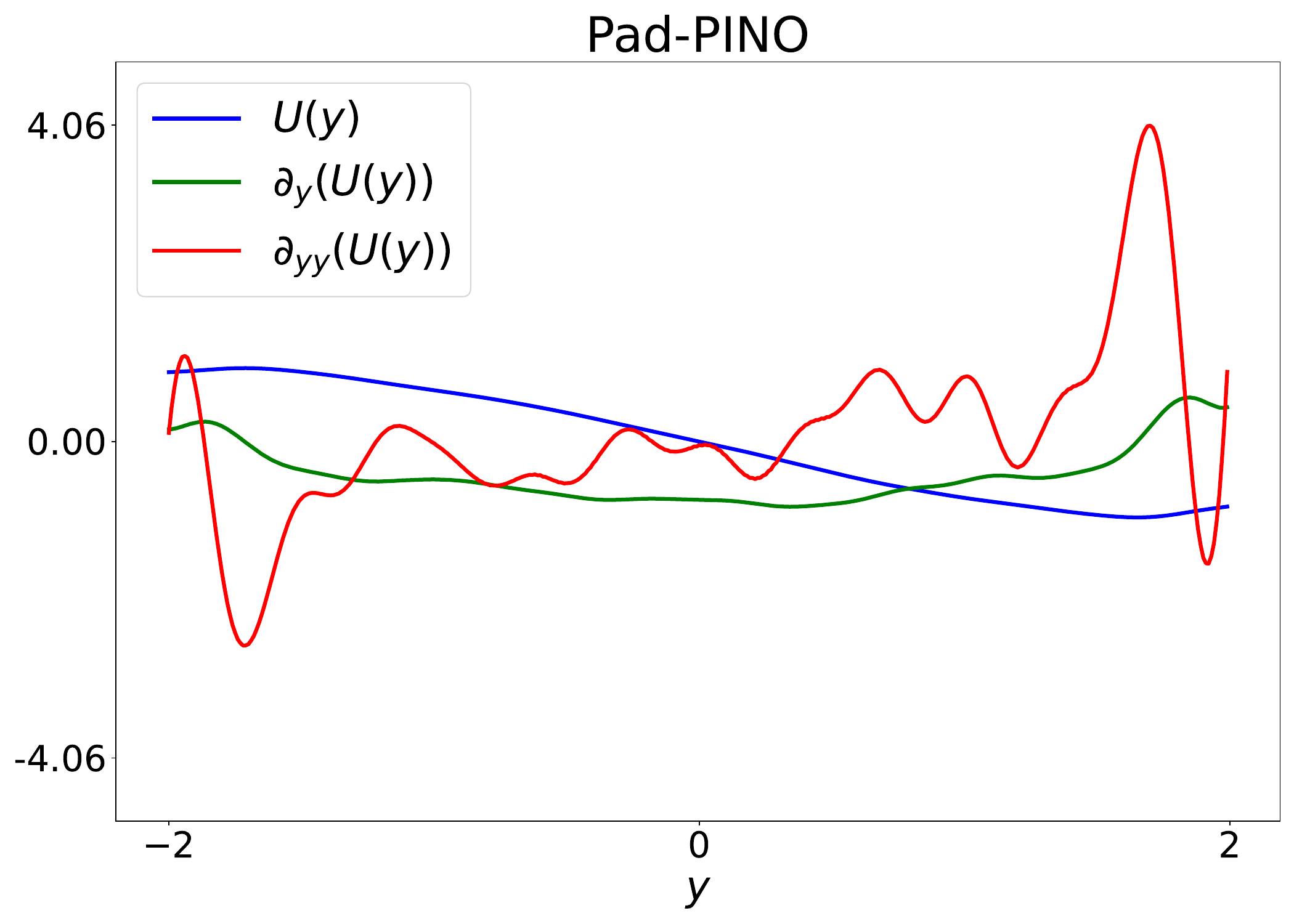}
    \end{minipage}
    \hfill
    \begin{minipage}[t]{0.492\textwidth}
        \centering
        
        \includegraphics[width=\linewidth]{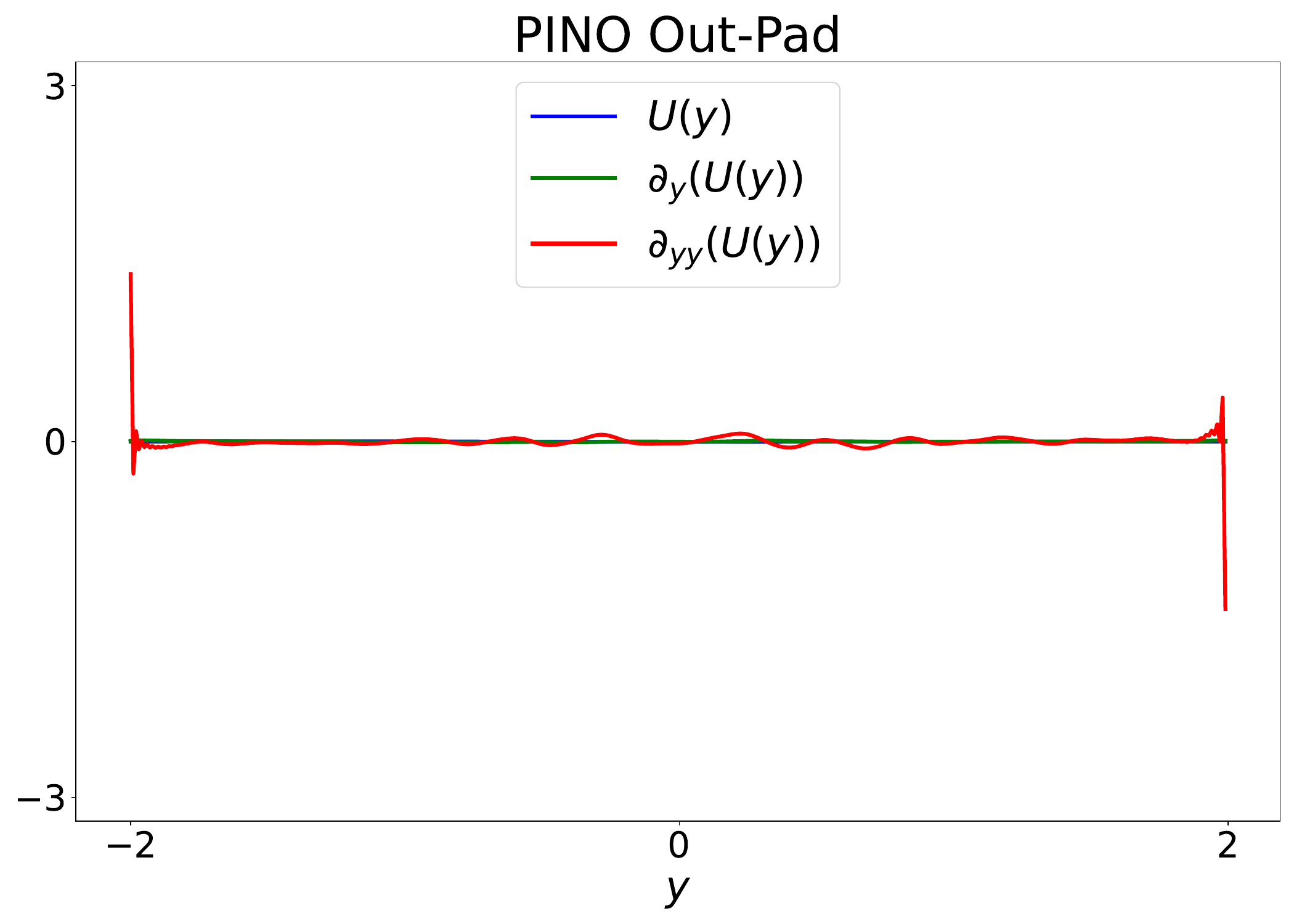}
    \end{minipage}

    \begin{minipage}[t]{0.492\textwidth}
        \centering
        \includegraphics[width=\linewidth]{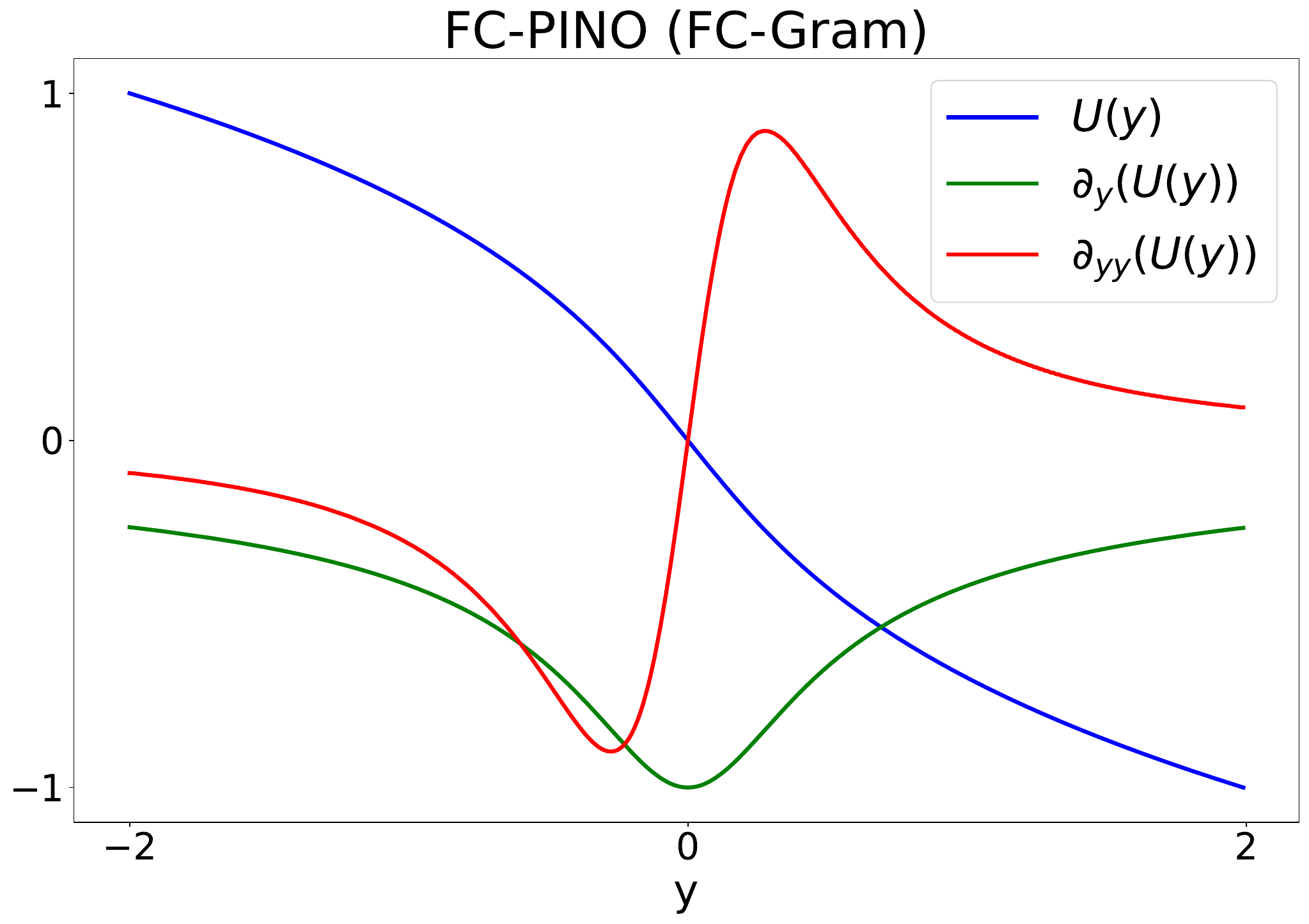}
    \end{minipage}
    \hfill
    \begin{minipage}[t]{0.492\textwidth}
        \centering
        \includegraphics[width=\linewidth]{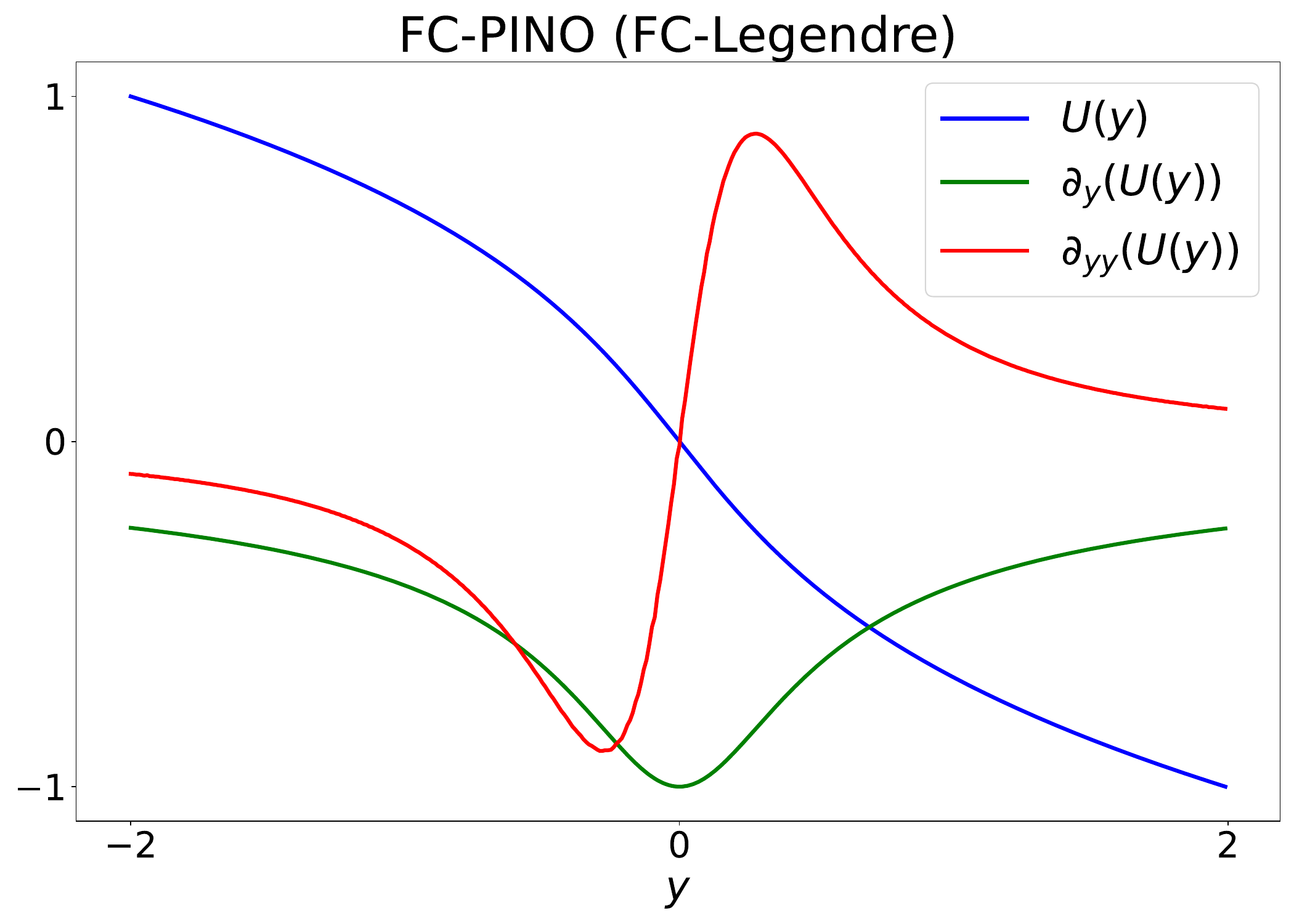}
    \end{minipage}

    \caption{Predicted solutions of PINO architectures for the 1D Burger's equation.}
    \label{fig:1d_burgers_1}
    \vspace{-16mm}
\end{figure}

\clearpage

\subsubsection{Instance-Wise Fine-tuning of the Pretrained \FCPINO{} Models}
\label{appx:fine_tuning_sols}

\vspace{3mm}

\begin{figure}[h] 
    \centering
    \begin{minipage}[t]{0.53\textwidth}
        \centering
        \includegraphics[width=\linewidth]{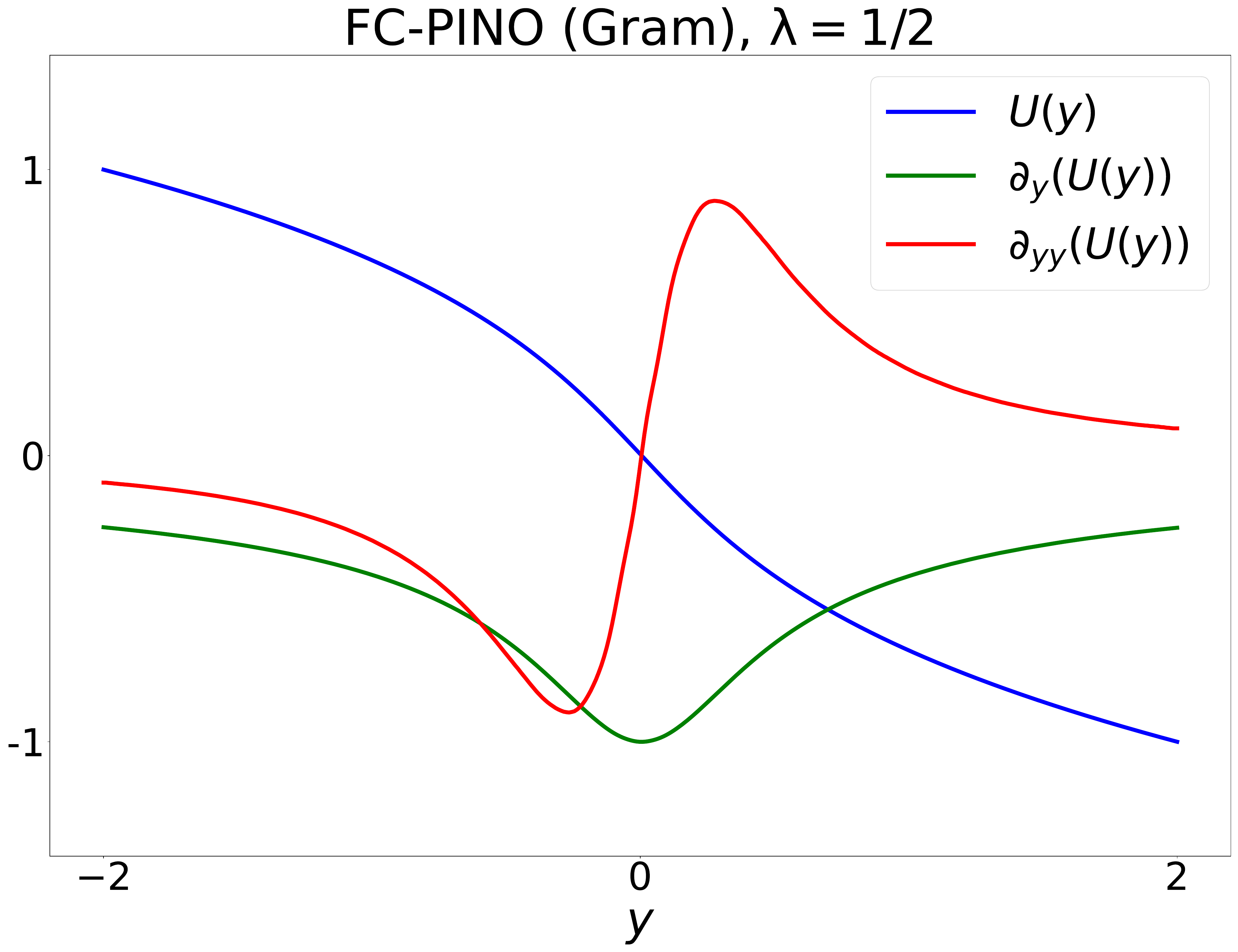}\\
        \includegraphics[width=\linewidth]{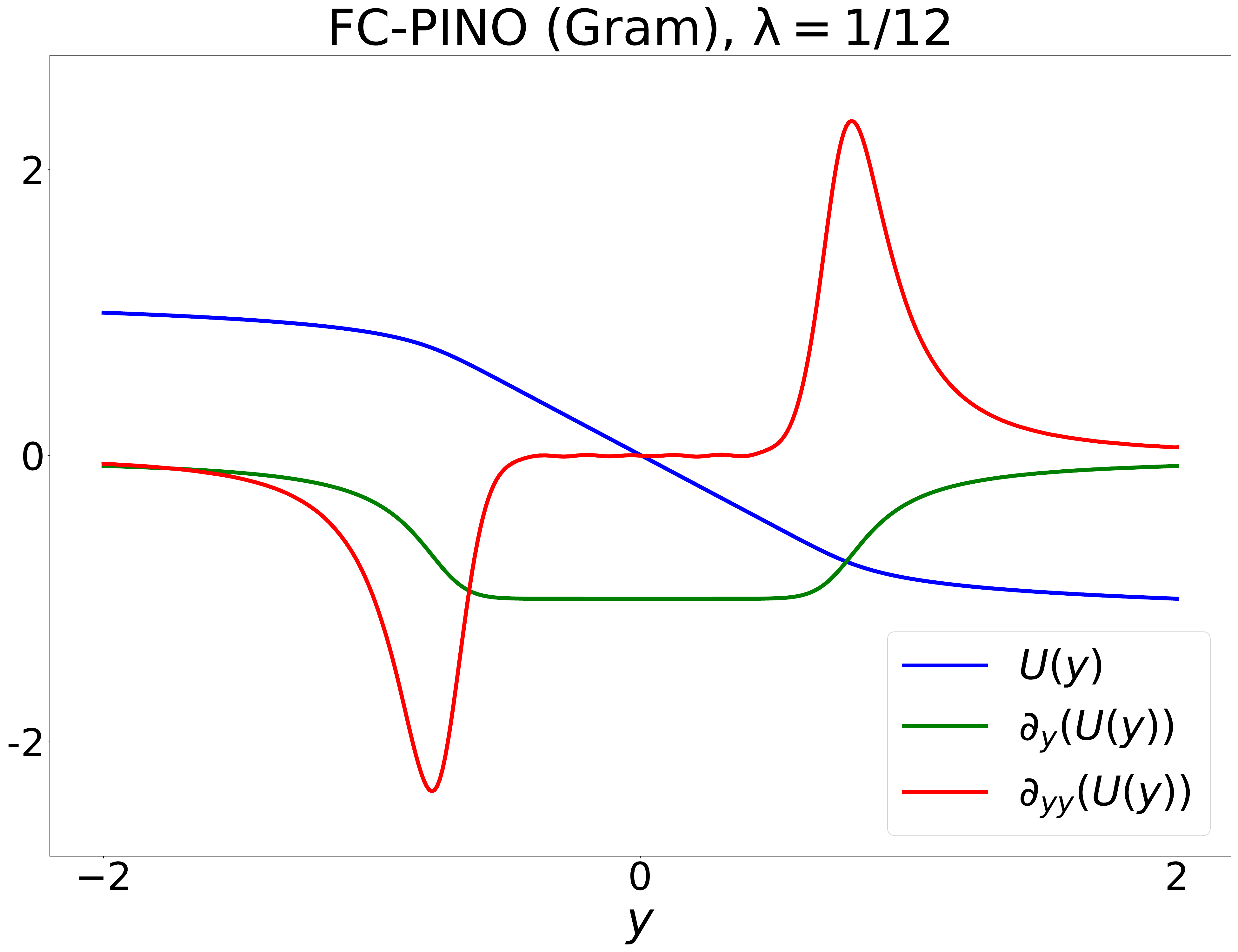}\\
        \includegraphics[width=\linewidth]{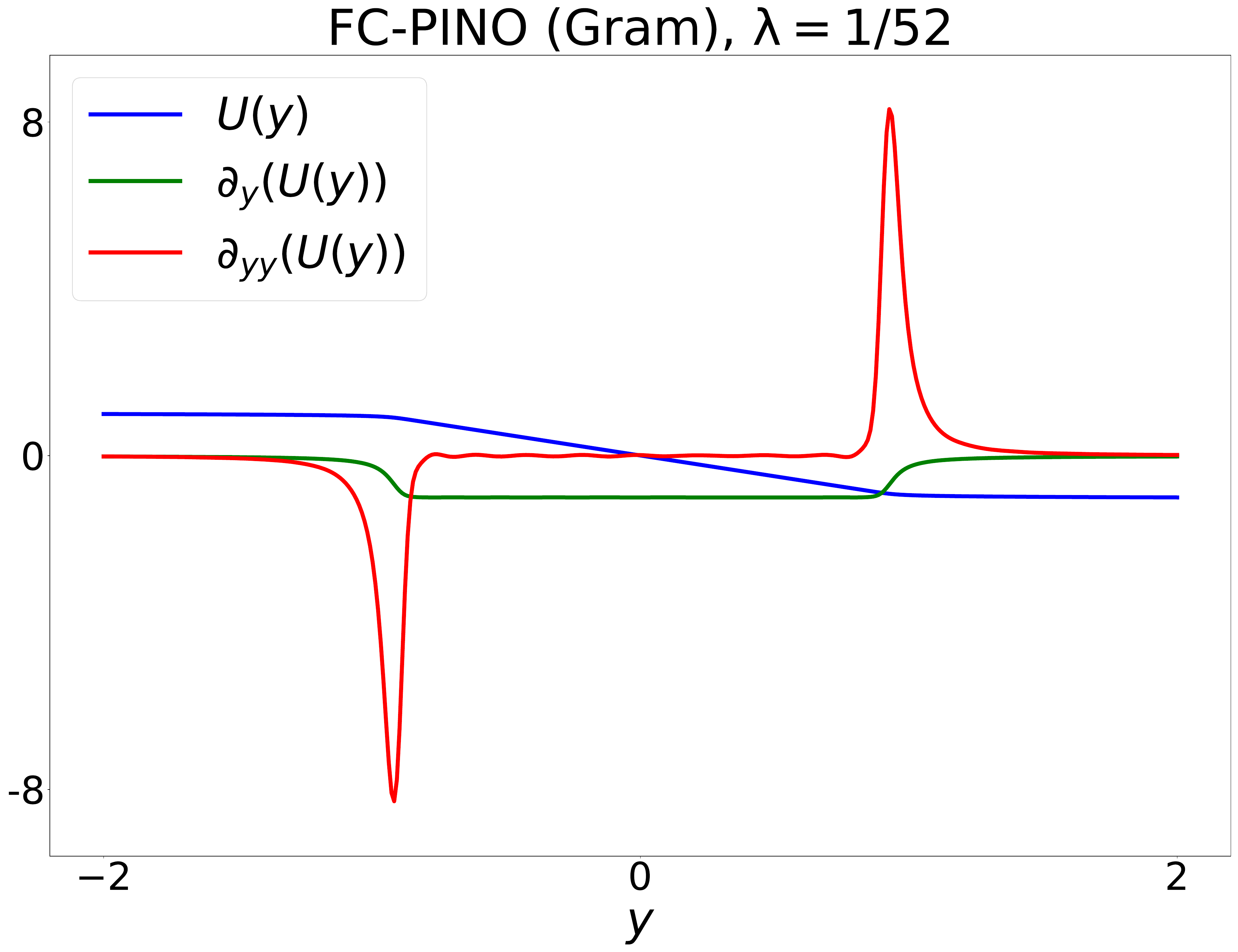}\\
    \end{minipage}

    \caption{Predicted solutions when fine-tuning pretrained \FCPINO{} models on different $\lambda$ values for the 1D Burgers' equation.}
    \label{fig:fine_tuning_1a} \vspace{-15mm}
\end{figure}

\clearpage

\hfill 
\vspace{-22mm}

\subsection{2D Burgers Equation} \label{appx:sol_2D_burgers}

\begin{figure}[h]
\vspace{-3mm}
    \centering
    \begin{minipage}[t]{0.325\textwidth}
        \centering
        \includegraphics[width=\linewidth]{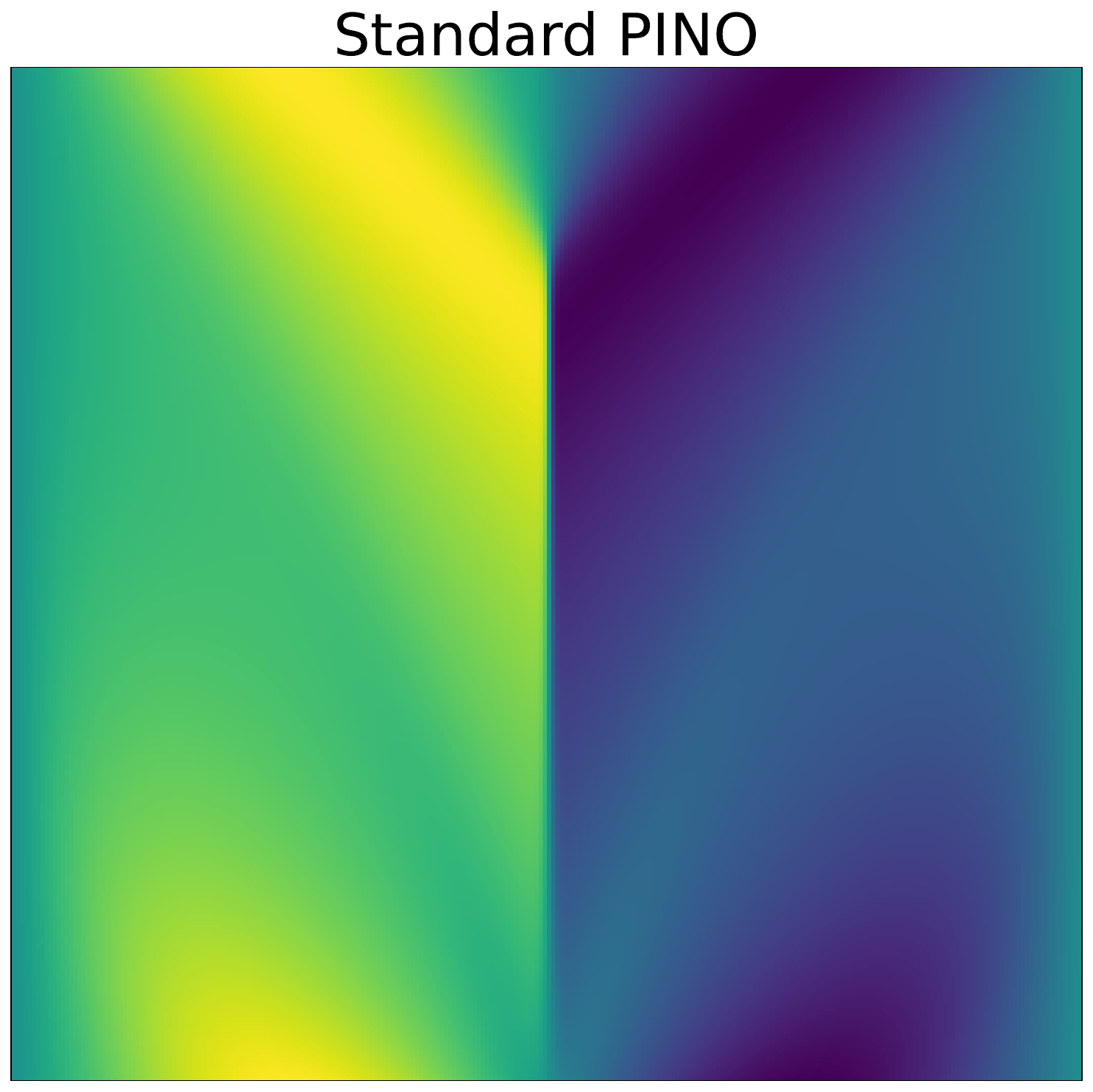}
    \end{minipage}
    \hfill
    \begin{minipage}[t]{0.325\textwidth}
        \centering
\includegraphics[width=\linewidth]{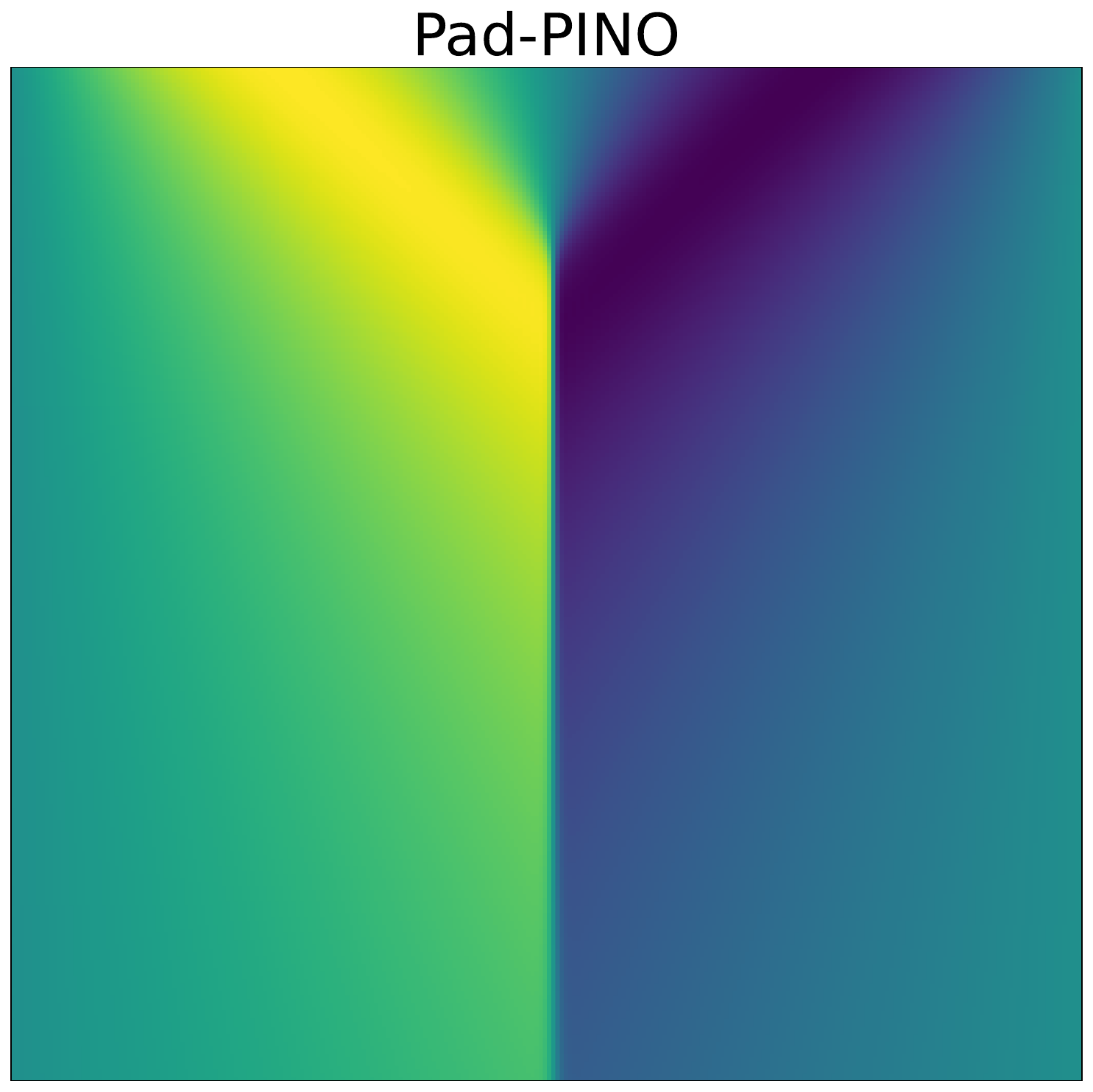}
    \end{minipage} \hfill
     \begin{minipage}[t]{0.325\textwidth}\includegraphics[width=\linewidth]{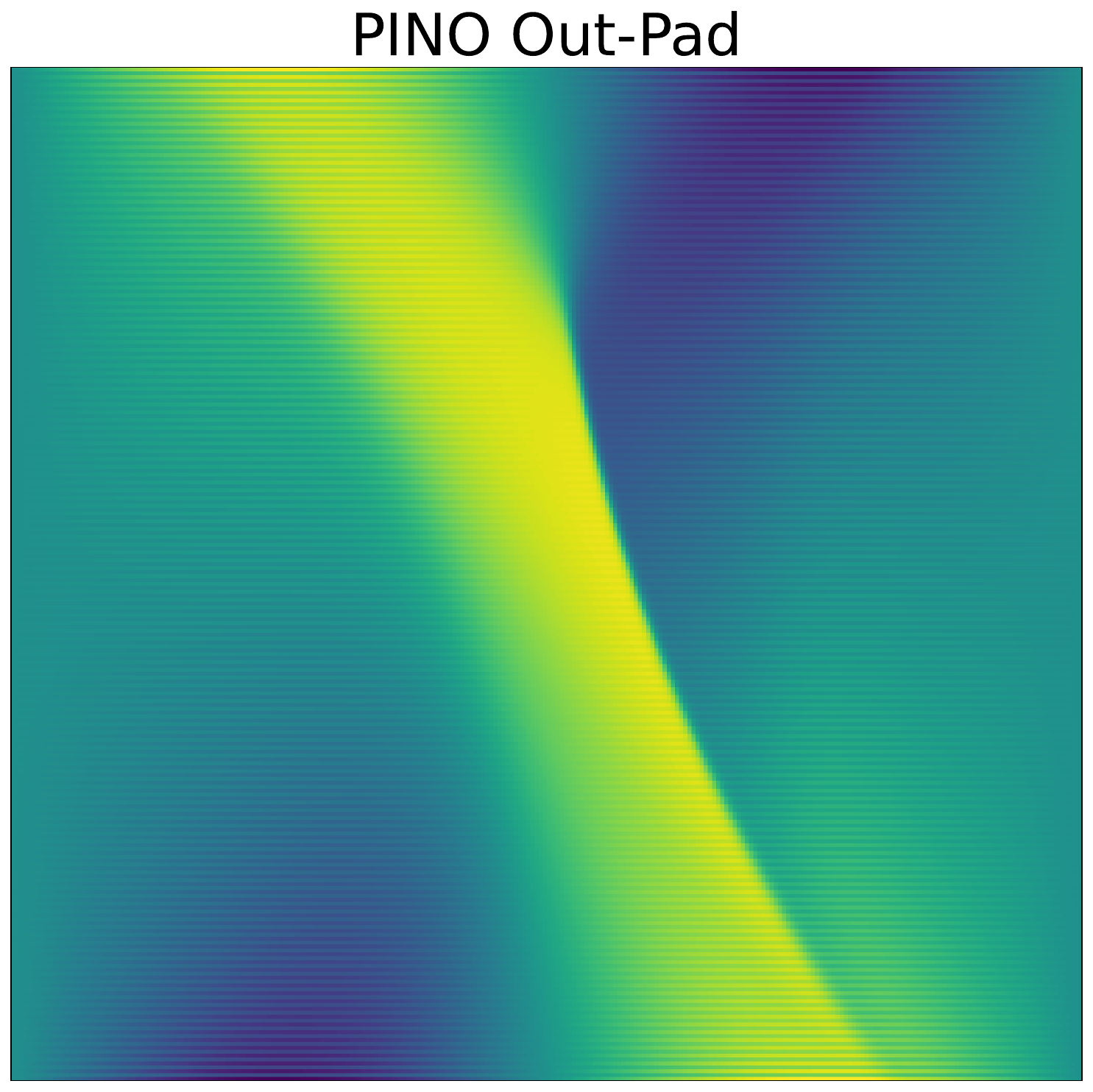}     \end{minipage}\\
\hspace{24mm}  
     \begin{minipage}[t]{0.325\textwidth}\includegraphics[width=\linewidth]{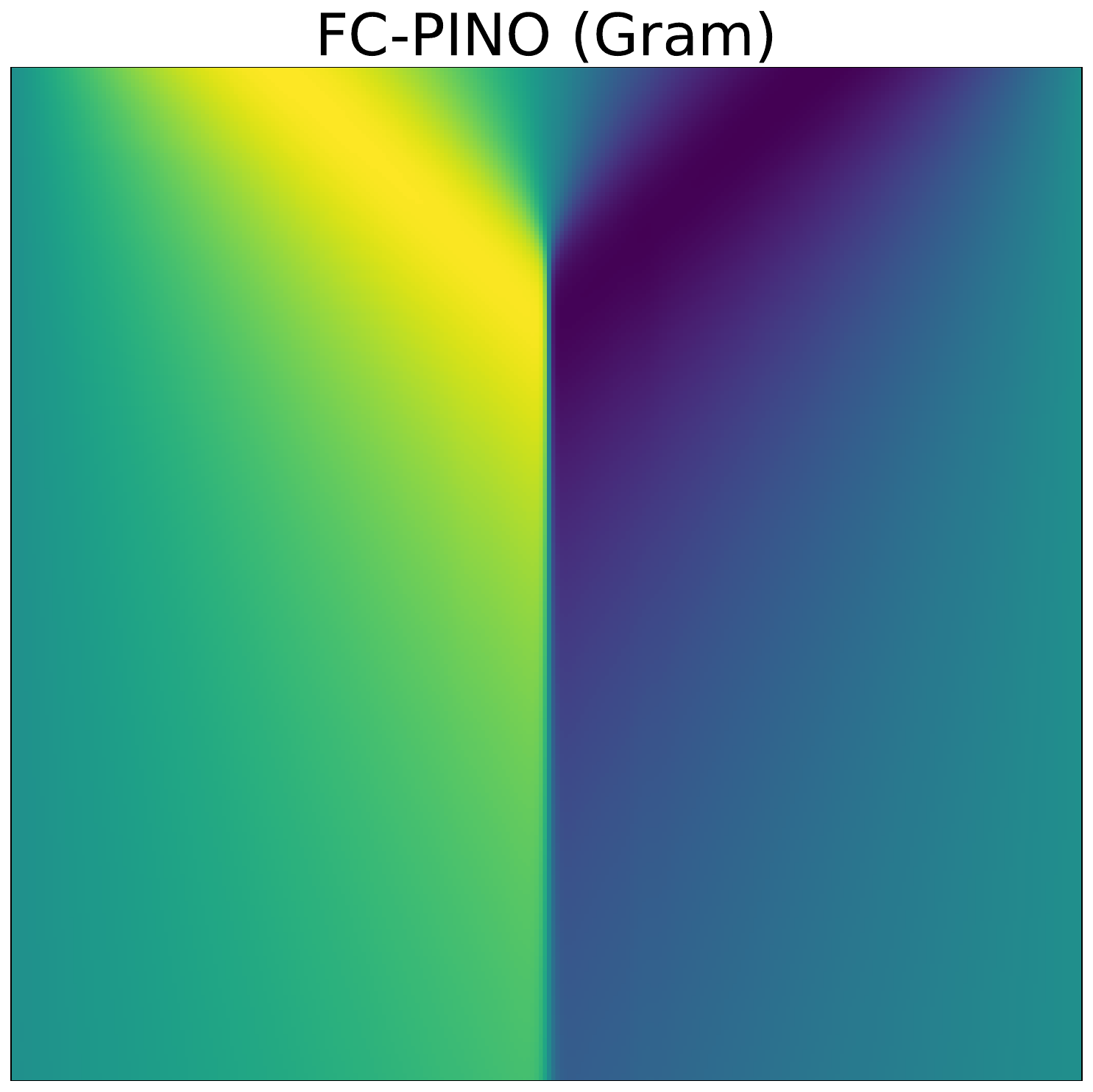}   \end{minipage}
\hfill
      \begin{minipage}[t]{0.325\textwidth}\includegraphics[width=\linewidth]{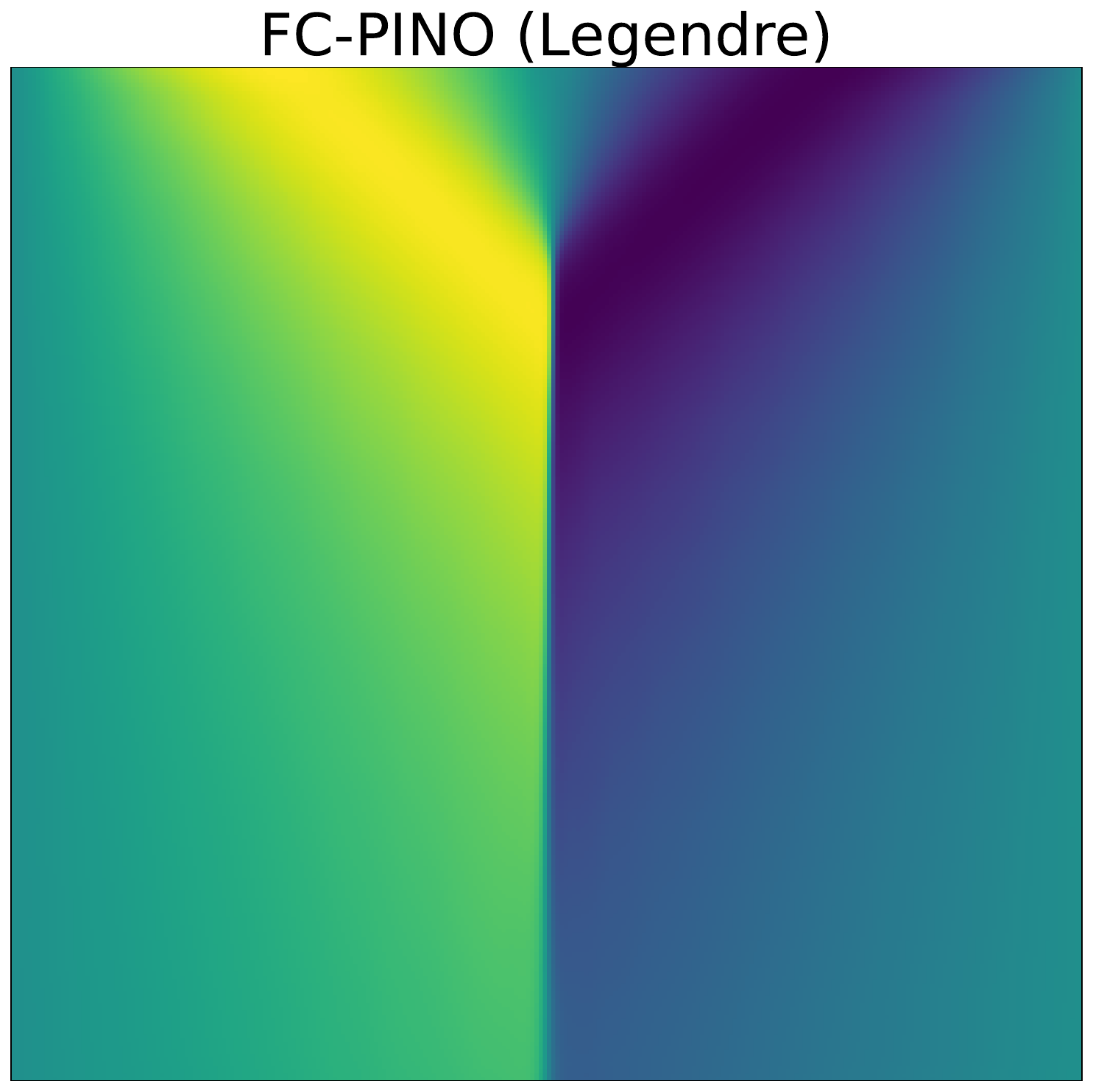}     \end{minipage} \hspace{24mm} 
      
    \vspace{2mm}
        
    \centering
    \begin{minipage}[t]{0.325\textwidth}
        \centering
        \includegraphics[width=\linewidth]{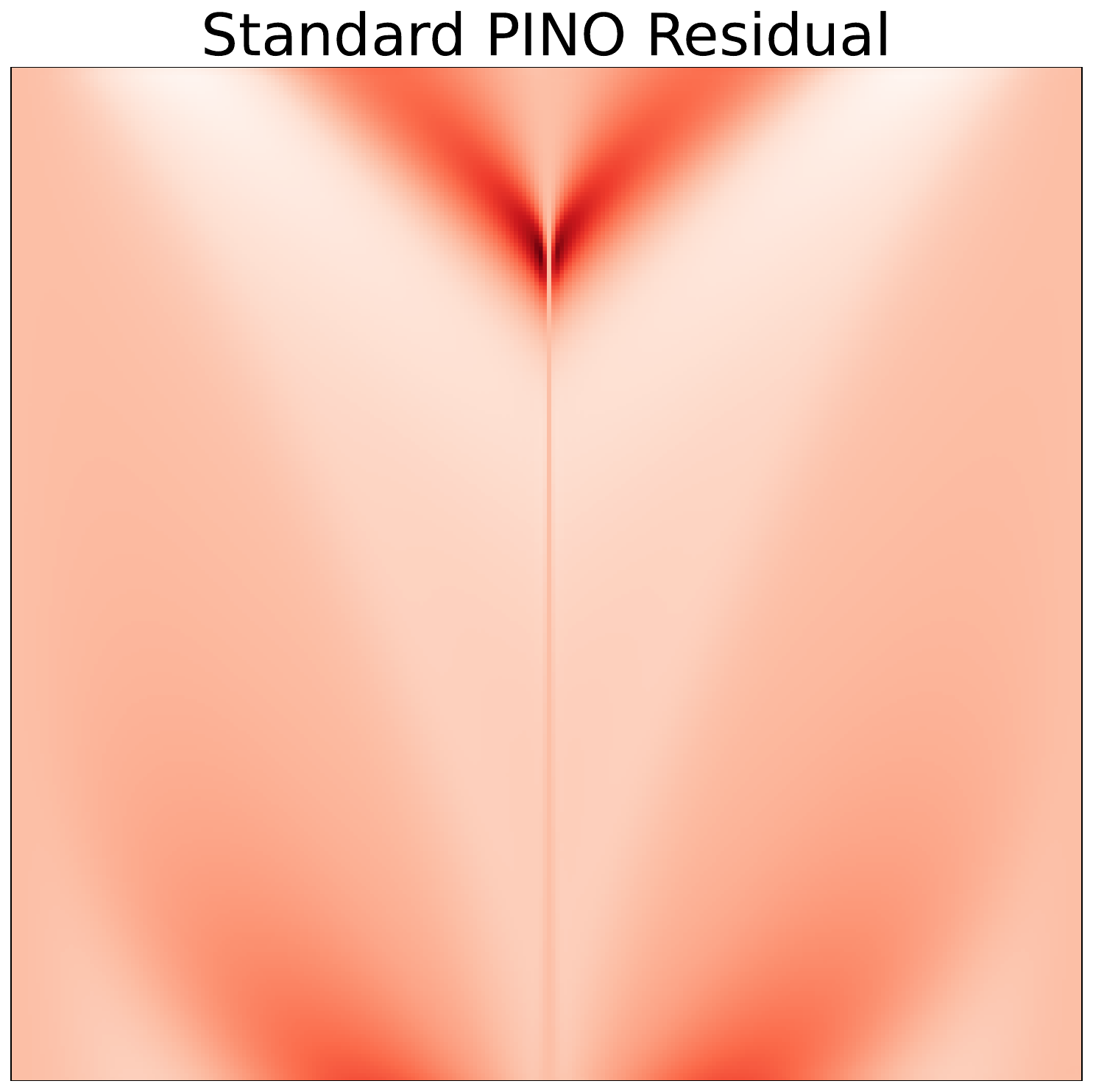}
    \end{minipage}
    \hfill
    \begin{minipage}[t]{0.325\textwidth}
        \centering
\includegraphics[width=\linewidth]{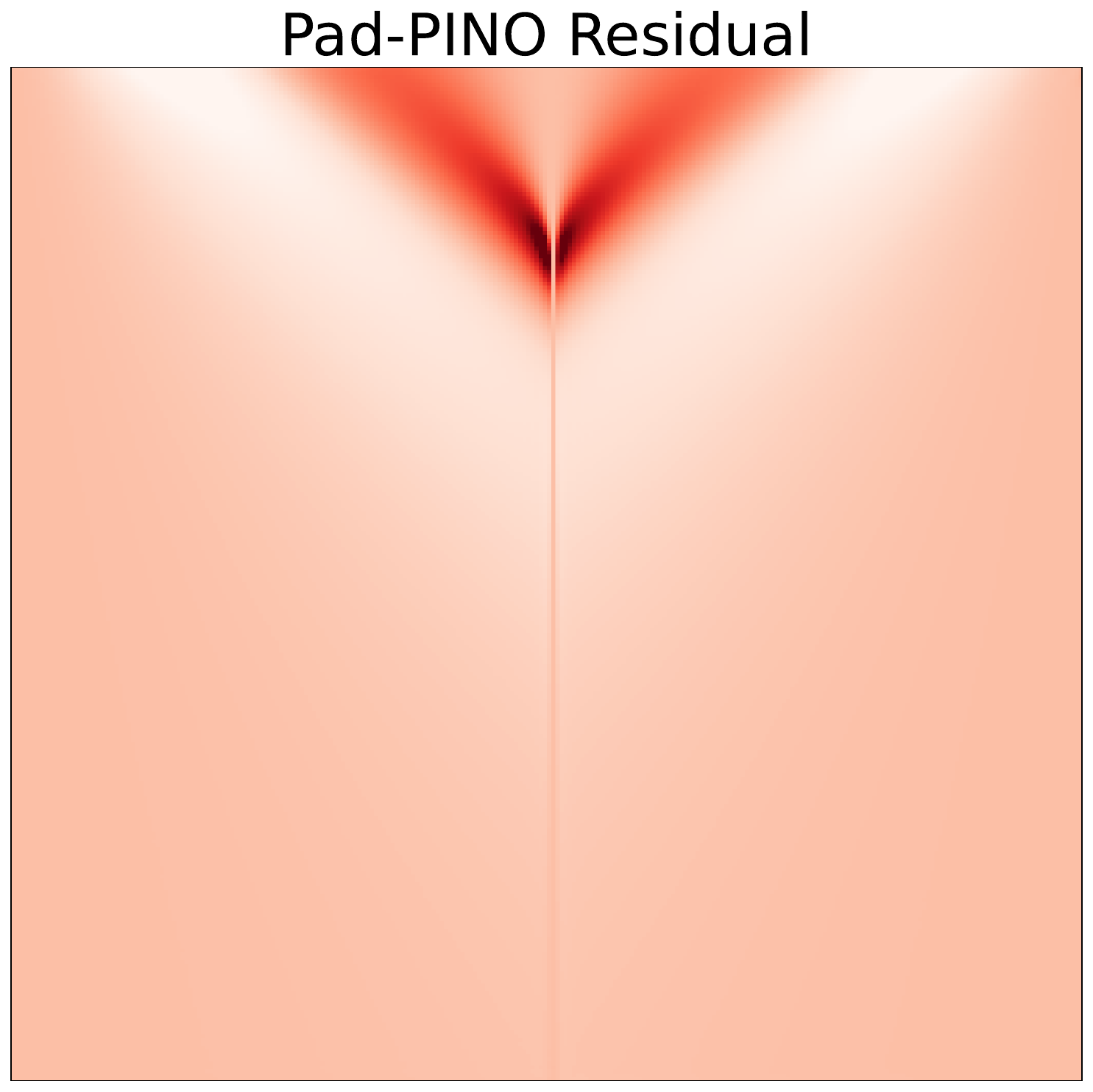}
    \end{minipage} \hfill
     \begin{minipage}[t]{0.325\textwidth}\includegraphics[width=\linewidth]{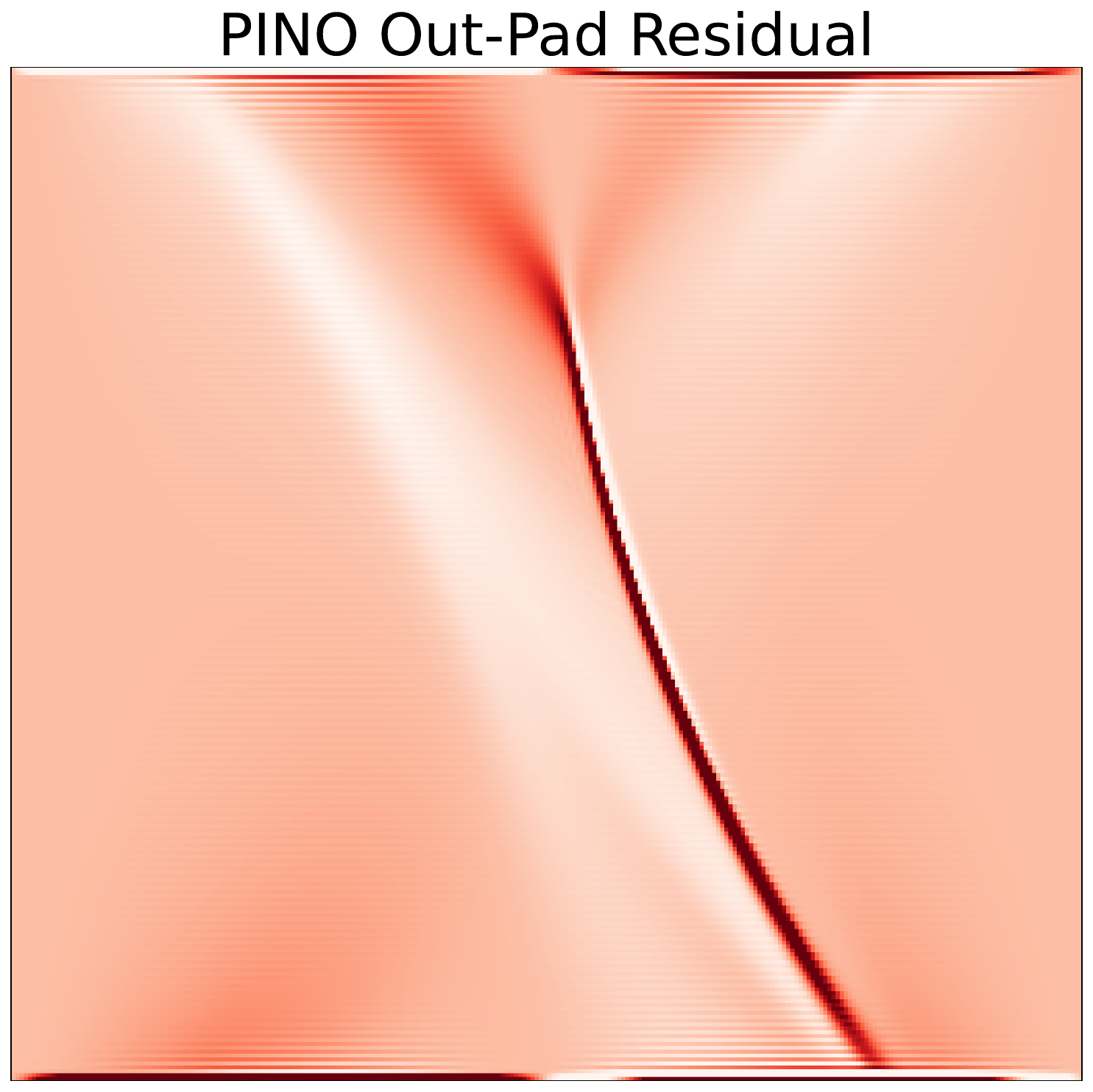}   \end{minipage}\\
\hspace{16mm}  
     \begin{minipage}[t]{0.325\textwidth}\includegraphics[width=\linewidth]{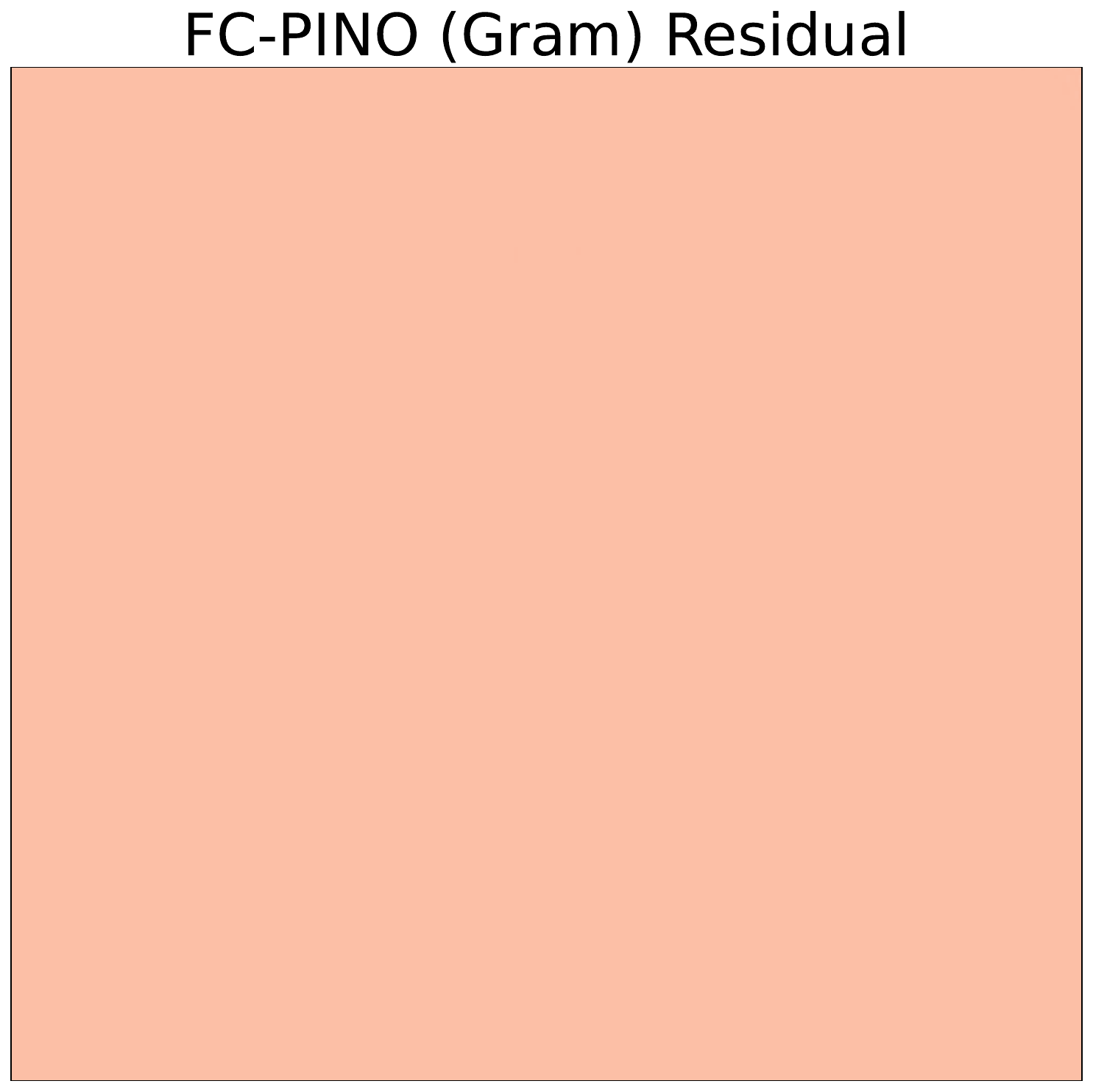} \end{minipage}
\hspace{0.5mm}
      \begin{minipage}[t]{0.325\textwidth}\includegraphics[width=\linewidth]{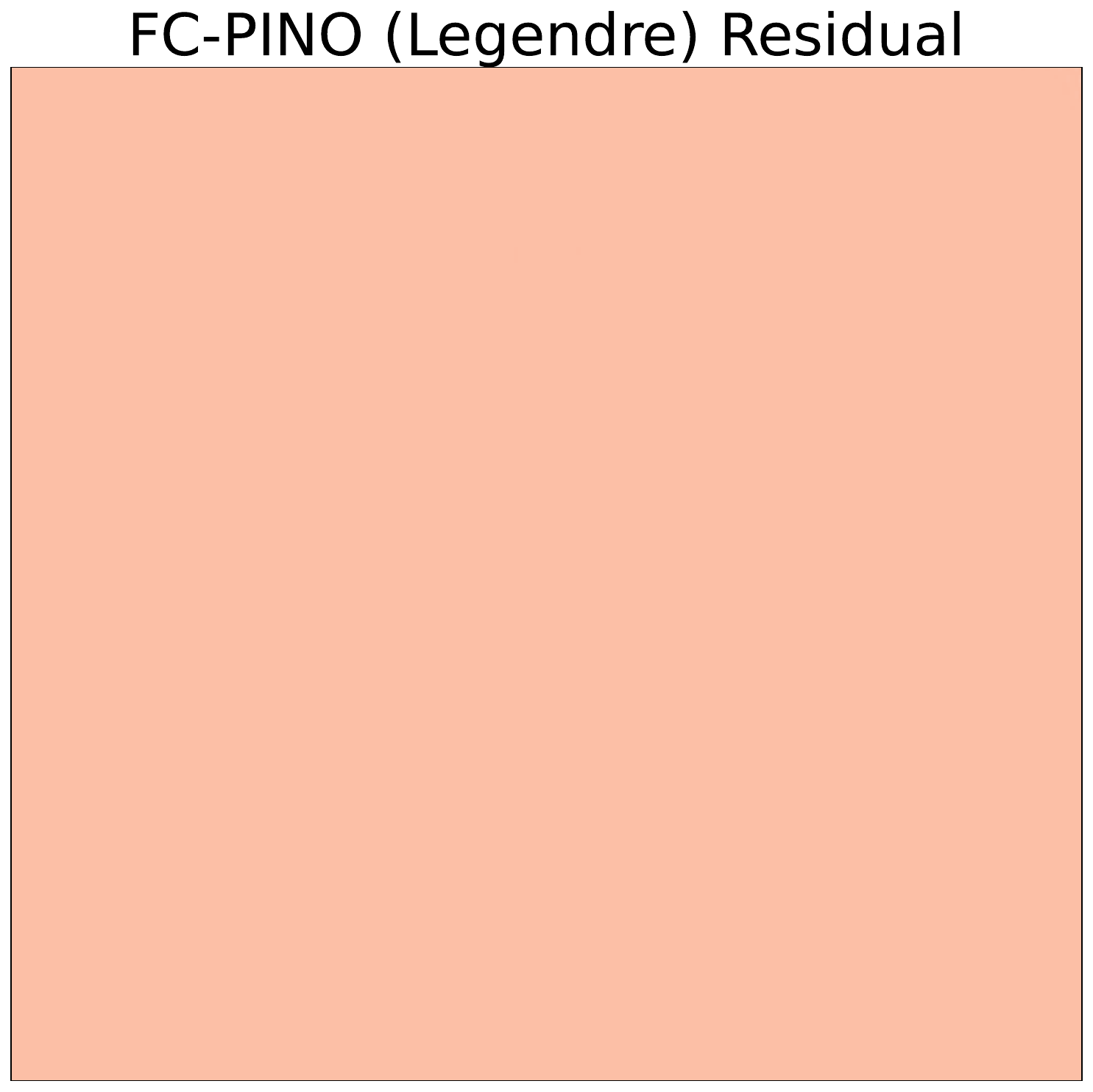}  \end{minipage} 
      \hspace{4mm} 
       \begin{minipage}[t]{0.066\textwidth}  \hfill \vspace{-48mm} \includegraphics[width=\linewidth]{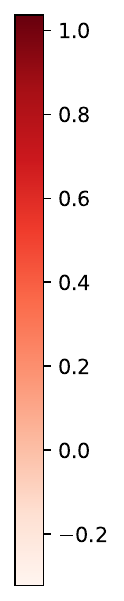}
    \end{minipage}   \hspace{26mm}
      
           \vspace{-5mm}  \caption{Predicted solutions (\emph{top five}) and corresponding PDE residuals (\emph{bottom five}) for baseline PINO architectures and \FCPINO{} variants for the 2D Burgers’ equation.}\label{fig:other_gram2D}\label{fig:2d_burgers}  \vspace{-16mm}
\end{figure}

\clearpage

\hfill 
\vspace{-21mm}

\subsection{Navier--Stokes Equations}
\label{appx:sol_Navier_Stokes}

\vspace{20mm}

\begin{figure}[h]
    \centering
    \vspace{-22.5mm}
    \begin{minipage}[t]{0.325\textwidth}
        \centering
        \includegraphics[width=\linewidth]{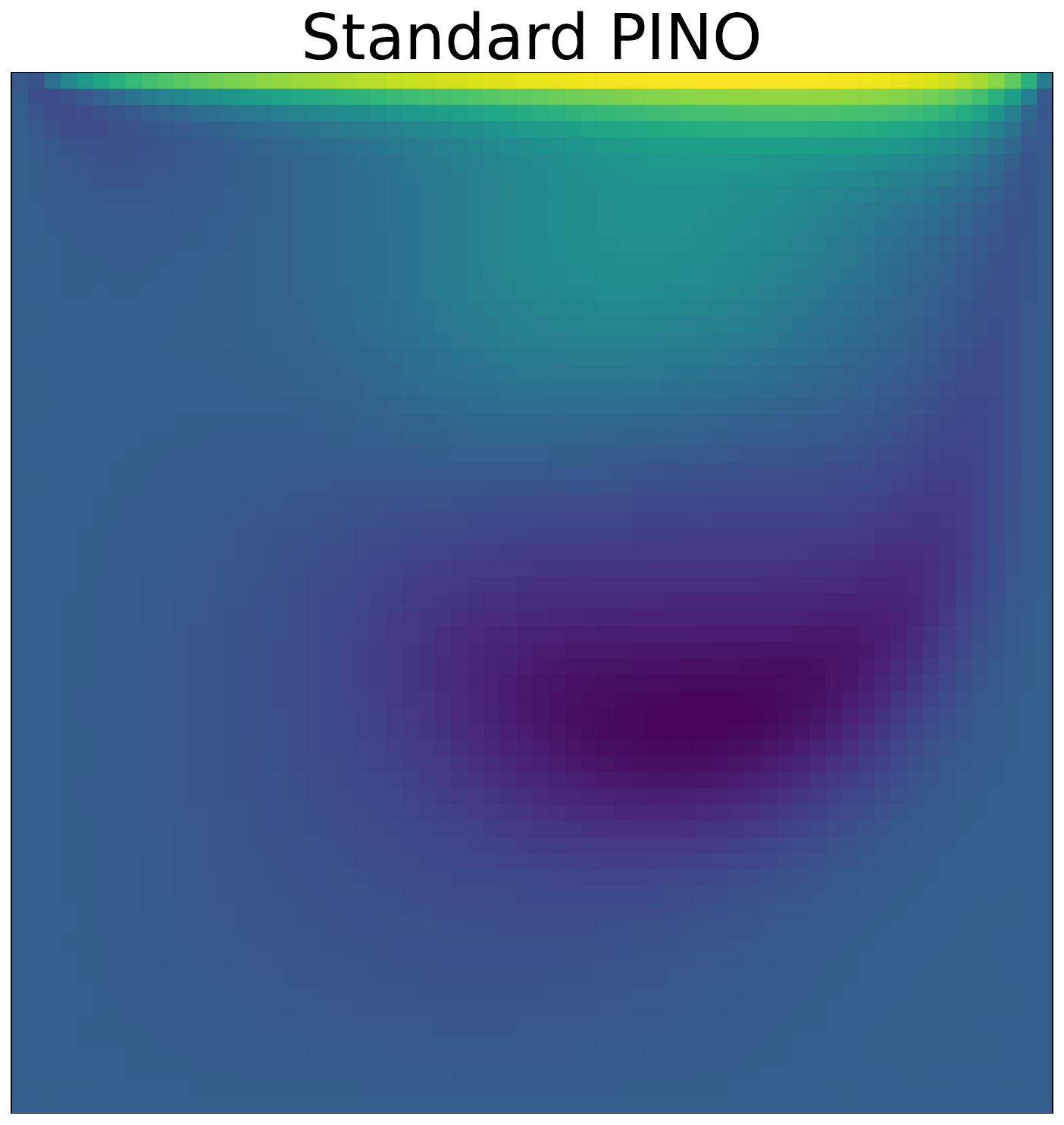}
    \end{minipage}
    \hfill
    \begin{minipage}[t]{0.325\textwidth}
        \centering
        \includegraphics[width=\linewidth]{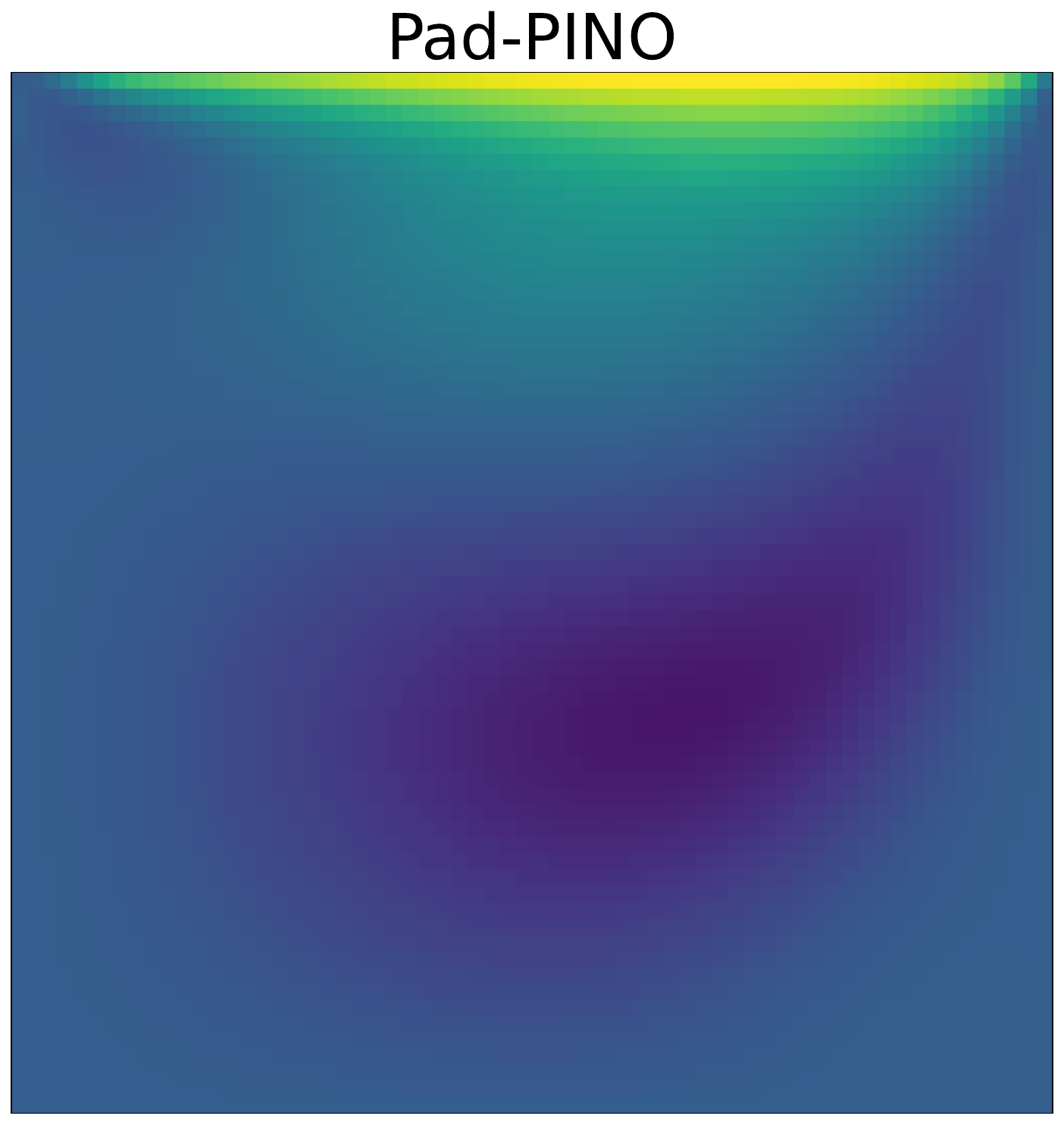}
    \end{minipage}
    \hfill
    \begin{minipage}[t]{0.325\textwidth}
        \centering
        \includegraphics[width=\linewidth]{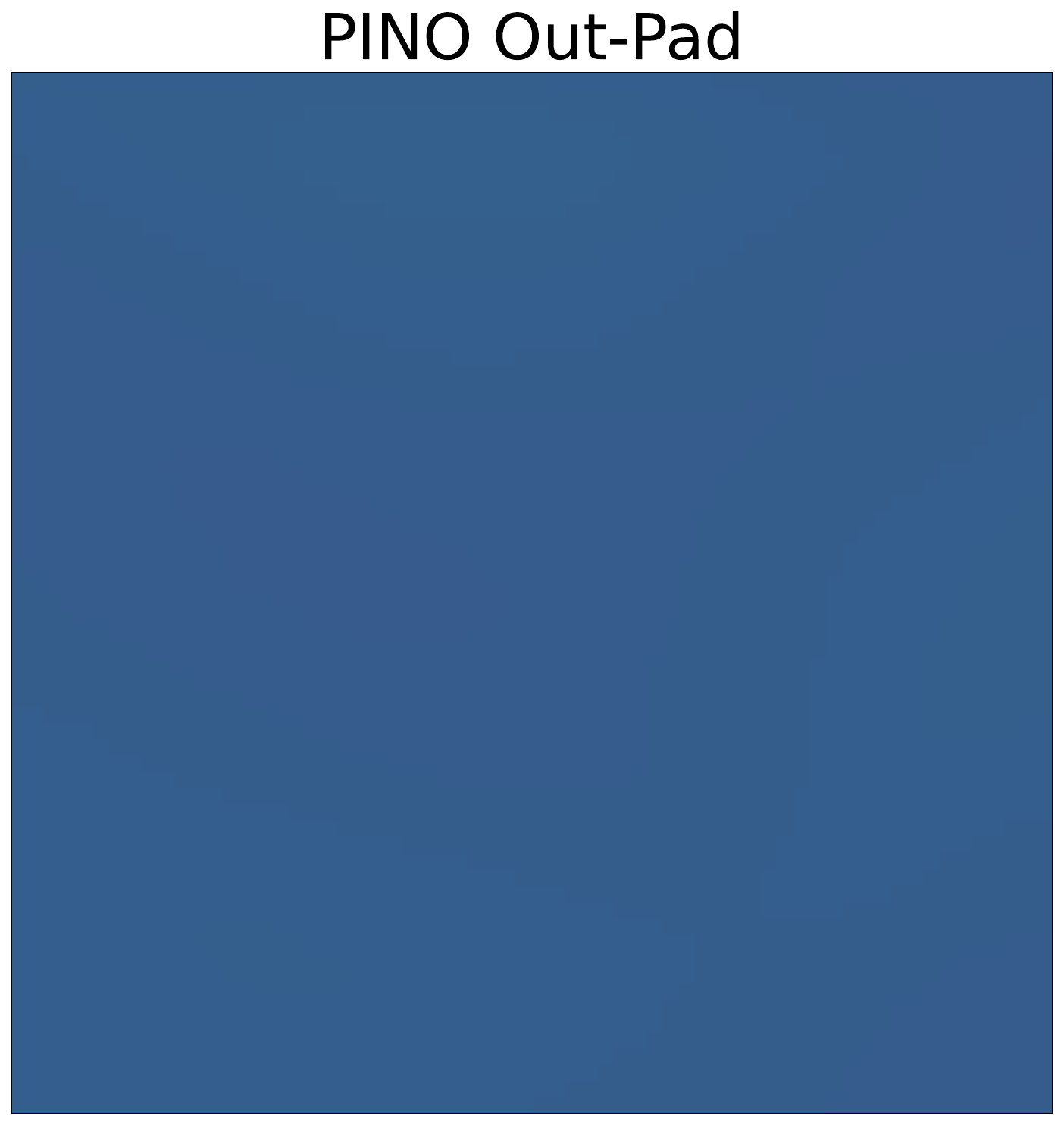}
    \end{minipage}\\

    \hspace{24mm}
    \begin{minipage}[t]{0.325\textwidth}
        \centering
        \includegraphics[width=\linewidth]{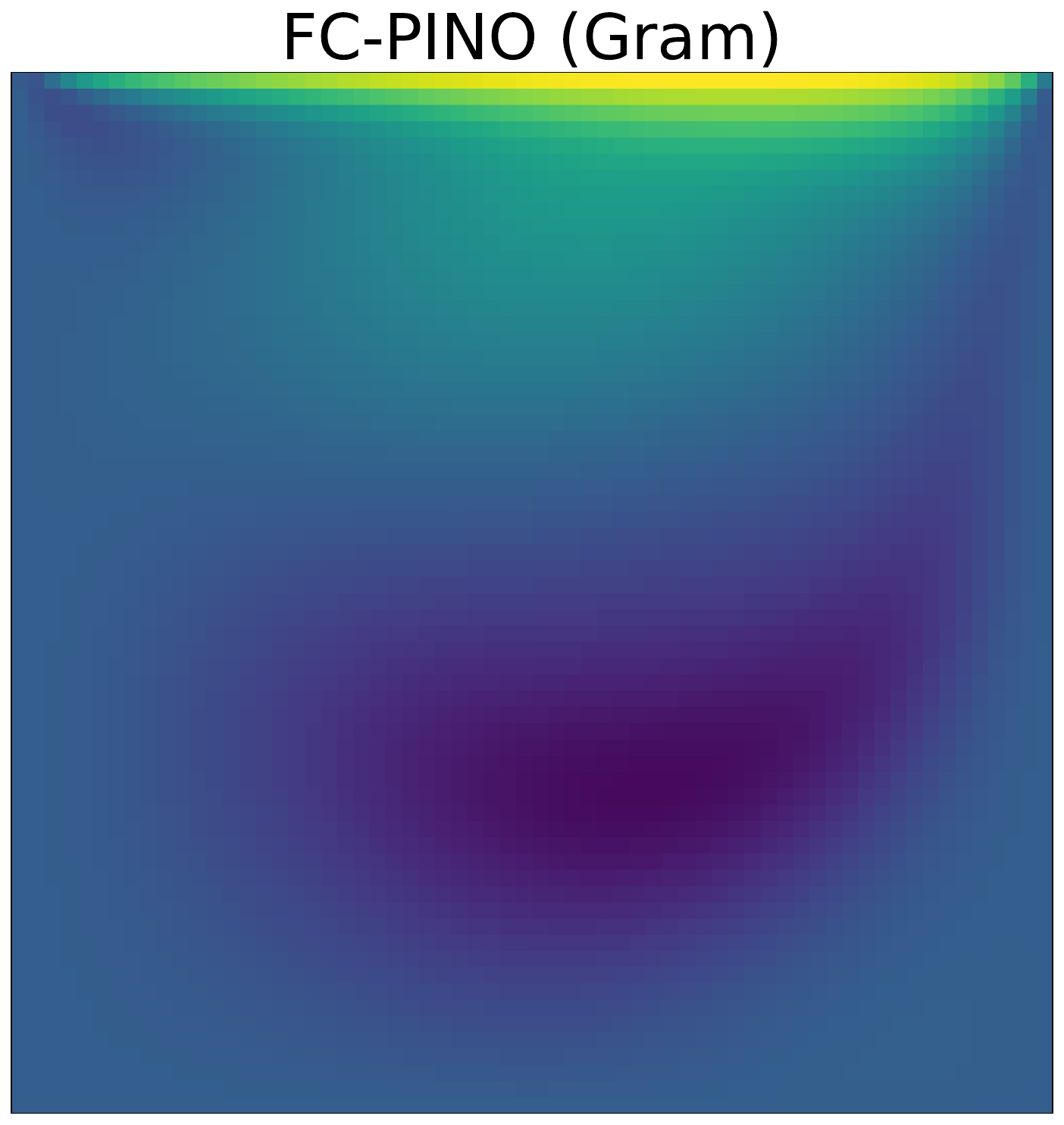}
    \end{minipage}
    \hfill
    \begin{minipage}[t]{0.325\textwidth}
        \centering
        \includegraphics[width=\linewidth]{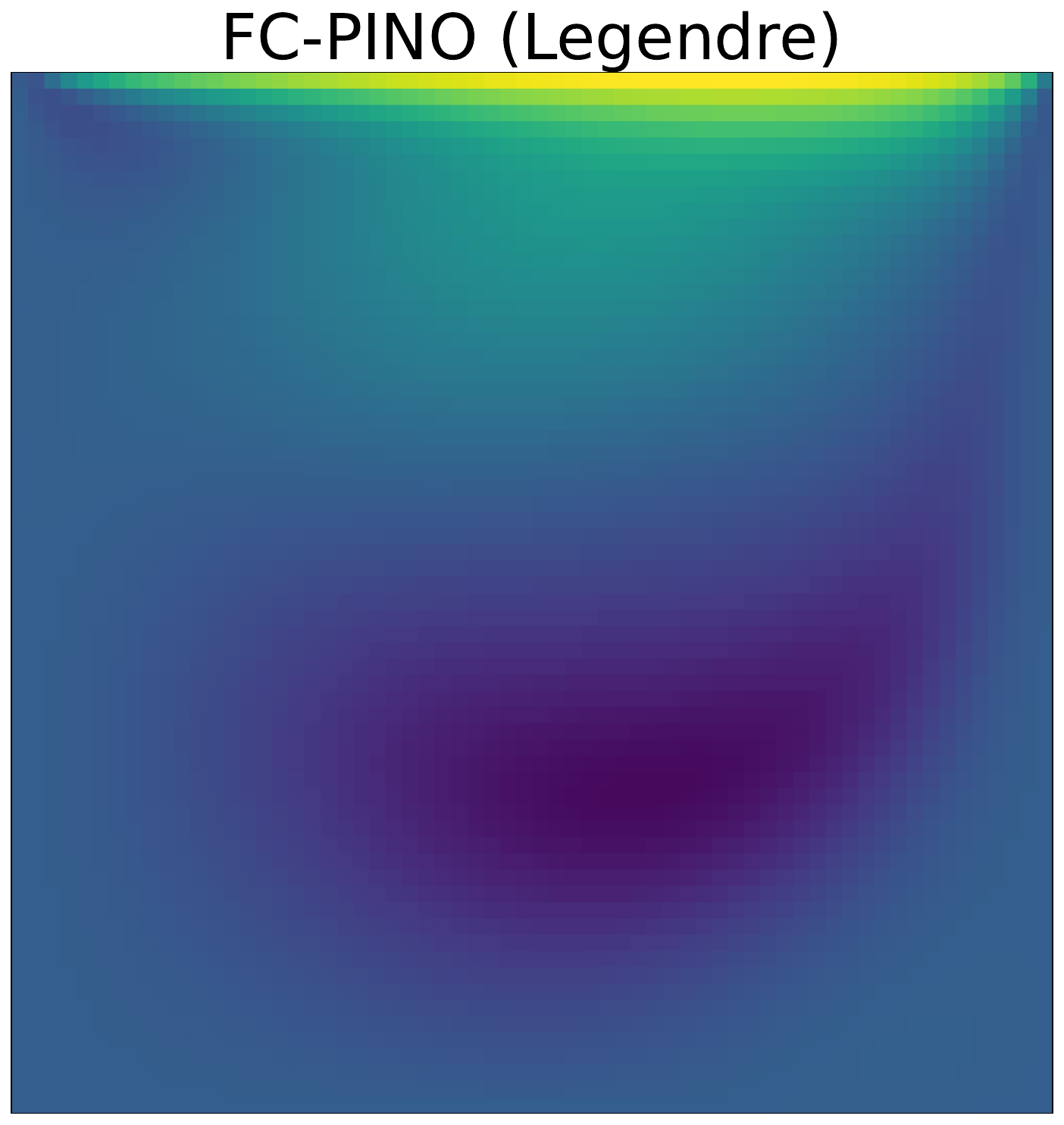}
    \end{minipage}
    \hspace{24mm}

    \vspace{2mm}

    \begin{minipage}[t]{0.325\textwidth}
        \centering
        \includegraphics[width=\linewidth]{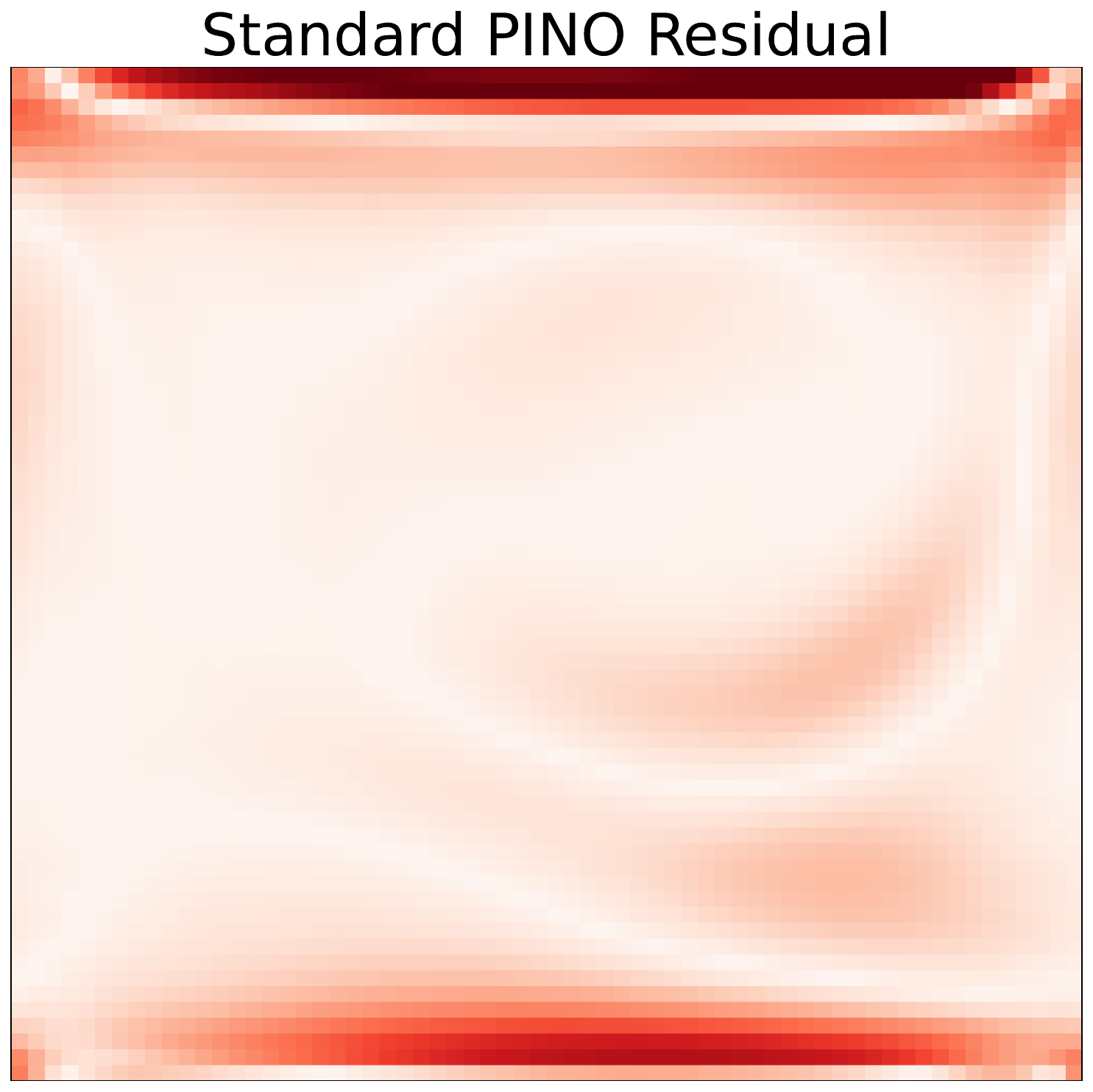}
    \end{minipage}
    \hfill
    \begin{minipage}[t]{0.325\textwidth}
        \centering
        \includegraphics[width=\linewidth]{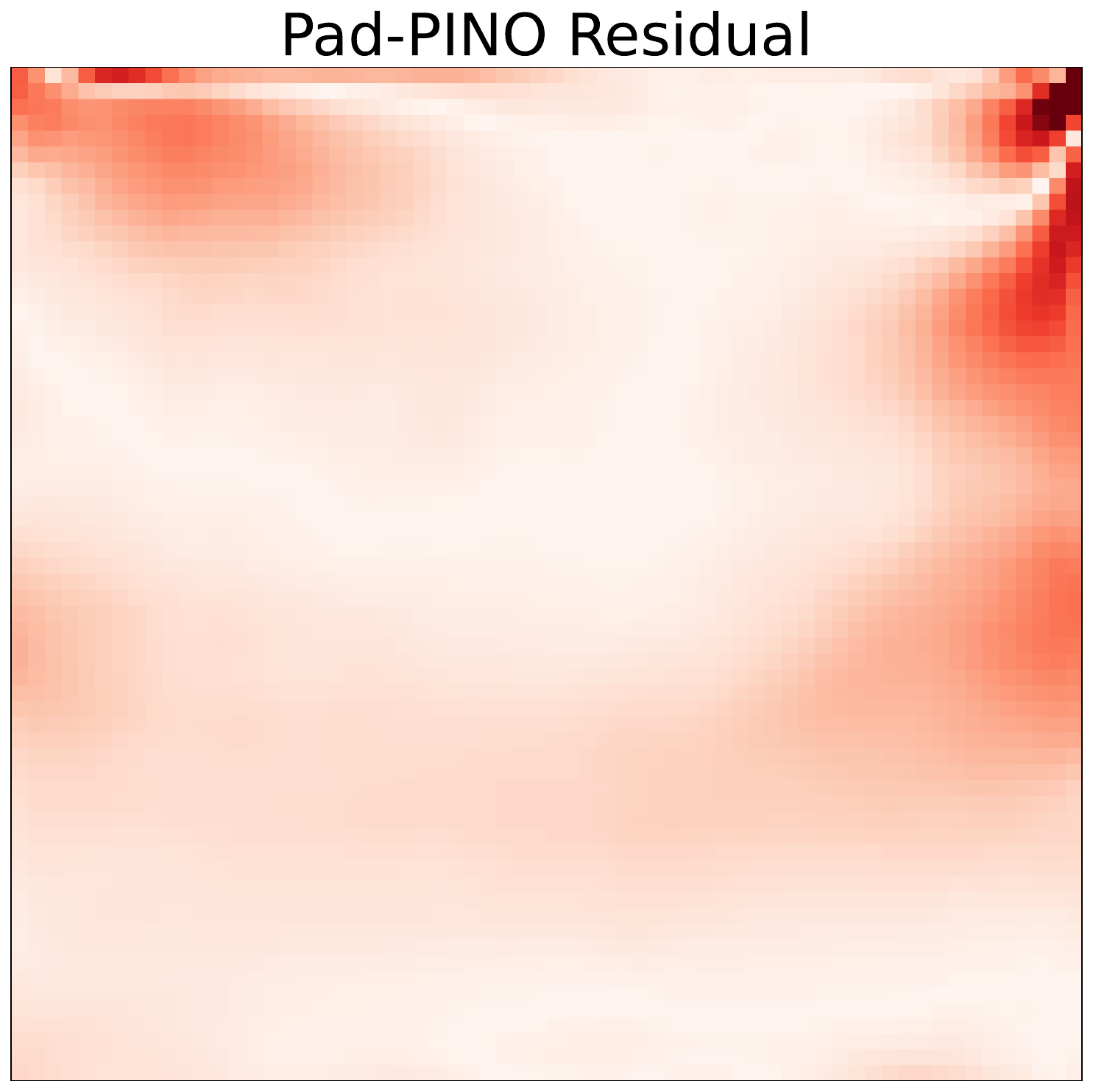}
    \end{minipage}
    \hfill
    \begin{minipage}[t]{0.325\textwidth}
        \centering
        \includegraphics[width=\linewidth]{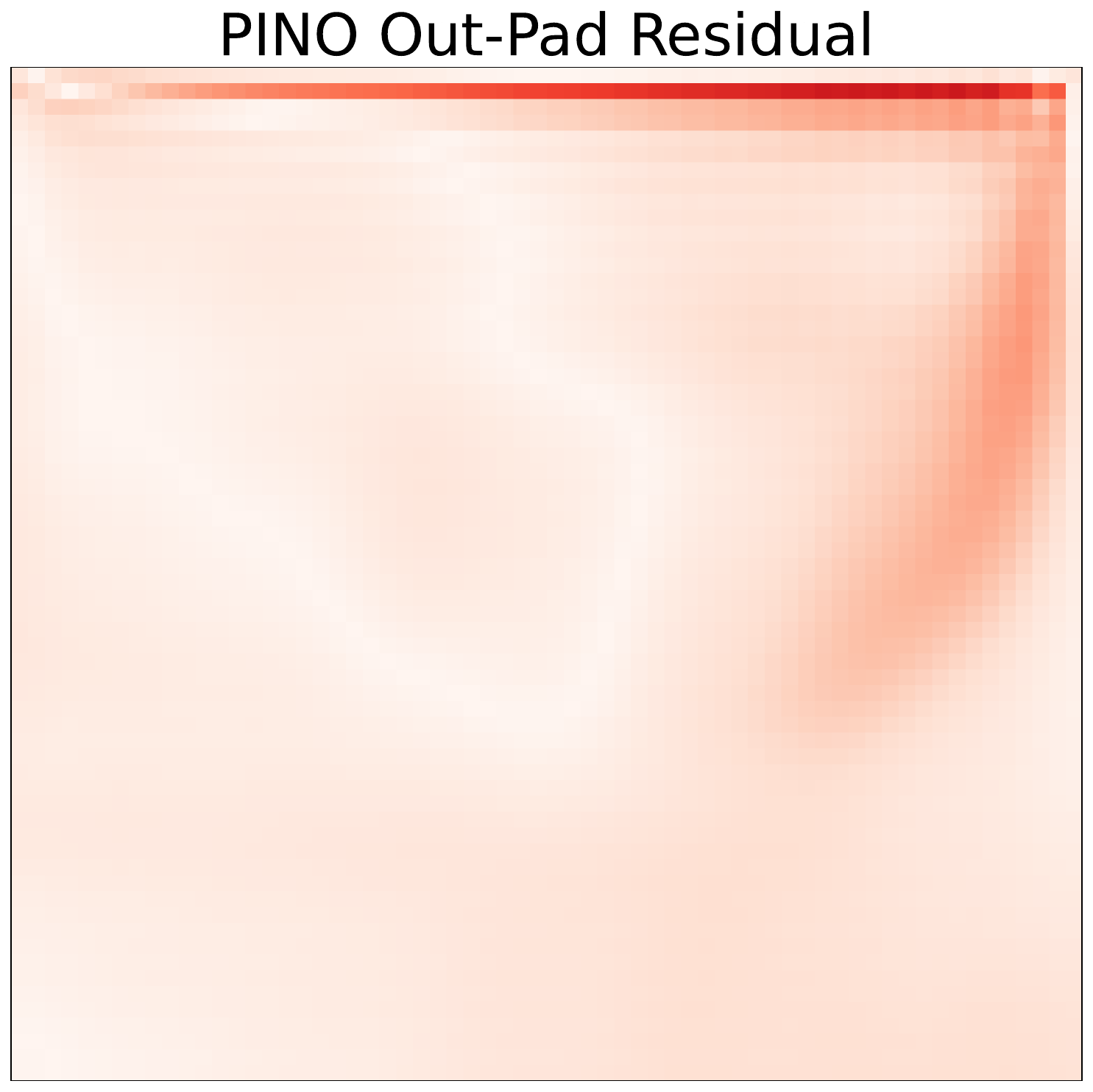}
    \end{minipage}\\

    \hspace{16mm}
    \begin{minipage}[t]{0.325\textwidth}
        \centering
        \includegraphics[width=\linewidth]{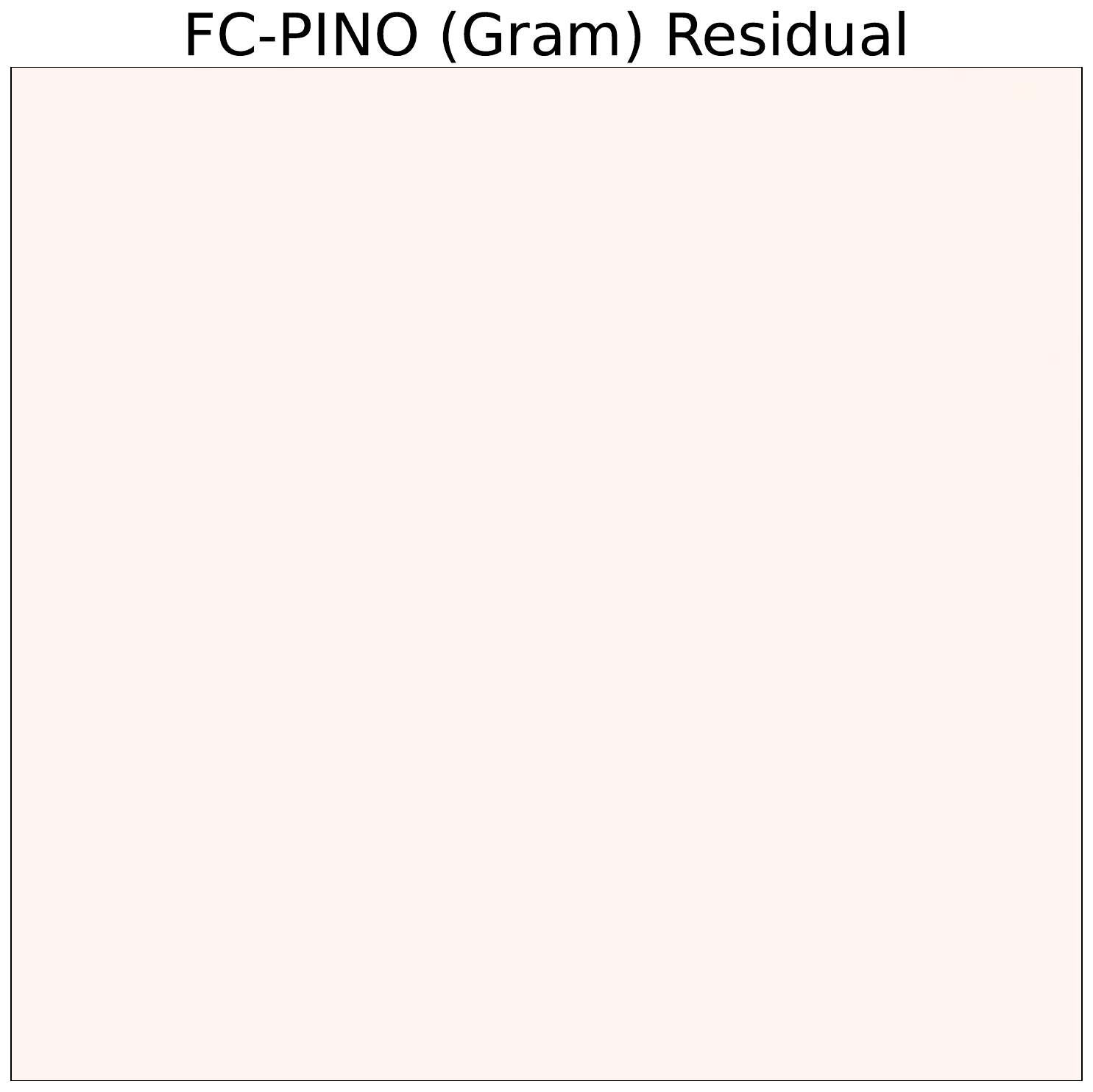}
    \end{minipage}
    \hspace{0.5mm}
    \begin{minipage}[t]{0.325\textwidth}
        \centering
        \includegraphics[width=\linewidth]{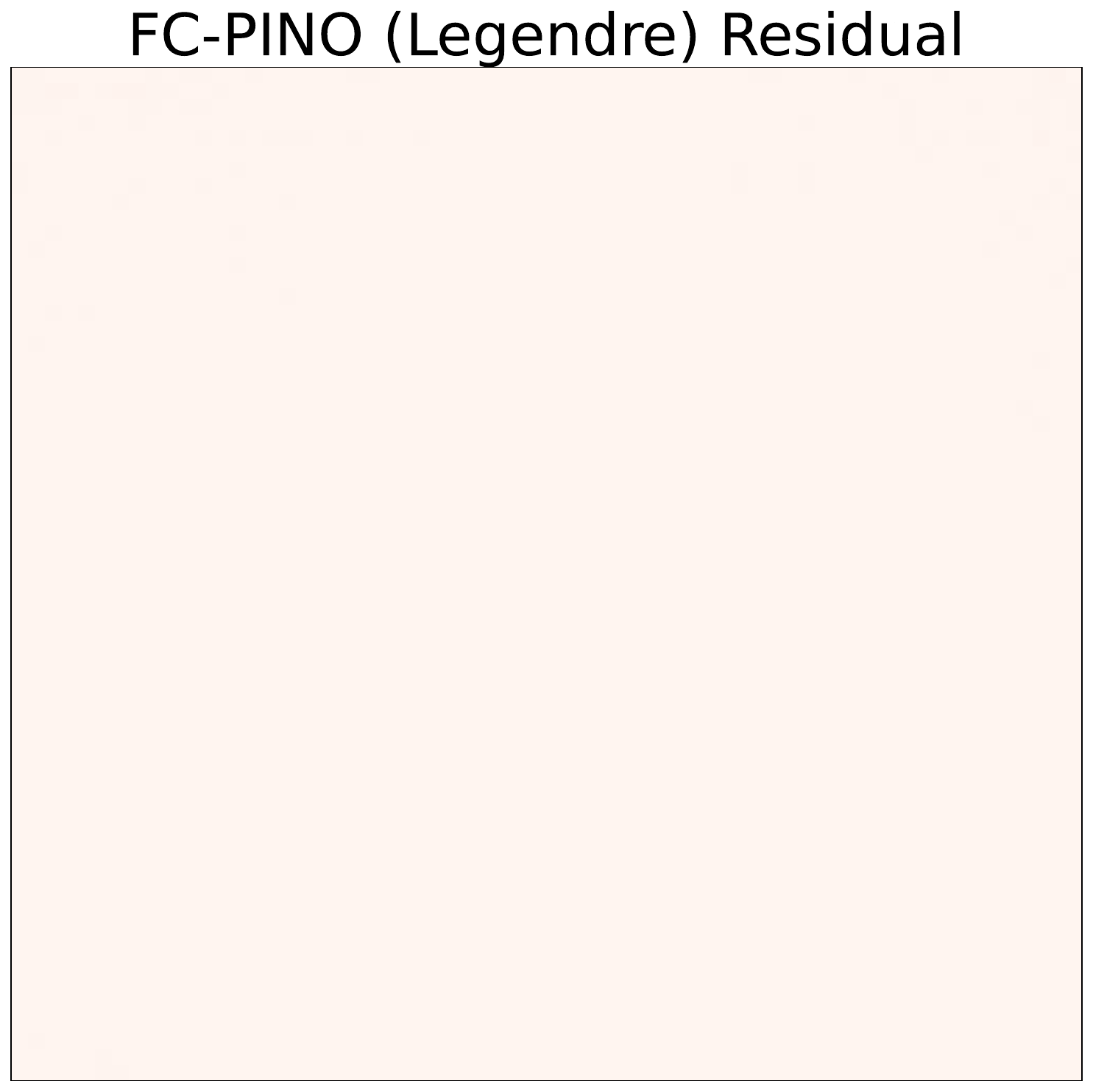}
    \end{minipage}
    \hspace{4mm}
    \begin{minipage}[t]{0.05\textwidth}
        \hfill
        \vspace{-48mm}
        \includegraphics[width=\linewidth]{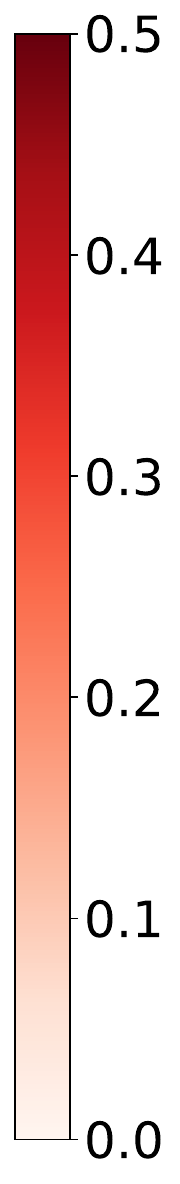}
    \end{minipage}
    \hspace{26mm}

    \vspace{-5mm}
    \caption{Predicted solutions (\emph{top five}) and corresponding PDE residuals (\emph{bottom five}) for baseline PINO architectures and \FCPINO{} variants for the horizontal velocity component of the Navier–Stokes equations.}
    \label{fig:NS_u_combined}
    \vspace{-16mm}
\end{figure}

\begin{figure}[h]
    \centering
    \vspace{-22mm}
    \begin{minipage}[t]{0.325\textwidth}
        \centering
        \includegraphics[width=\linewidth]{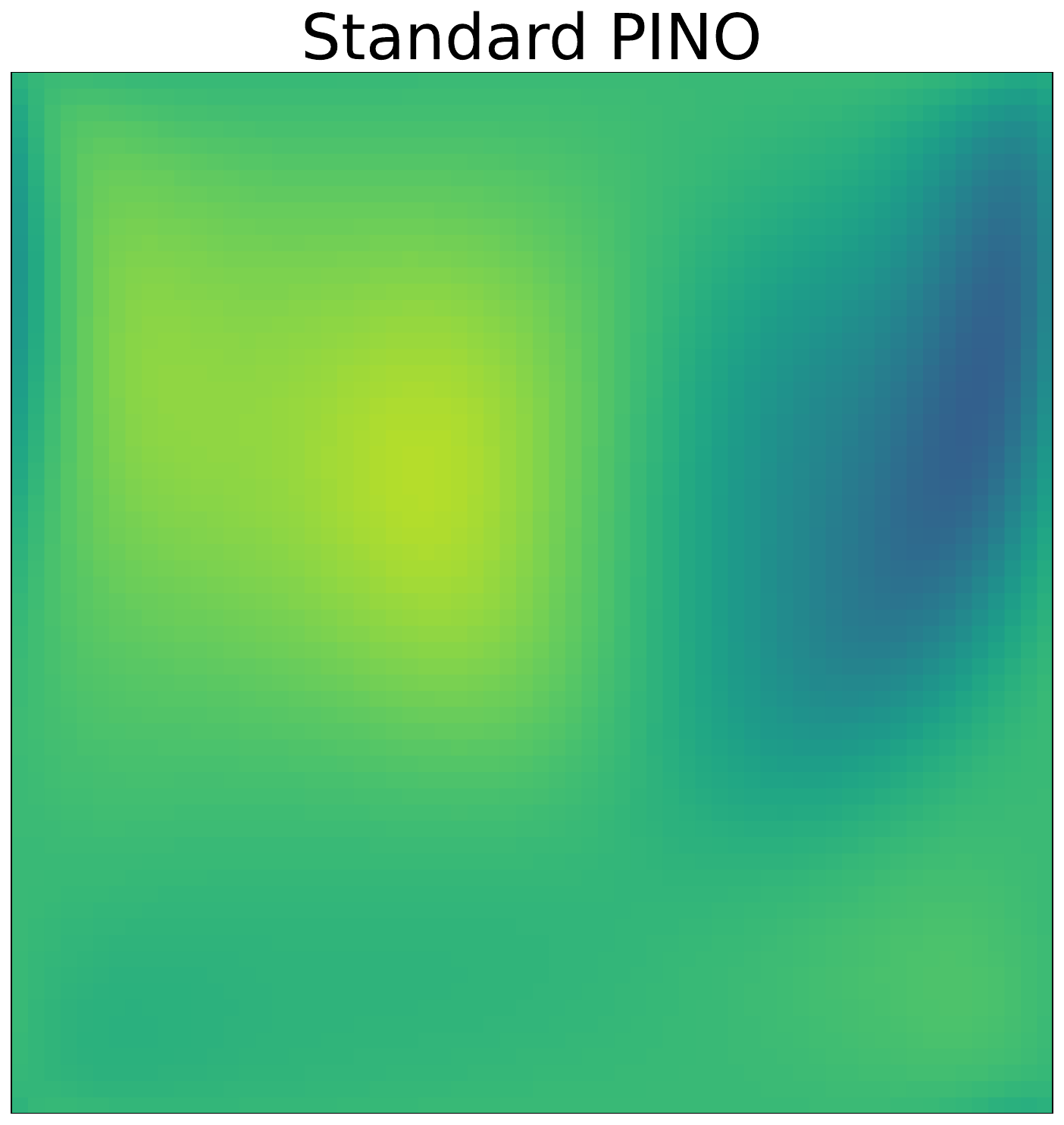}
    \end{minipage}
    \hfill
    \begin{minipage}[t]{0.325\textwidth}
        \centering
        \includegraphics[width=\linewidth]{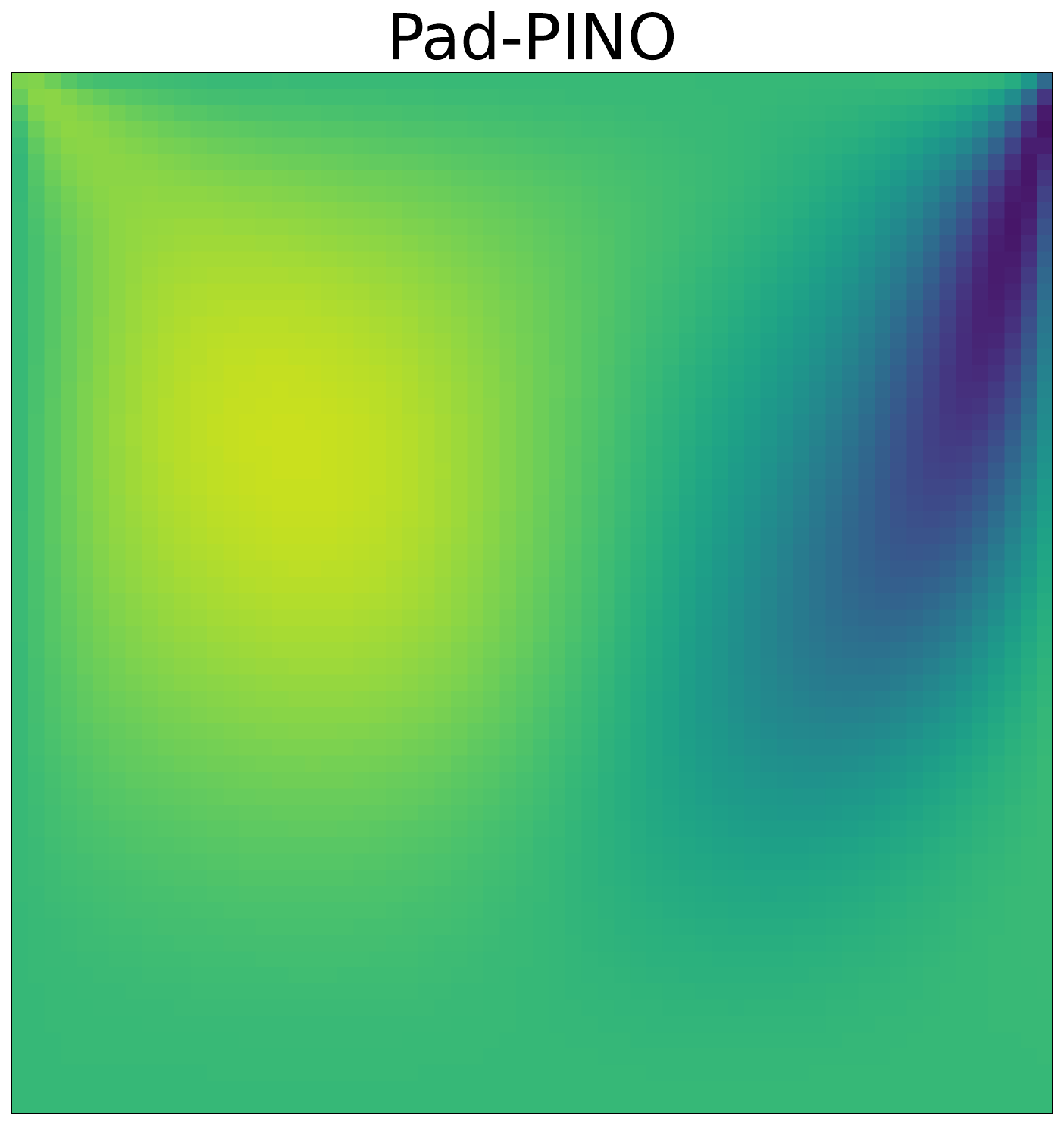}
    \end{minipage}
    \hfill
    \begin{minipage}[t]{0.325\textwidth}
        \centering
        \includegraphics[width=\linewidth]{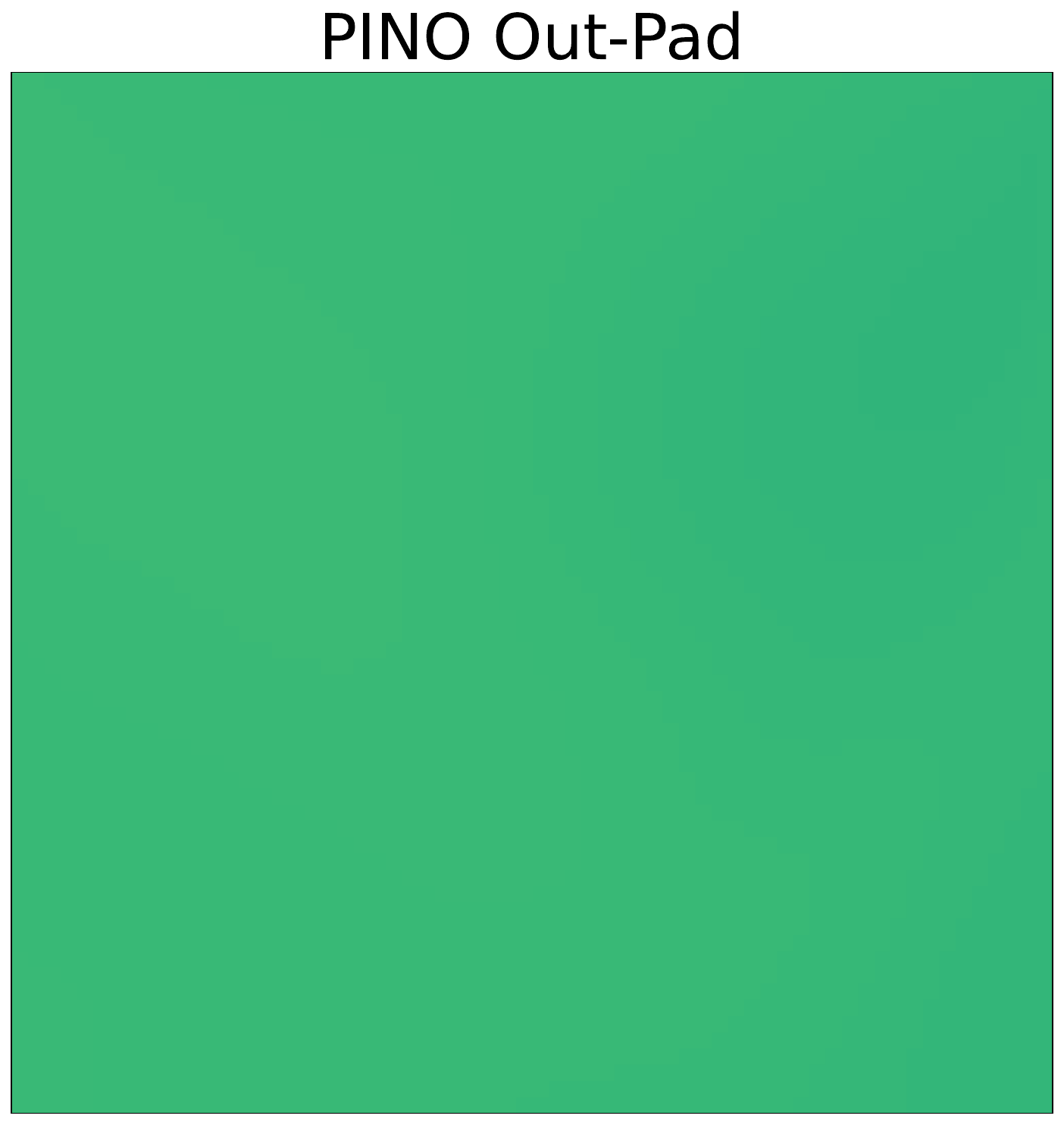}
    \end{minipage}\\

    \hspace{24mm}
    \begin{minipage}[t]{0.325\textwidth}
        \centering
        \includegraphics[width=\linewidth]{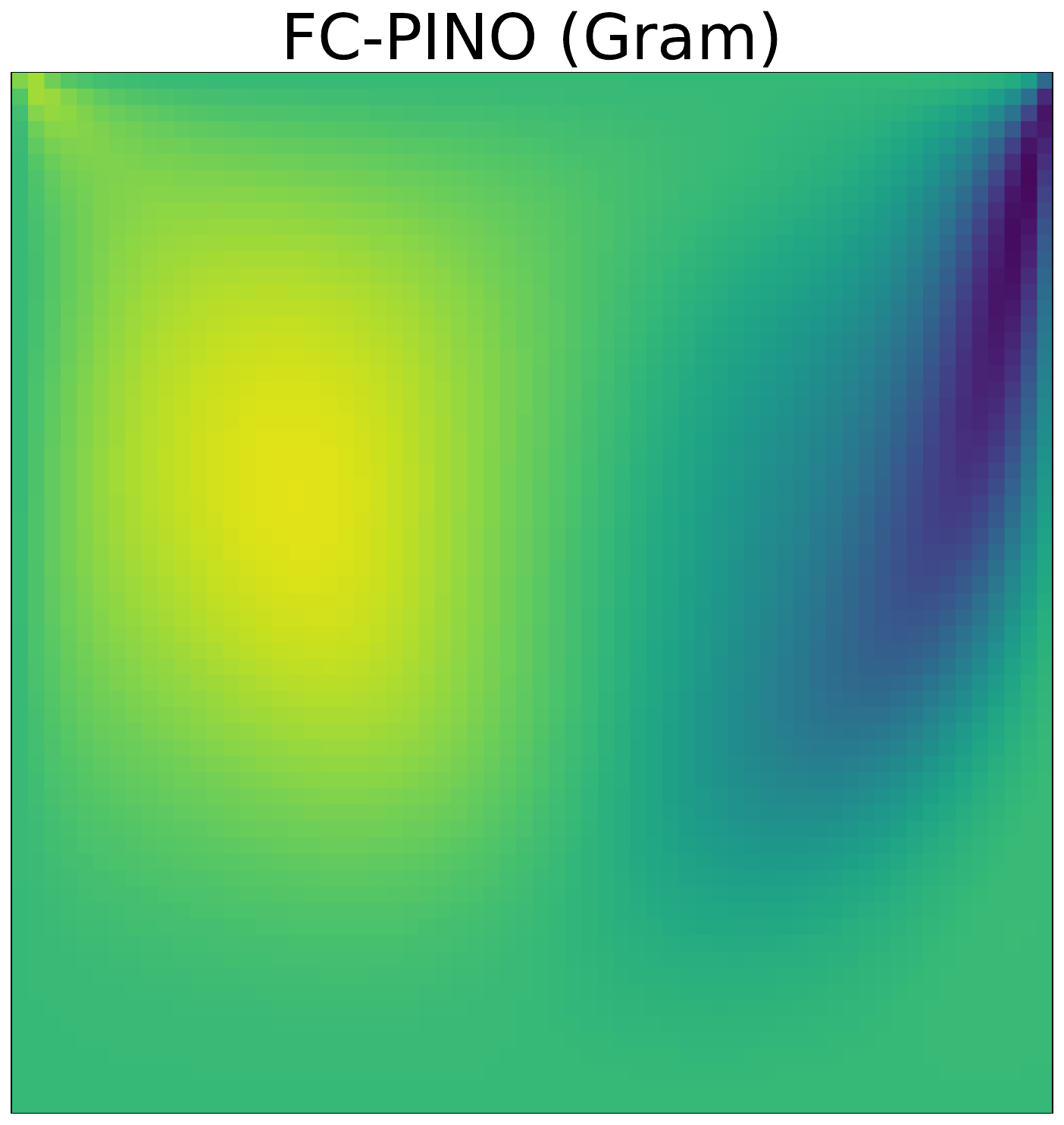}
    \end{minipage}
    \hfill
    \begin{minipage}[t]{0.325\textwidth}
        \centering
        \includegraphics[width=\linewidth]{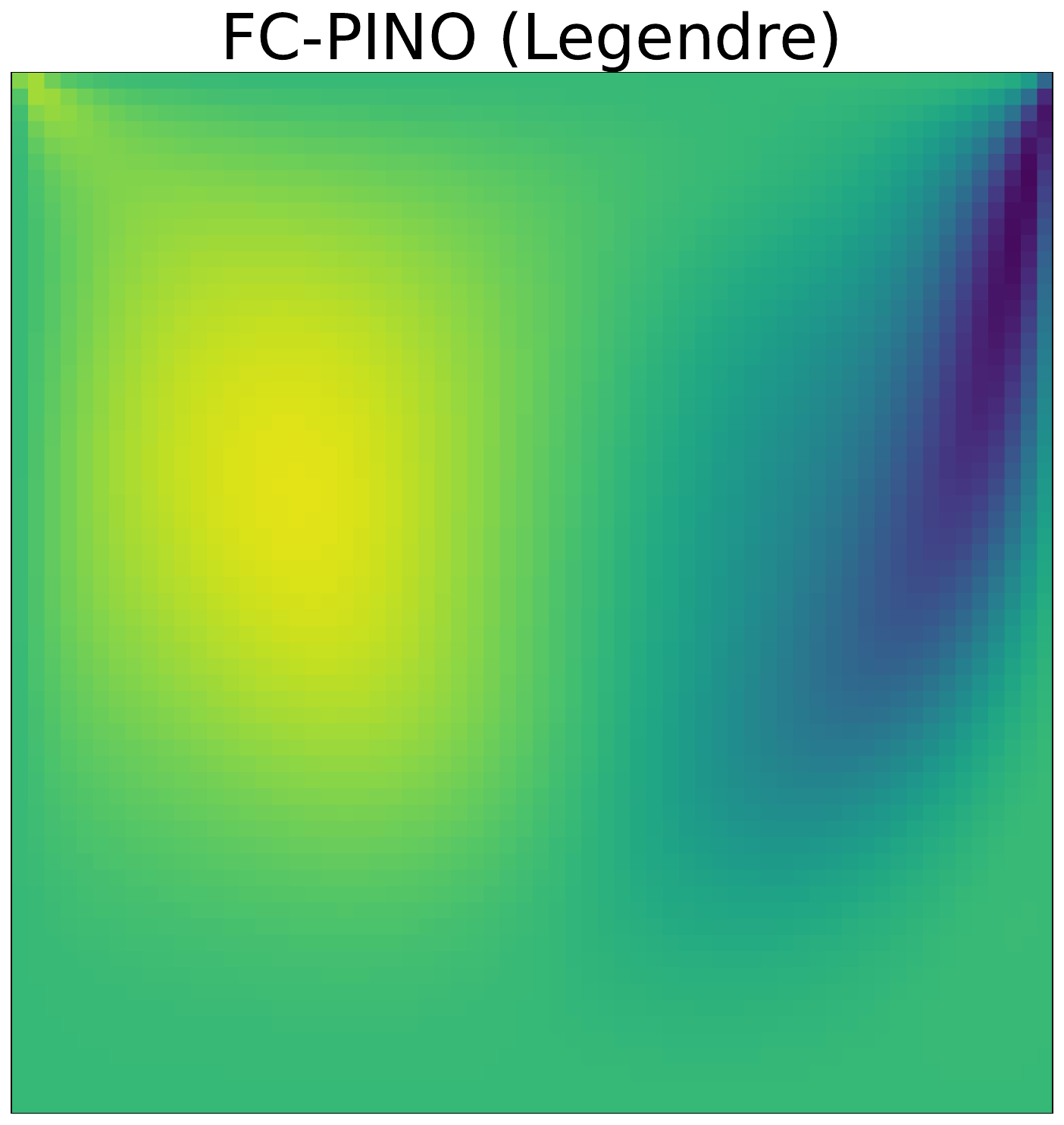}
    \end{minipage}
    \hspace{24mm}

    \vspace{2mm}

    \begin{minipage}[t]{0.325\textwidth}
        \centering
        \includegraphics[width=\linewidth]{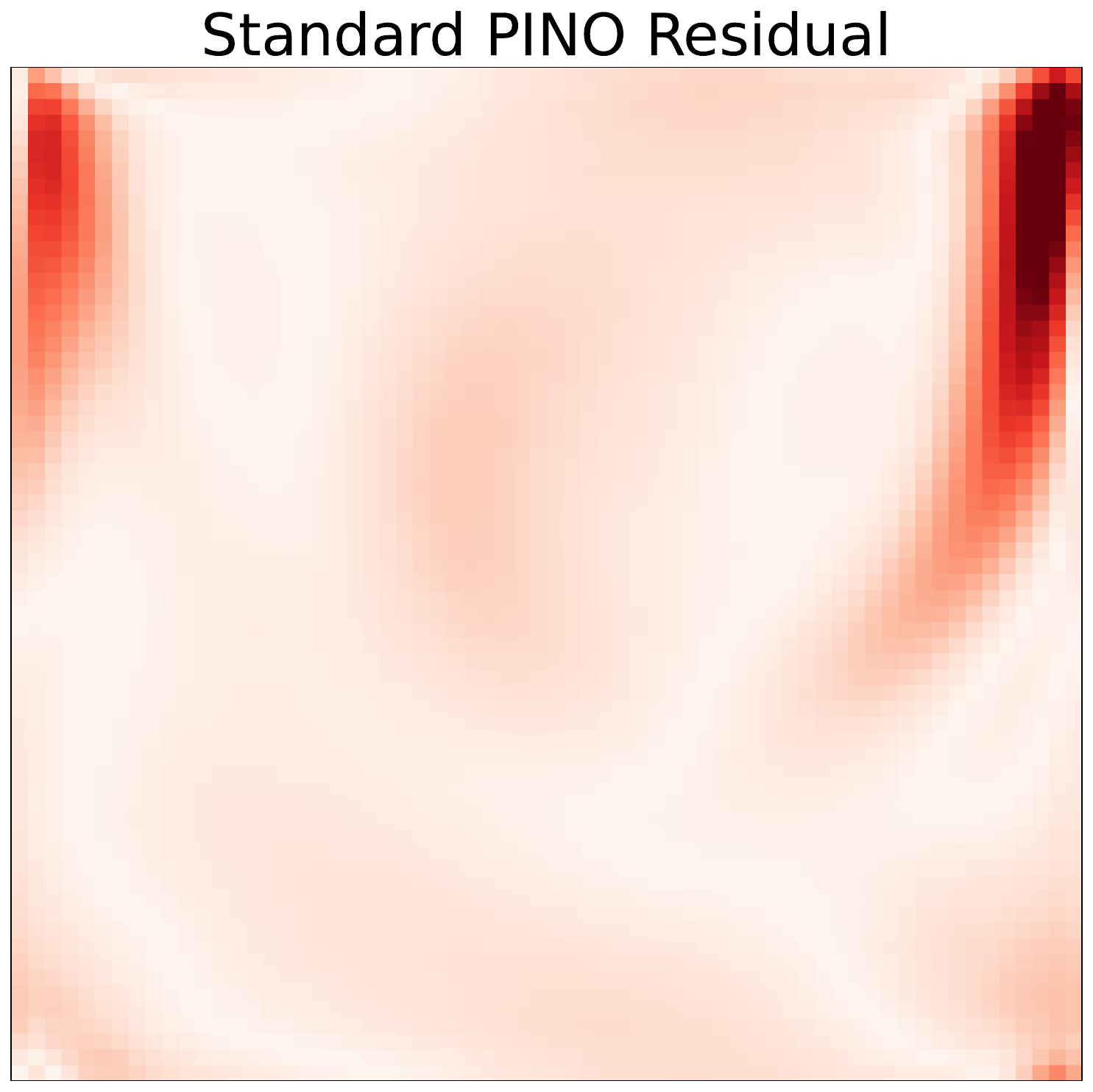}
    \end{minipage}
    \hfill
    \begin{minipage}[t]{0.325\textwidth}
        \centering
        \includegraphics[width=\linewidth]{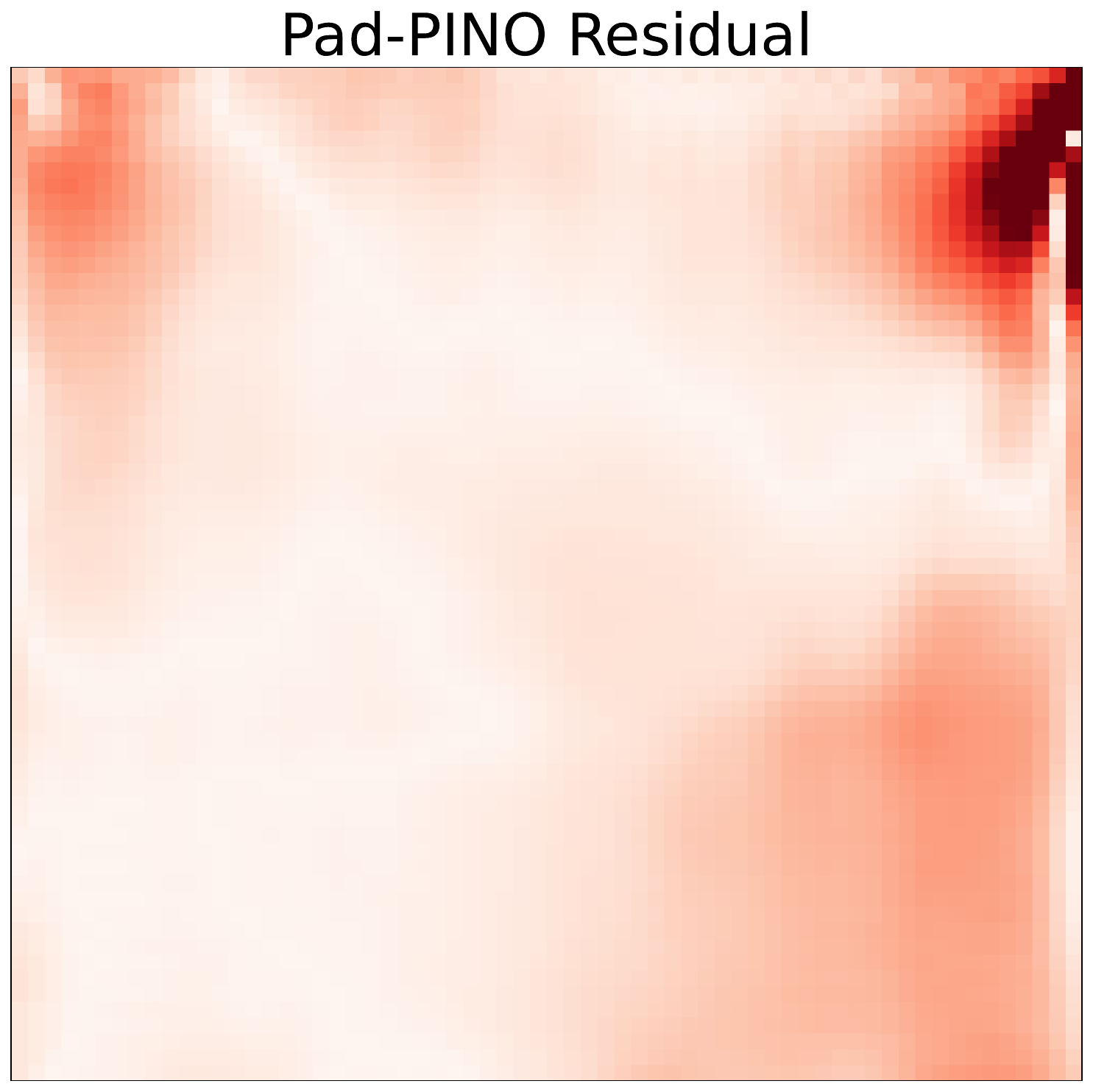}
    \end{minipage}
    \hfill
    \begin{minipage}[t]{0.325\textwidth}
        \centering
        \includegraphics[width=\linewidth]{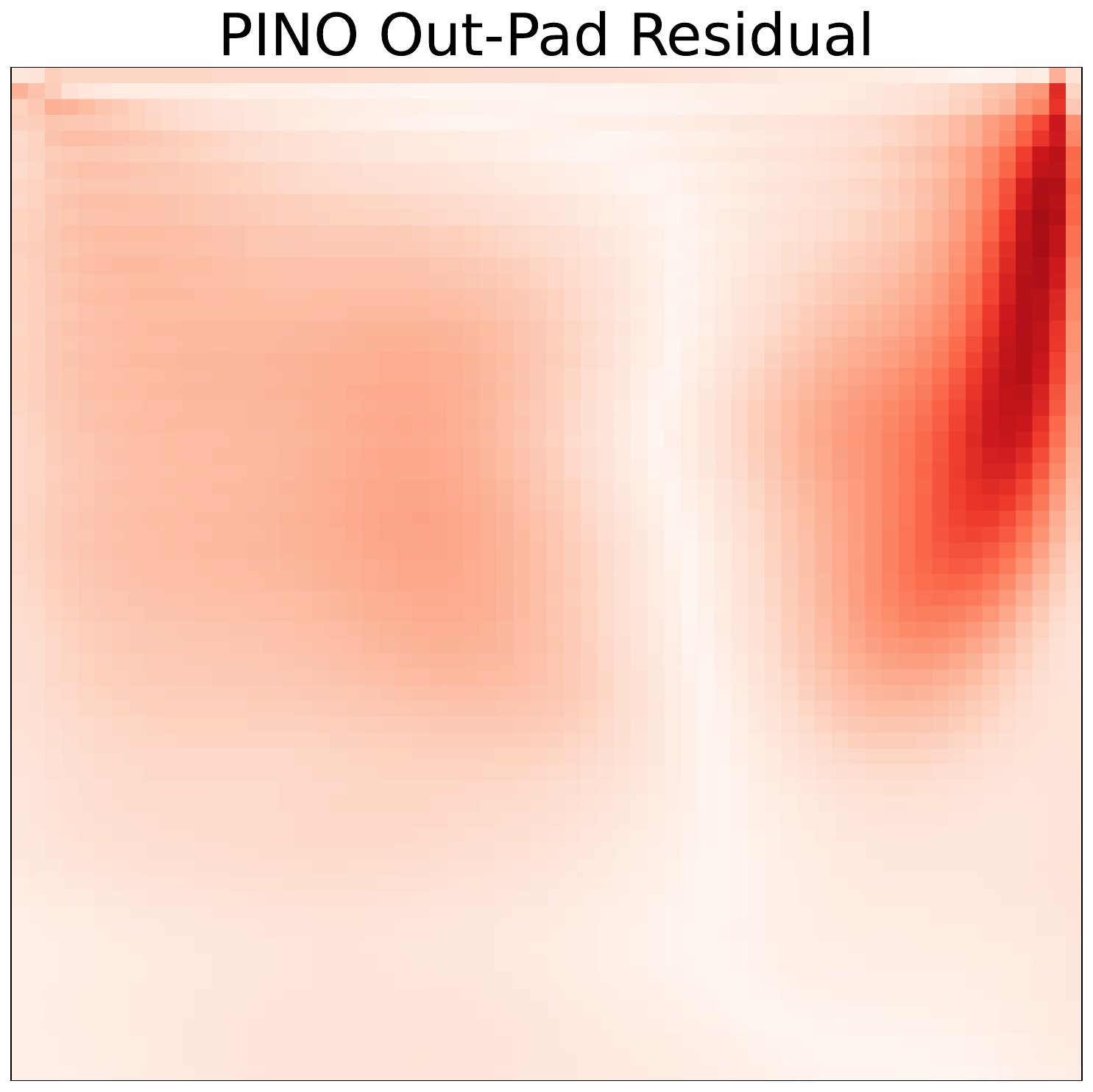}
    \end{minipage}\\

    \hspace{16mm}
    \begin{minipage}[t]{0.325\textwidth}
        \centering
        \includegraphics[width=\linewidth]{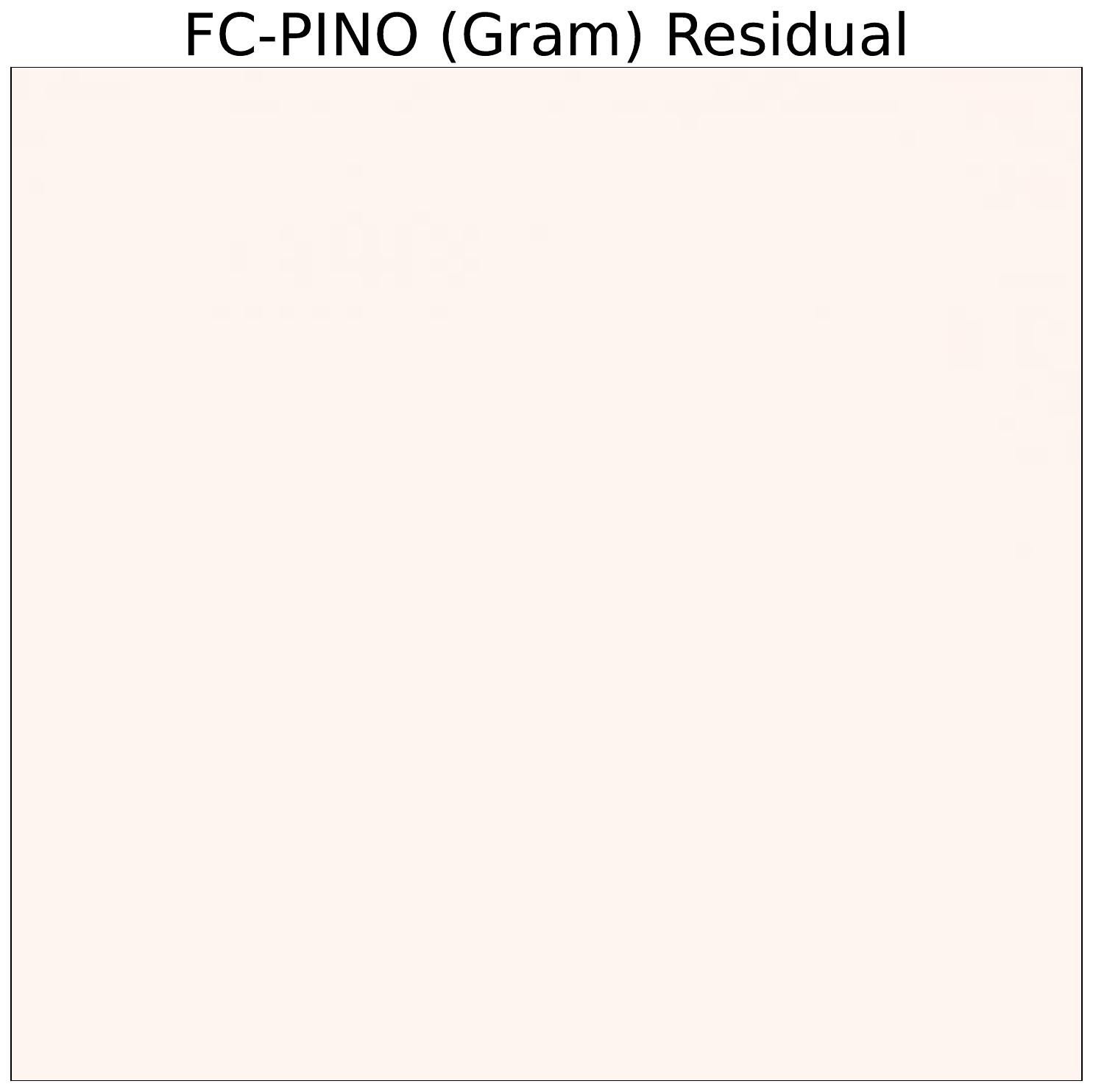}
    \end{minipage}
    \hspace{0.5mm}
    \begin{minipage}[t]{0.325\textwidth}
        \centering
        \includegraphics[width=\linewidth]{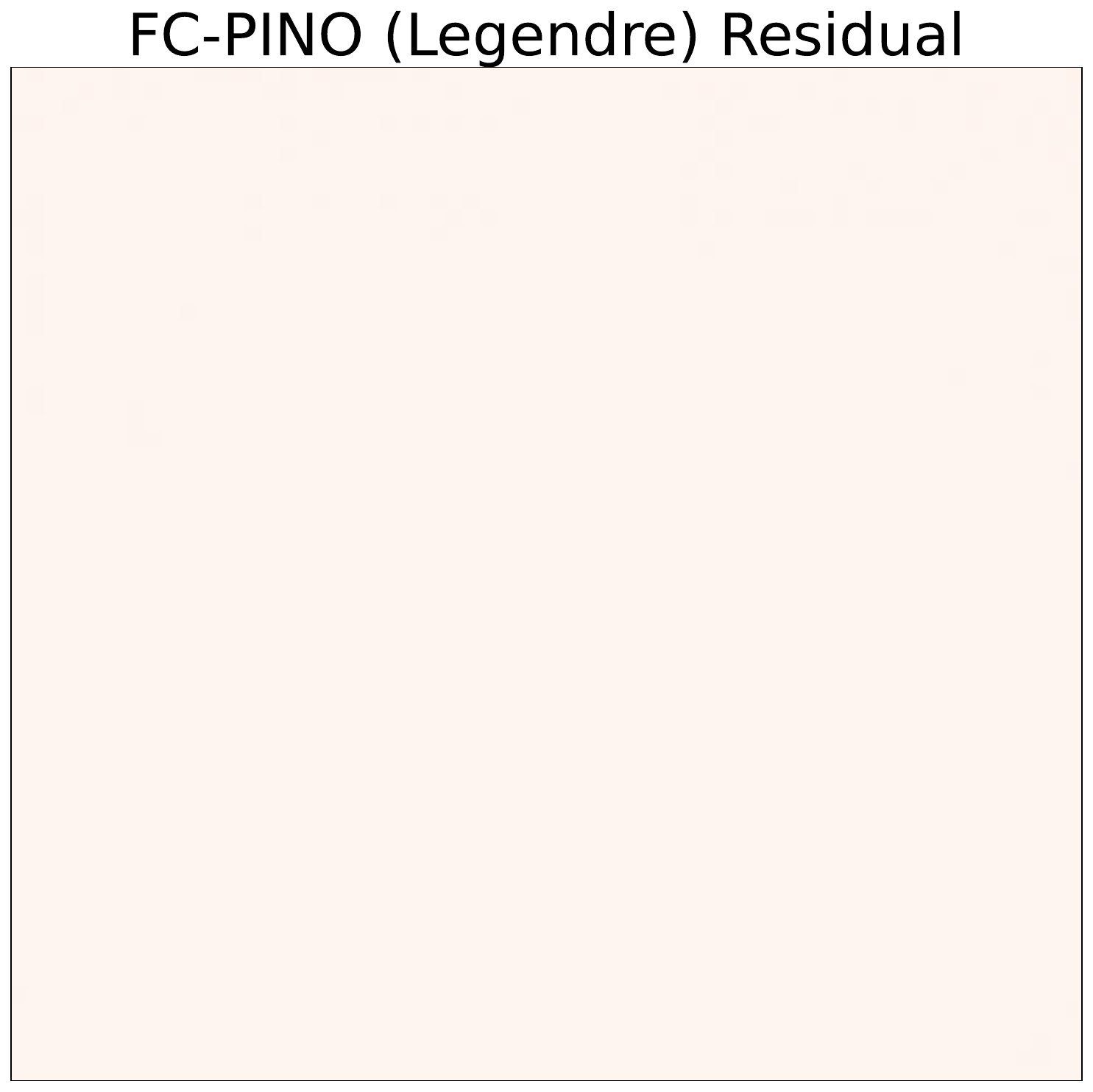}
    \end{minipage}
    \hspace{4mm}
    \begin{minipage}[t]{0.05\textwidth}
        \hfill
        \vspace{-48mm}
        \includegraphics[width=\linewidth]{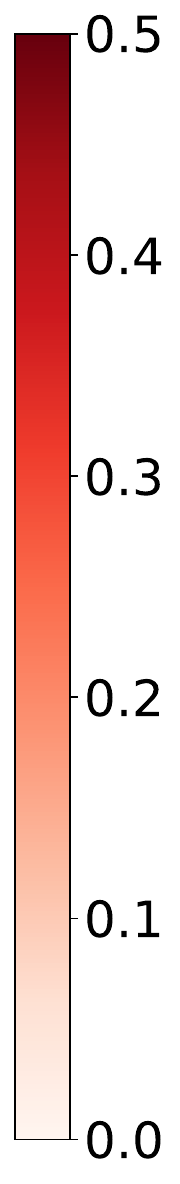}
    \end{minipage}
    \hspace{26mm}

    \vspace{-5mm}
    \caption{Predicted solutions (\emph{top five}) and corresponding PDE residuals (\emph{bottom five}) for baseline PINO architectures and \FCPINO{} variants for the horizontal velocity component of the Navier–Stokes equations.}
    \label{fig:NS_v_combined}
    \vspace{-16mm}
\end{figure}

\clearpage

\section{Convergence Plots (PDE Residual versus Iterations)}
\label{appx:convergence_iters}

\vspace{2mm}

In this appendix, we display convergence plots showing the evolution of the PDE residual during training, as a function of the number of training iterations. 

\vspace{1mm}

\subsection{1D Self-Similar Burger's Equation}
\label{appx:conv_iters_1d_burgers}

\begin{figure}[h]
    \centering
    \begin{minipage}[t]{0.76\textwidth}
        \centering
        \includegraphics[width=\linewidth]{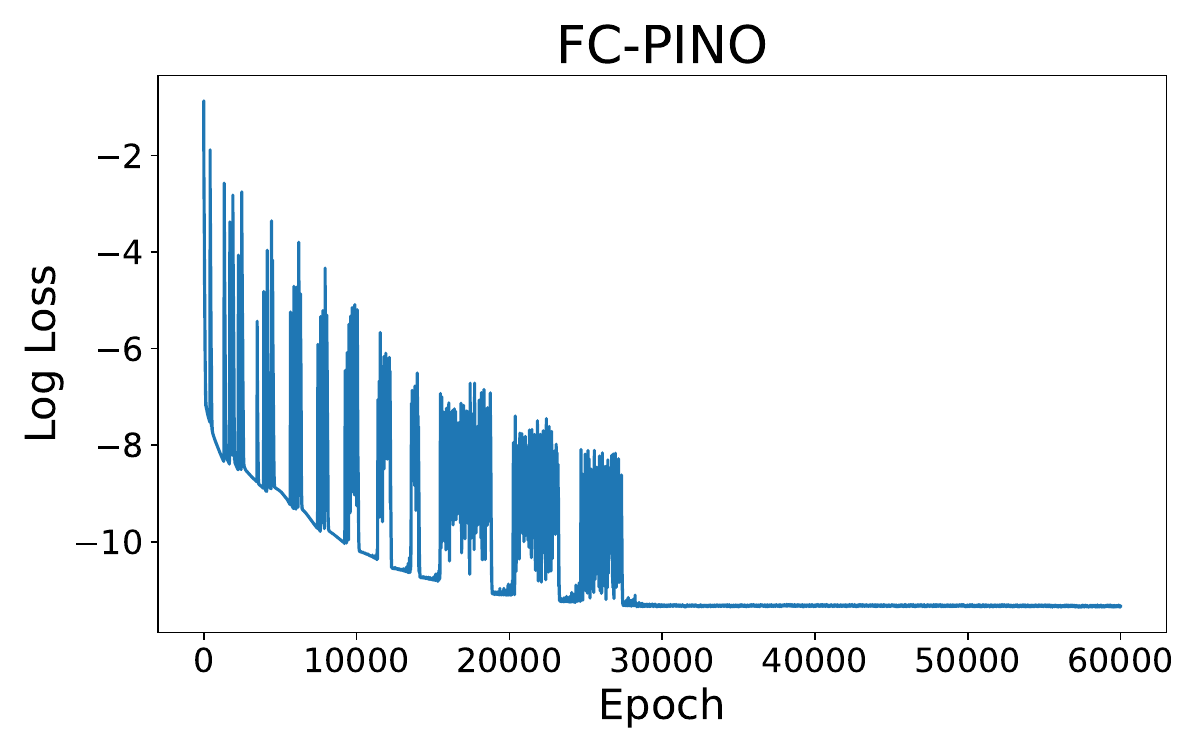}\\
    \end{minipage}
\vspace{-2mm}
    \caption{Evolution of the PDE residual loss for the 1D Burgers' equation with \FCPINO.}
\end{figure}

\vspace{2mm} 

\subsection{2D Burger's Equation}
\label{appx:conv_iters_2d_burgers}

\begin{figure}[h]
    \centering
    \begin{minipage}[t]{0.76\textwidth}
        \centering
        \includegraphics[width=\linewidth]{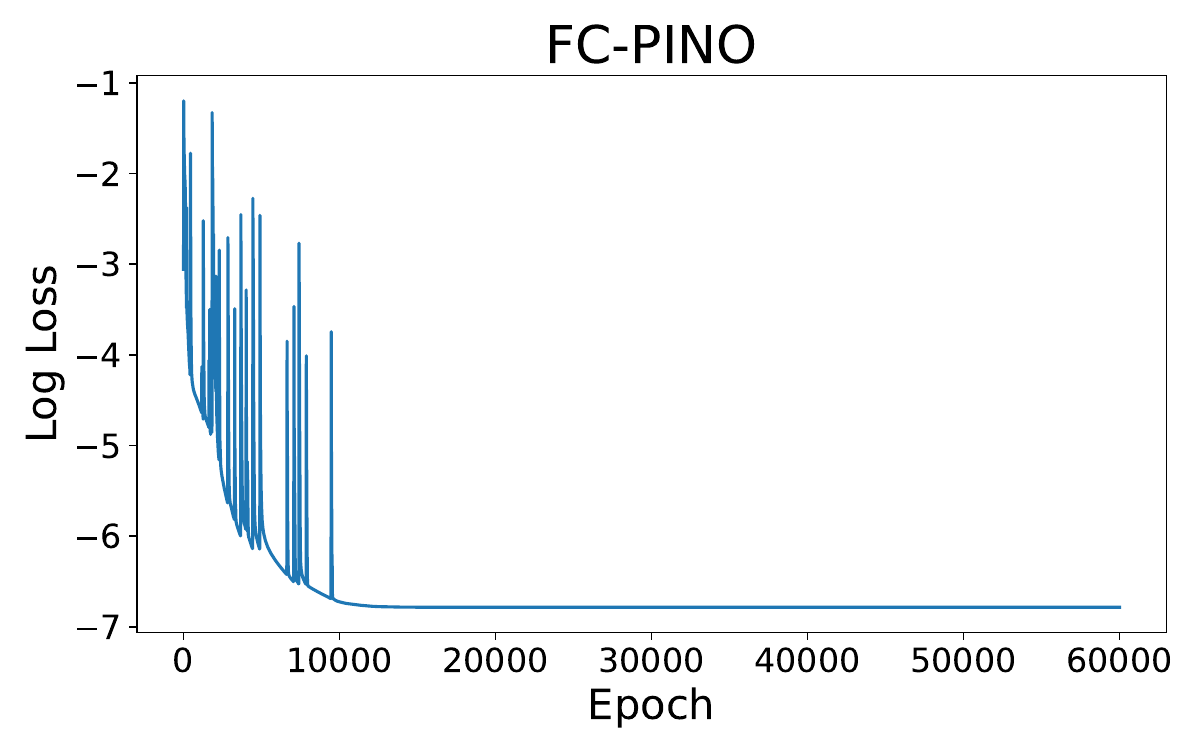}\\
    \end{minipage}
    \vspace{-2mm}
    \caption{Evolution of the PDE residual loss for the 2D Burgers' equation with \FCPINO.}
\end{figure}

\clearpage

\hfill 
\vspace{-18mm}

\subsection{Navier--Stokes Equations}
\label{appx:conv_iters_NS}

\vspace{-3mm}

\begin{figure}[h]
    \centering
    \begin{minipage}[t]{0.63\textwidth}
        \centering
        \includegraphics[width=\linewidth]{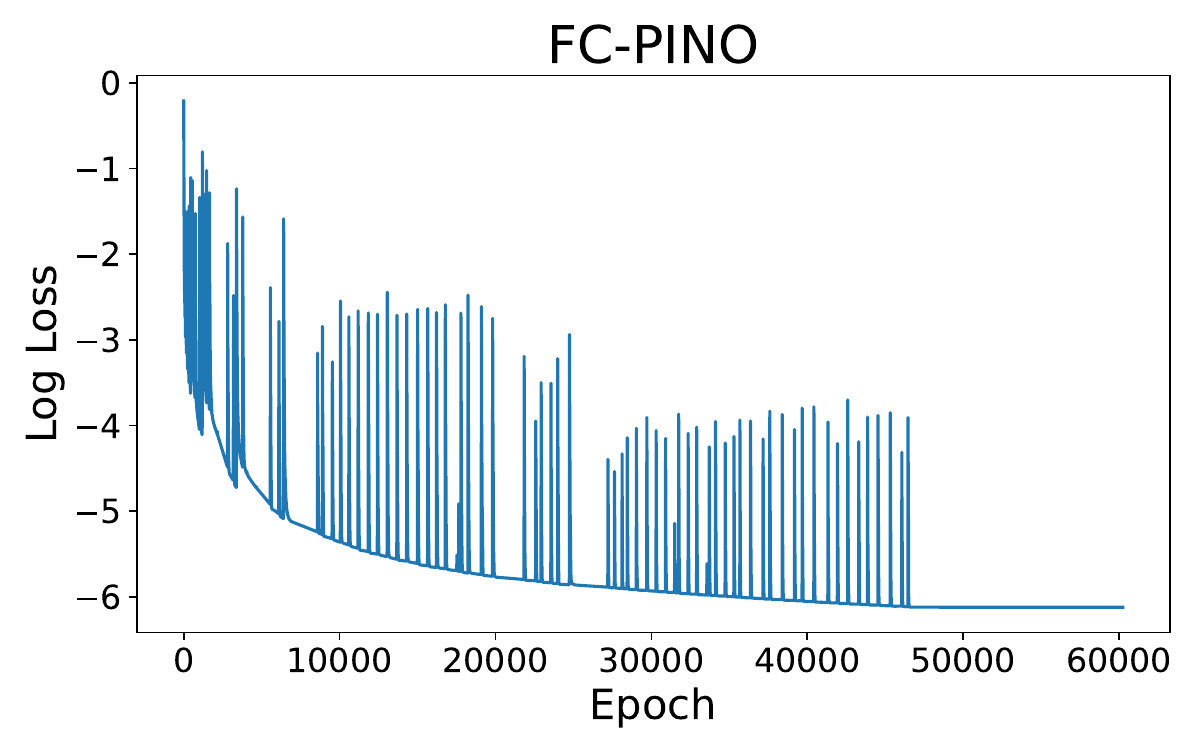}\\
    \end{minipage} \vspace{-3mm}
    \caption{Evolution of the PDE residual for the Navier--Stokes equations with \FCPINO.}
\end{figure}

\section{Convergence Plots (Total Loss versus Time)}
\label{appx:convergence_time}

In this appendix, we display convergence plots showing the evolution of the total loss as a function of training time, for \textsc{FC--Gram}. For a fair comparison, the codes for the 1D and 2D Burgers' equations were run on the same workstation equipped with a single NVIDIA RTX 4090 GPU (24GB VRAM), an AMD Ryzen 9 7900X CPU, and 64GB of system RAM. For the Navier--Stokes equation, all models are run on a NVIDIA H100 80GB HBM3.

\subsection{1D Self-Similar Burger's Equation}
\label{appx:conv_time_1d_burgers}

\vspace{-3mm}

\begin{figure}[h]
    \centering
    \includegraphics[width=0.82\linewidth]{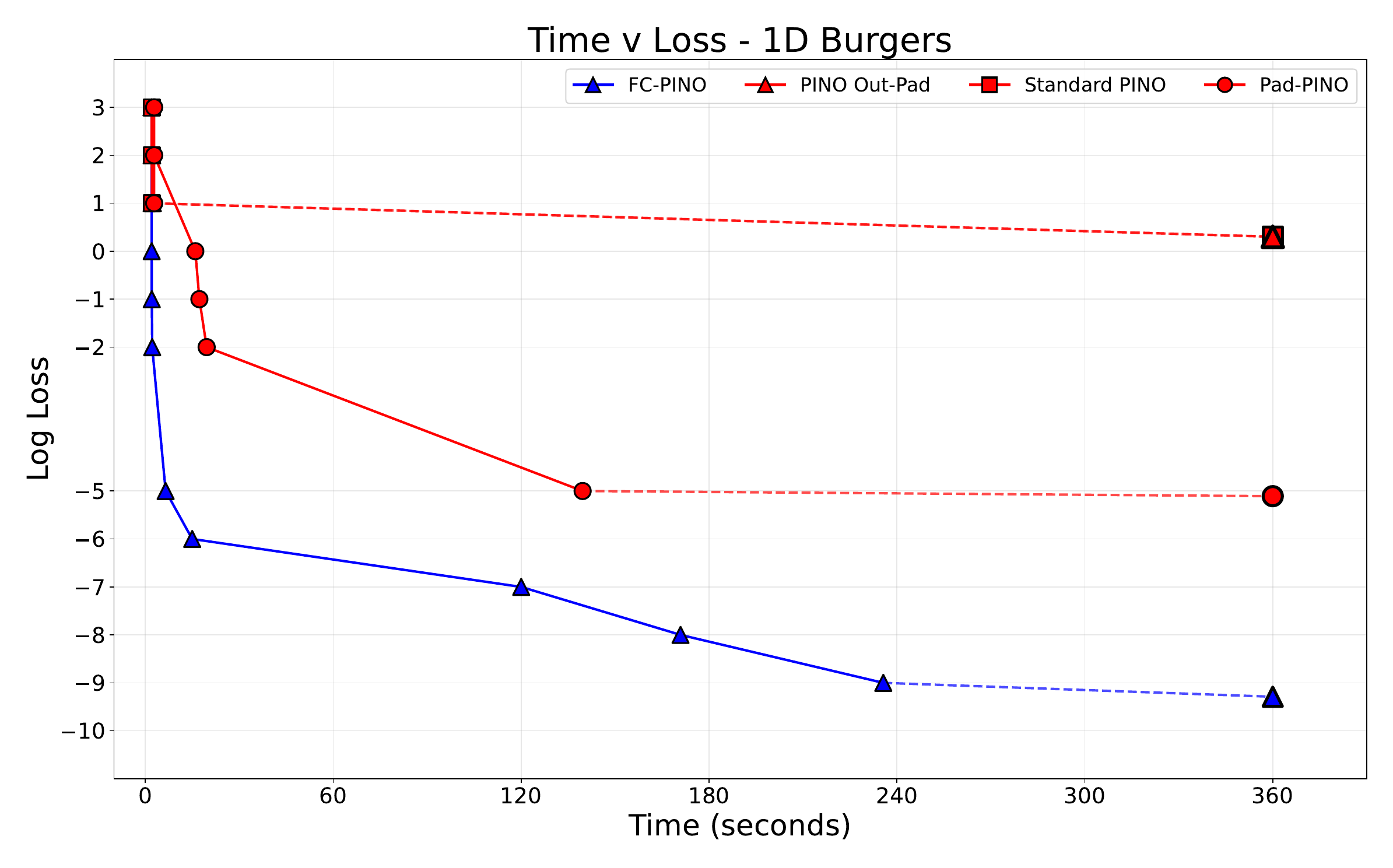} \vspace{-3.5mm}
    \caption{Evolution of the total loss as a function of time for the 1D Burgers' equation with the different PINO approaches. To allow for a cleaner visualization, we display the time taken to reach certain threshold loss values for the first time as a solid line. The dotted line then connects the loss at the last threshold value and the loss at $360$ seconds. }
    \label{fig:threshold_crossing} \vspace{-16mm}
\end{figure}

\clearpage

\hfill 
\vspace{-16mm}

\subsection{2D Burger's Equation}
\label{appx:conv_time_2d_burgers}

\vspace{-5mm}

\begin{figure}[h]
    \centering
    \includegraphics[width= 0.86 \linewidth]{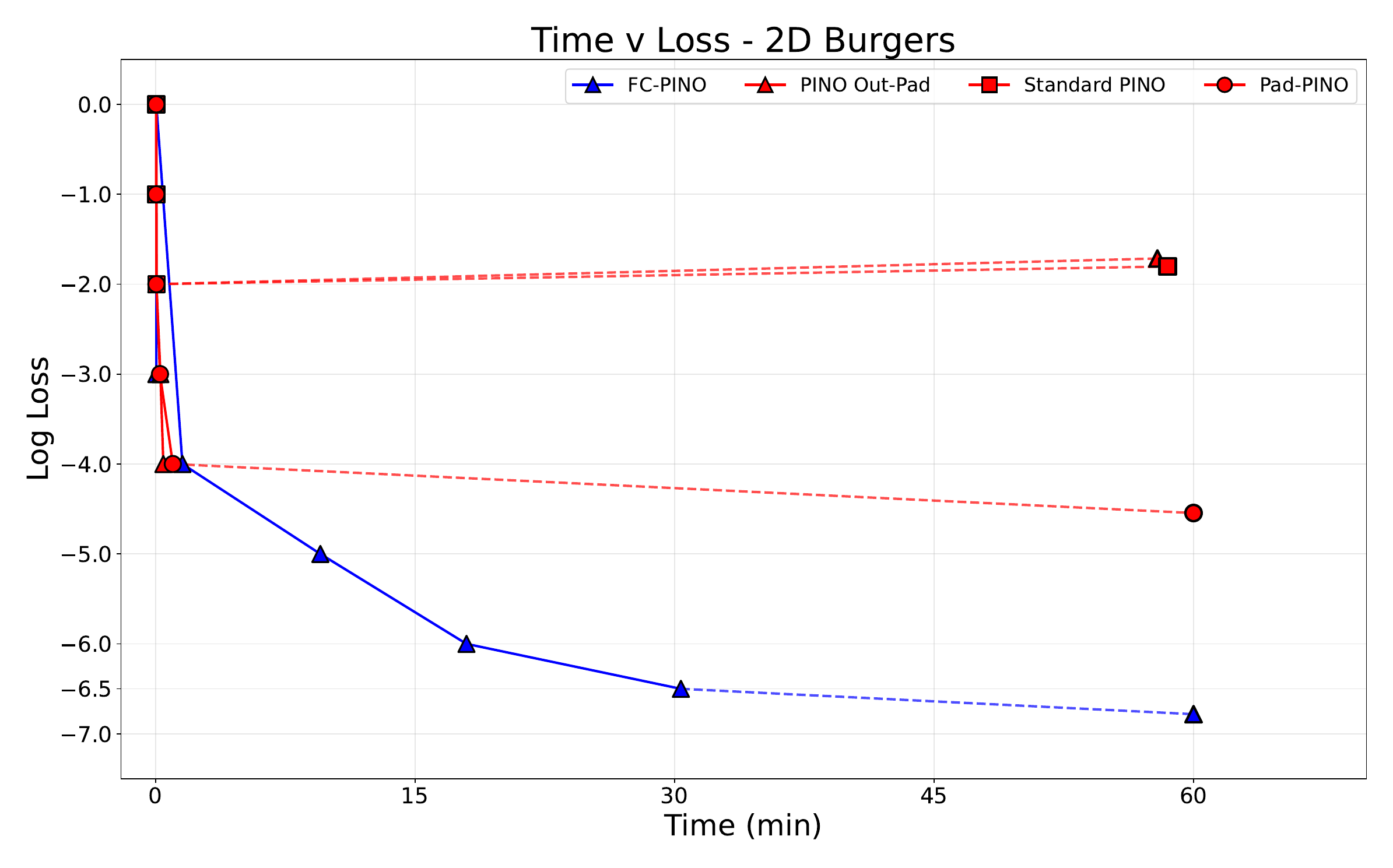} \vspace{-3.5mm}
    \caption{Evolution of the PDE residual loss as a function of time for the 2D Burgers' equation with the different PINO approaches. To allow for a cleaner visualization, we display the time taken to reach certain threshold loss values for the first time as a solid line. The dotted line then connects the loss at the last threshold value and the loss at $60$ minutes. }
    \label{fig:threshold_crossing_2D} 
\end{figure}

\vspace{-2mm}

\subsection{Navier--Stokes Equations}
\label{appx:conv_time_NS}

\vspace{-5mm}

\begin{figure}[h]
    \centering
    \includegraphics[width=0.86\linewidth]{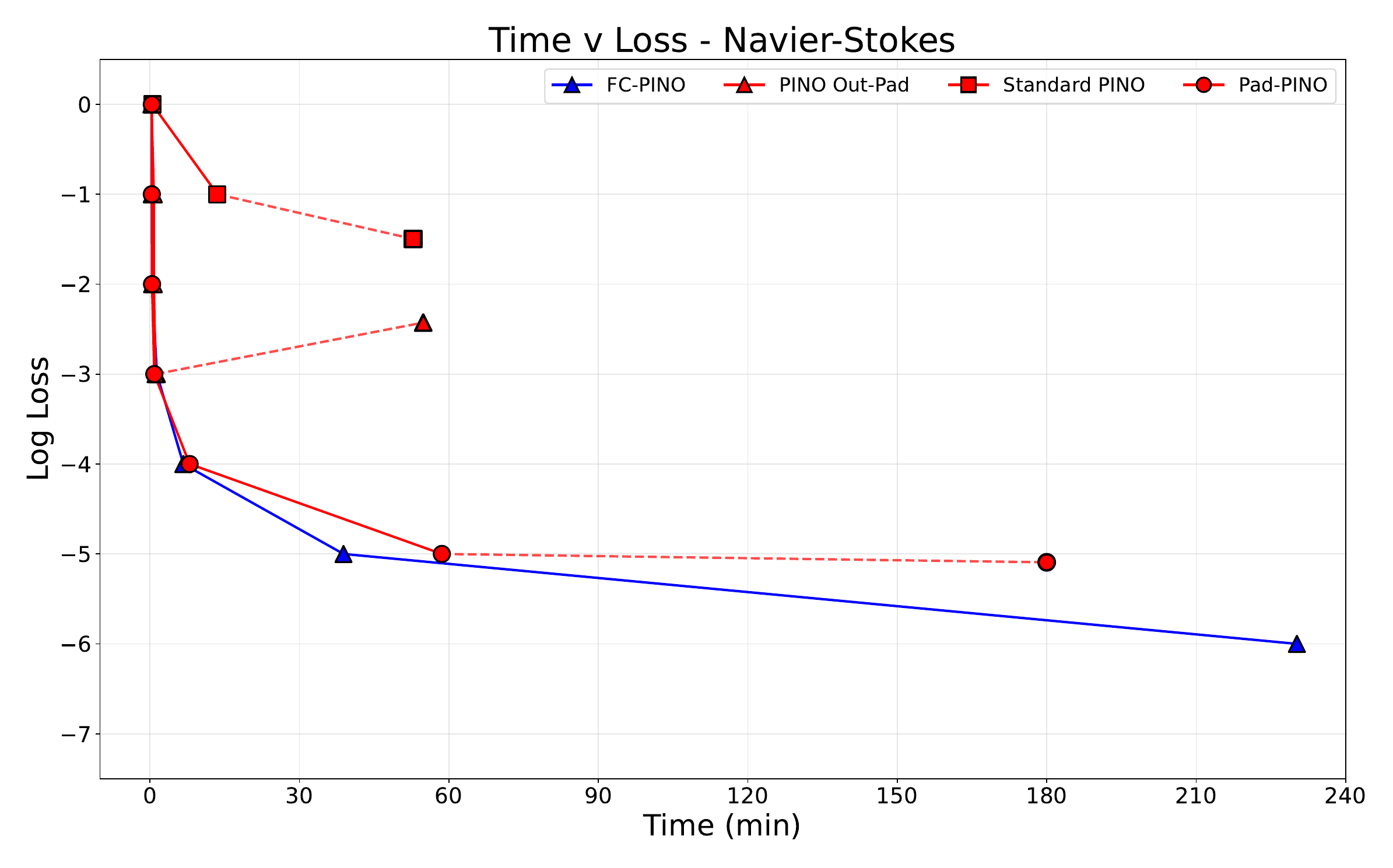} \vspace{-3.5mm}
    \caption{Evolution of the PDE residual as a function of time for the Navier--Stokes equations with the different PINO approaches. To allow for a cleaner visualization, we display the time taken to reach certain threshold loss values for the first time as a solid line. The dotted line then connects the loss at the last threshold value and the final loss at the final value.}
    \label{fig:threshold_crossing_3D} \vspace{-14mm}
\end{figure}

\end{document}